\documentclass[lettersize,journal]{IEEEtran}
\usepackage{amsmath,amsfonts}
\usepackage{algorithmic}
\usepackage{algorithm}
\usepackage{array}
\usepackage[caption=false,font=normalsize,labelfont=sf,textfont=sf]{subfig}
\usepackage[font=normalsize, justification=centering]{caption}
\usepackage{textcomp}
\usepackage{stfloats}
\usepackage{url}
\usepackage{multirow}
\usepackage{verbatim}
\usepackage{booktabs}
\usepackage{array}
\usepackage{graphicx}
\usepackage{cite}
\usepackage[margin=1in]{geometry}
\renewcommand{\arraystretch}{1.3}

\usepackage{booktabs}
\usepackage{makecell}

\begin{document}

\title{Remote Sensing Image Intelligent Interpretation from the Language-Centered Perspective: Principles, Methods and Challenges}

\author{Haifeng Li,~\IEEEmembership{Senior Member,~IEEE},
        Wang Guo,
        Haiyang Wu,
        Mengwei Wu,
        Jipeng Zhang,
        Qing Zhu,
        Yu Liu,
        Xin Huang, ~\IEEEmembership{Fellow,~IEEE},
        Chao Tao
\thanks{
This work was supported by the National Natural Science Foundation of China (Grant Number 42271481). The authors also appreciate the High-Performance Computing Platform of Central South University and HPC Central of the Department of GIS for providing HPC resources. \textit{(Corresponding author: Chao Tao.)}

Haifeng Li, Wang Guo, Haiyang Wu, Mengwei Wu, Jipeng Zhang and Chao Tao are with the School of Geosciences and Info-Physics, Central South University, Changsha 410083, China. (e-mails: lihaifeng@csu.edu.cn; wang\_guo@csu.edu.cn; 245001024@csu.edu.cn; 235007016@csu.edu.cn; 245002055@csu.edu.cn; kingtaochao@csu.edu.cn) 

Qing Zhu is with the Faculty of Geosciences and Environmental Engineering, Southwest Jiaotong University, Chengdu 611756, China. 

Yu Liu is with the School of Earth and Space Sciences, Peking University, Beijing 100871, China. 

Xin Huang is the head of the Institute of Remote Sensing Information Processing (IRSIP), Wuhan University, Wuhan 430072, China. (e-mail: xhuang@whu.edu.cn)
}
}

\maketitle

\begin{abstract}
Remote sensing image interpretation has predominantly followed a visual-centered paradigm, relying on visual feature extraction to achieve semantic understanding. However, faced with increasingly complex interpretation demands, such methods have inherent limitations in cross-modal reasoning, abstract semantic representation, and decision-making. Although large language models (LLMs) have recently been applied to the remote sensing domain, existing research focuses mostly on performance improvements in specific downstream tasks and lacks a systematic theoretical explanation of the core role of language in the cognitive process. This article aims to reveal the paradigm shift in remote sensing interpretation from ``visual-centere'' to ``language-centered'' Inspired by global workspace theory (GWT) in cognitive science, we construct a language-centered remote sensing interpretation framework. This framework positions the LLM as a cognitive hub, coordinating the four spaces of perception, task, knowledge, and action, thereby aiming to achieve a unified loop of understanding, reasoning, and decision-making. We first analyze the potential of LLM as core cognitive components and then systematically review key challenges in realizing this paradigm: multimodal representation, knowledge association and reasoning, and decision-making and execution. To address these challenges, we summarize existing solutions, training datasets and evaluation benchmarks. Finally, we outline future research directions regarding the adaptive alignment of multimodal data, task understanding under dynamic knowledge constraints, trustworthy reasoning, and autonomous interaction. This review aims to provide a theoretical perspective for understanding the role of language in remote sensing cognition and to serve as a reference for building next-generation intelligent cognitive geospatial analysis systems.
\end{abstract}

\begin{IEEEkeywords}
Language-Centered Remote Sensing Interpretation, Global Workspace Theory, Multimodal Representation, Knowledge Association, Reasoning and Decision-Making, Language-based Agent.
\end{IEEEkeywords}

\section{Introduction}
\IEEEPARstart{L}{anguage} and vision, as two fundamental paradigms of human cognition, have maintained a dialectical relationship throughout the development of civilization. Martin Jay\cite{jay1988rise} reveals the profound impact of the privileging of vision on modern cognitive science, where the visual system, owing to its intuitive spatial perception, has been granted epistemological primacy within the framework of empiricist cognition. In contrast, the centrality of language is rooted in the Platonic tradition of idealism and underwent a modern transformation through the works of Descartes, Kant, and ultimately, Wittgenstein's linguistic turn, which emphasized the abstract capacity of language as the foundation of rational cognition. The interplay between these two cognitive paradigms is also reflected in the history of biological evolution: the emergence of visual organs triggered the Cambrian explosion of biodiversity, while the birth of language marked the beginning of the human intelligence revolution. Kant, in his famous dictum ``Thoughts without content are empty, intuitions without concepts are blind."\cite{preussischen1902akademie}, revealed the complementarity between the two: Vision offers the richness of concrete perception, while language provides the stability of abstract representation. Together, they form a closed loop through which humans cognitively engage with the physical world. In the interdisciplinary field of remote sensing science and artificial intelligence, this duality of cognitive paradigms maps onto the evolutionary trajectory of image interpretation technologies. 

Current mainstream remote sensing intelligent interpretation systems adhere to a visual-centered paradigm: they rely on vision models to extract pixel-level or region-level features from imagery and accomplish tasks such as scene classification, object recognition, and target detection through the mapping between features and semantic labels and have achieved significant success in these standardized scenarios\cite{zhu2017deep, li2020object}. However, as remote sensing interpretation tasks continuously expand in both form and reasoning depth, the information obtainable solely from visual signals has become increasingly insufficient. The visual-centered interpretation paradigm, limited by explicit and local perceptual cues, struggles to perform complex reasoning based on deep semantic relationships, leaving them stretched when facing demands for high-order object relationship mining and dynamic semantic reasoning.

\begin{table*}[t]
\caption{Comparison with Related Reviews
\label{tab:1}}
\centering
\begin{tabular}{>{\raggedright\arraybackslash}p{1.6cm} 
                >{\raggedright\arraybackslash}p{1.4cm} 
                >{\raggedright\arraybackslash}p{11.3cm}}
\toprule
\multirow{1}{*}{Reviews} & Year & Key Focus \\

\midrule

\multirow{2}{*}{Li et al.\cite{li2024vision}} & \multirow{2}{*}{2024} & {\textbf{Applications of VLMs}: Focuses on applications in specific tasks such as VQA and image captioning.}\\
 
\midrule

\multirow{2}{*}{Huang et al.\cite{huang2025survey}} & \multirow{2}{*}{2025 Mar} & {\textbf{The transition from vision to multimodality}: Provides a detailed discussion of the technological evolution from CNN/ViT-based vision models to MLLMs.}\\ 

\midrule

\multirow{2}{*}{Xiao et al.\cite{xiao2025foundation}} & \multirow{2}{*}{2025 Jun} & {\textbf{Remote Sensing Foundation Models}: Integration of VFMs, VLMs, and LLMs; highlights general-purpose applicability and zero-shot transferability.}\\
 
\midrule

\multirow{2}{*}{Lu et al.\cite{lu2025vision}} & \multirow{2}{*}{2025 Sep} & {\textbf{Vision Foundation Models}: Focuses on VFM (e.g., SAM, MAE) for pixel-level and region-level perception tasks in remote sensing.}\\
 
\midrule

\multirow{2}{*}{\textbf{Ours}} & \multirow{2}{*}{-} & {\textbf{Language-Centered Interpretation Paradigm}: Incorporates global workspace theory (GWT), positioning language as the cognitive hub to shift \textbf{from passive interpretation to active interaction and decision-making}.}\\
 
\bottomrule
\end{tabular}
\end{table*}

To surmount the bottlenecks of the aforementioned visual-centered paradigm in terms of semantic reasoning and interaction capabilities, intelligent remote sensing interpretation technology has undergone a series of significant evolutions and iterations. Visual foundation models (VFMs)\cite{sun2022visual, manas2021seasonal, lu2025vision} significantly enhanced general feature representation capabilities through large-scale pretraining; however, they remained essentially confined to visual pattern matching, lacking the explicit expression of abstract concepts. Subsequently, the introduction of vision-language models (VLMs)\cite{li2024vision} overcame this limitation, preliminarily achieving cross-modal semantic association by establishing alignment relationships between visual features and linguistic semantics. With breakthroughs in large language model technology, multimodal large language models (MLLMs)\cite{huang2025survey} emerged. Works such as RS-LLaVA\cite{bazi2024rs} and EarthGPT\cite{zhang2024earthgpt} deeply integrated visual encoders with LLMs, utilizing the language modality to complement high-level semantic information missing in visual cues and unifying the interaction interface for multiple tasks. Recent research has further pushed this trend to the stage of LLM-based agents\cite{xu2024rs, shabbir2025thinkgeo}. For instance, models such as EarthMarker\cite{zhang2024earthmarker} treat the LLM as a cognitive hub, endowing the interpretation system with capabilities for proactive planning, complex logical reasoning, and decision-making. This technological evolutionary trajectory from  perception (VFM) to alignment (VLM) and then to understanding (MLLM) and cognition (agent) reveals that remote sensing interpretation is undergoing a fundamental paradigm shift: transitioning from a ``visual-centered paradigm" that relies on explicit visual features to a ``language-centered paradigm" centered on semantic understanding and logical reasoning.

Although several recent reviews have systematically summarized new breakthroughs in the field of remote sensing image interpretation (as shown in Table\ref{tab:1}), they focus primarily on the categorization of technical implementations and performance comparisons. For instance, Lu et al.\cite{lu2025vision} focus on the architecture and pretraining strategies of VFMs, emphasizing pixel-level feature representation capabilities; Li et al.\cite{li2024vision} and Huang et al.\cite{huang2025survey} discuss in detail the application of VLMs in multimodal alignment and downstream tasks (such as VQA and image captioning). However, most of these works remain at the level of ``instrumental rationality", focusing on ``how to train stronger models" or ``how to fuse more data", while rarely offering reflections at the level of cognitive essence: Why does language occupy a central position in the evolution of remote sensing interpretation? What kind of cognitive paradigm improvement does the shift from visual feature mapping to linguistic logical reasoning truly signify?

\begin{figure*}[!t]
\centering
\includegraphics[width=\textwidth]{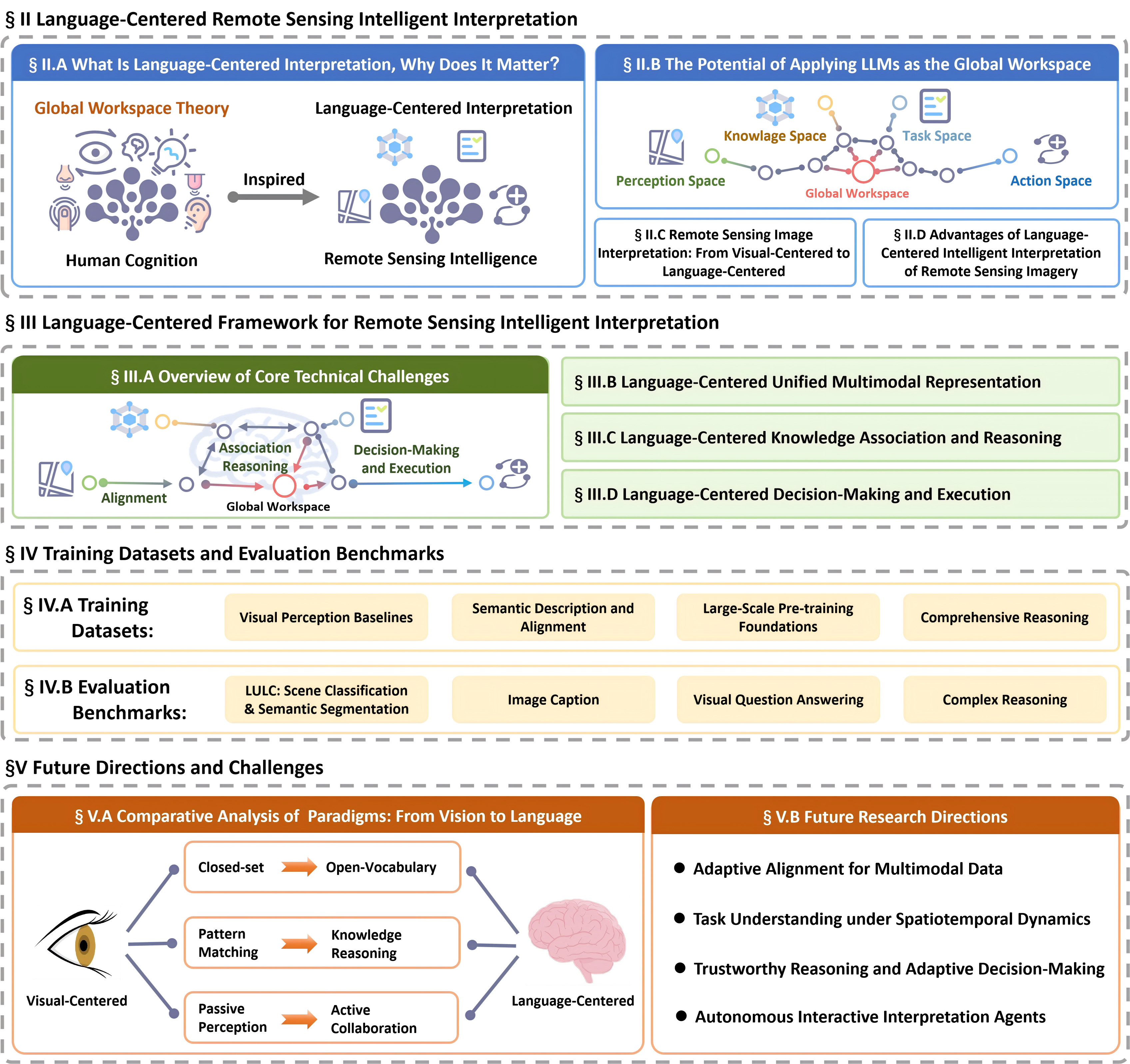}%
\caption{Overall structure}
\label{overview}
\end{figure*}

To bridge this gap, this paper aims to establish a language-centered cognitive theoretical framework. We introduce global workspace theory (GWT)\cite{baars2005global} from cognitive science, redefining the LLM from a mere ``interaction interface" into a ``cognitive hub". The overall structure of this review is illustrated in Fig.\ref{overview}. Its main contributions are as follows:

\begin{enumerate}
\item{\emph{Rethinking Remote Sensing Image Interpretation from a Cognitive Perspective}: Starting from the human cognitive process, this review analyzes the cognitive limitations of current remote sensing image interpretation technologies and emphasizes a language-centered ``perception-cognition-decision" paradigm, thereby laying a theoretical foundation for future research.}

\item{\emph{Global Workspace-Driven Interpretation Framework}: In light of recent advances in LLMs, this work explores the potential application of GWT in remote sensing image interpretation. It provides a systematic review of relevant technologies and analyzes the feasibility of implementing such a framework within the current technological landscape.}

\item{\emph{Key Challenges and Technical Bottlenecks in Building a Language-Centered Intelligent Interpretation Framework}: This work summarizes the core issues faced in implementing language-centered remote sensing interpretation frameworks. It focuses on three major challenges: 

\begin{itemize}
    \item The difficulty of constructing unified multimodal representations; 
    \item The lack of effective mechanisms for combining multisource knowledge and reasoning;
    \item The limited decision-making and execution capabilities. 
\end{itemize}

}

\item{\emph{Outlook and Emerging Research Opportunities}: Finally, this review summarizes and discusses the new opportunities brought by this language-centered paradigm.}
\end{enumerate}

\begin{figure*}[!t]
\centering
\includegraphics[width=0.8\textwidth]{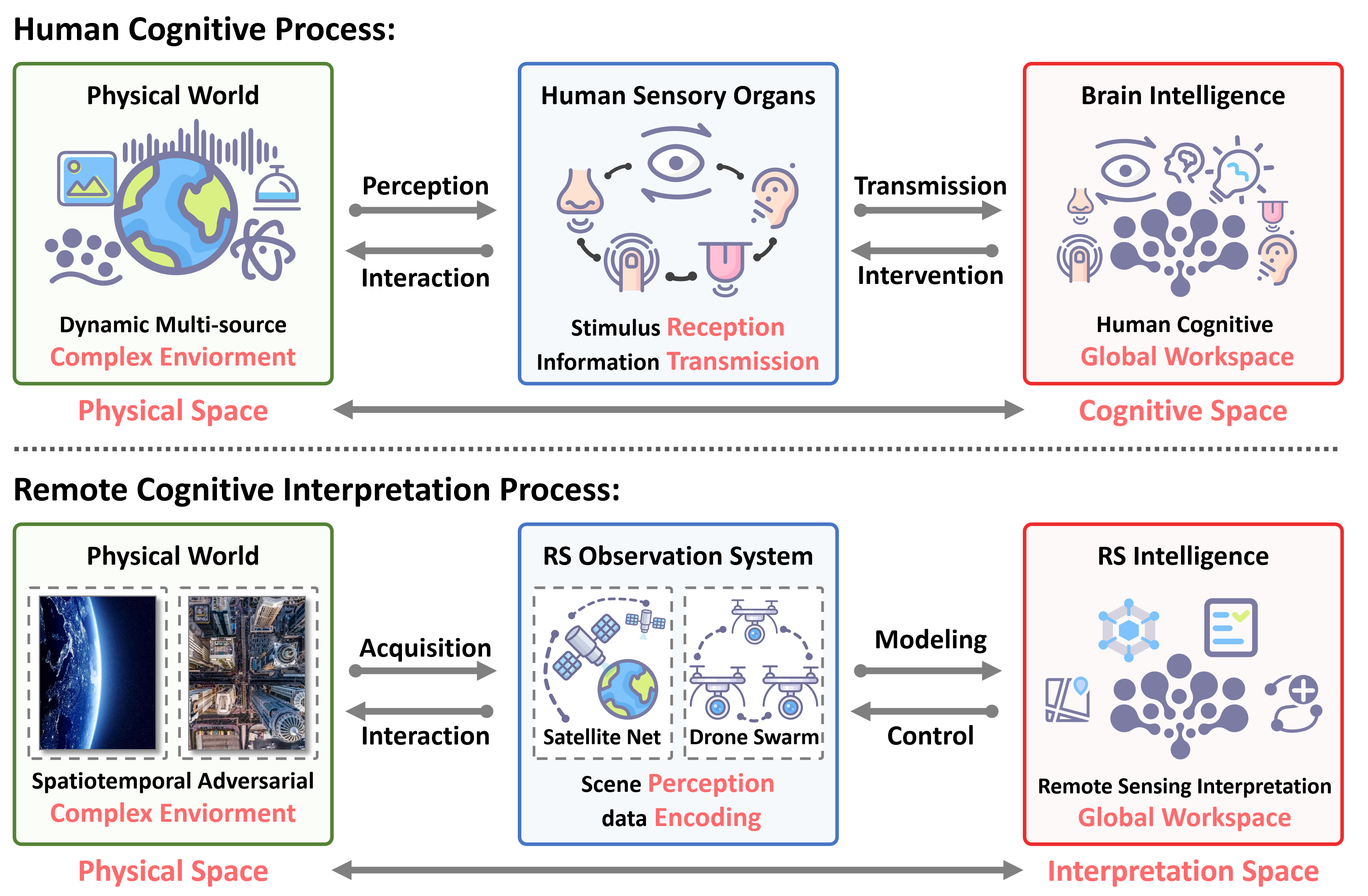}%
\caption{Correspondence between Human Cognition and Intelligent Remote Sensing Interpretation}
\label{Cognitive}
\end{figure*}

\section{Language-Centered \\ Remote Sensing Intelligent Interpretation}

\subsection{What Is Language-Centered Interpretation and Why Does It Matter?}

Language-centered remote sensing (RS) interpretation is a paradigm that employs the language modality as a unified semantic interface to organize perception, task representation, and knowledge reasoning. As a symbolic system, language represents concepts, attributes, and relationships through discrete symbols and combinatorial rules, enabling information to be represented and manipulated in a structured and compositional manner. In cognitive studies, language is regarded as the container of human thought\cite{carruthers2002cognitive}, providing the essential mechanism for concept formation, relationship building, and high-level reasoning, which allows knowledge to be organized and controlled at an abstract level. In 1998, Bernard Baars proposed global workspace theory (GWT)\cite{baars2005global}. This theory posits that the brain contains a ``global workspace" that serves as a shared platform for information convergence and interaction. It integrates content from diverse sources, such as perception and memory, to support complex cognitive functions. Within this framework, as shown in Fig. \ref{Cognitive}, perceptual information from complex physical scenes and knowledge stored in memory intersect and correlate within the workspace, driving reasoning processes and generating actionable decisions under the guidance of intent. Subsequent actions affect the external world, and the resulting environmental changes provide feedback to the cognitive system through new perceptual inputs, thereby forming a continuously updating cycle of interactive engagement. 

Inspired by this cognitive framework, {\bf{intelligent remote sensing (RS) interpretation can be viewed as a continuous, cyclic interactive process}}. Complex and dynamic spatiotemporal scenes are first captured by RS observation systems and transformed into digitized perceptual information. Once parsed and modeled, this information enters the ``global workspace" of the interpretation system to support scene understanding and task reasoning (as shown in Fig. \ref{Cognitive}). Guided by the decisions generated within the workspace, the system issues commands back to the observation equipment. Based on the requirements of the interpretation task, it proactively acquires supplementary information to continuously refine and optimize its understanding of the external world. In this context, RS observation represents a bottom-up perceptual process, whereas RS interpretation employs top-down strategy adjustments to secure necessary information; together, they form a continuous closed loop. Within this closed-loop framework, language serves as the pivotal mechanism for task expression, semantic organization, and reasoning guidance. This feature enables the system to transcend the limitations of purely visual representations, endowing it with enhanced task-expressive power, deeper reasoning capabilities, and a more robust basis for decision-making.

\begin{figure*}[!t]
\centering
\includegraphics[width=0.7\textwidth]{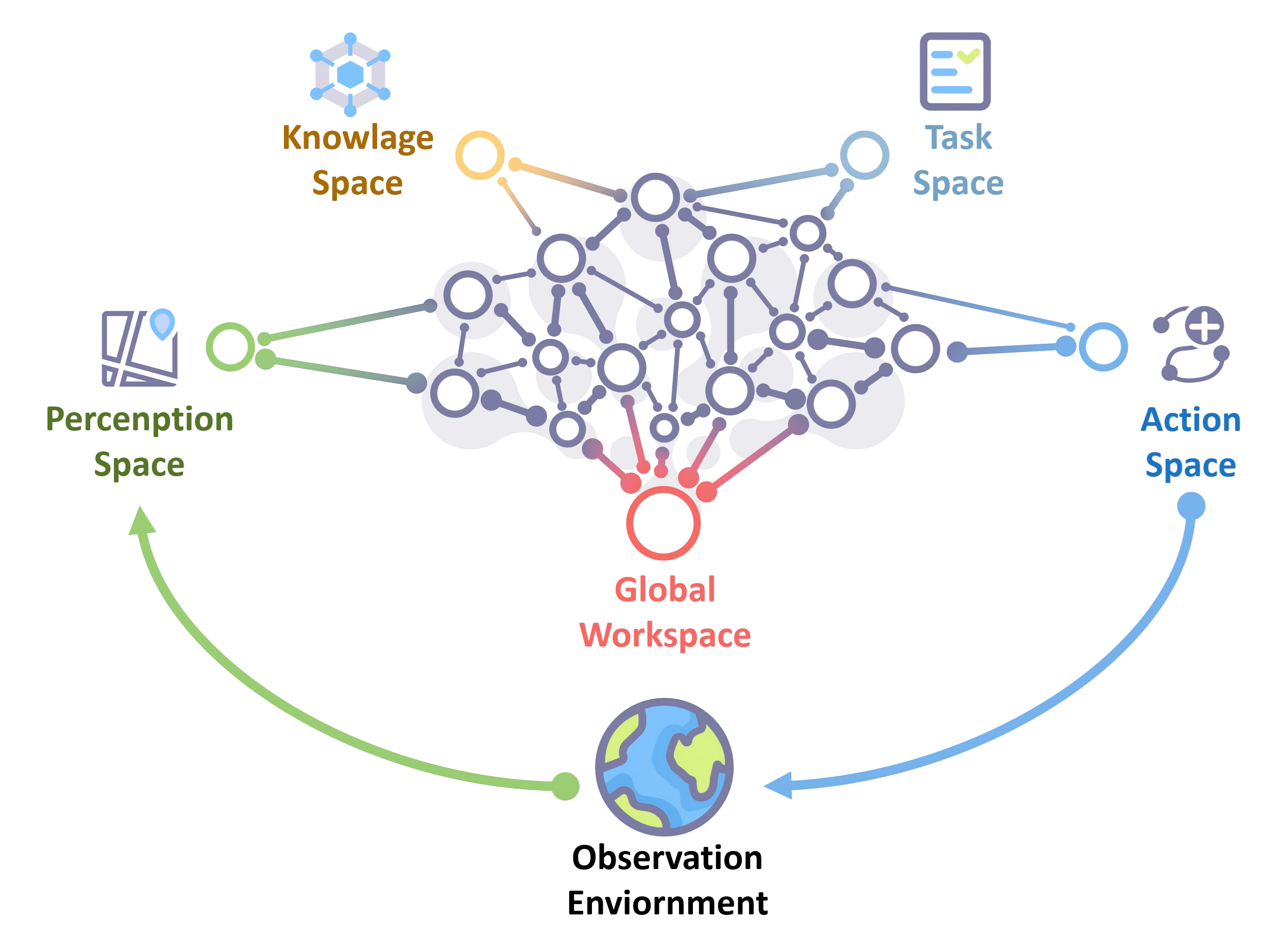}%
\caption{Cyclic Interactive Interpretation Driven by a Global Workspace}
\label{GWT}
\end{figure*}

This process can be driven by a global workspace and formally represented as cyclic interaction interpretation:

\[
C = \left( G(P, K, T),\ A \right)
\]

The entire process is shown in Fig.\ref{GWT}, where

\begin{itemize}
\item{$P$ represents the perception space and denotes information such as scene context, object characteristics, and spatiotemporal attributes. This space captures a detailed description of geo-objects within the observation environment, including their spectral signatures, geometric properties, texture features, and temporal dynamics.}
\item{$K$ represents the knowledge space, referring to supplementary information derived from remote sensing expert knowledge. This information includes prior knowledge of object features, geographical distribution patterns, seasonal variation rules, and other details. Such knowledge can be acquired through expert experience or accumulated historical data and is used during the interpretation process to enhance the understanding and reasoning of observed information.}
\item{$T$ denotes the task space, which represents the user's interpretation requirements. Tasks within the task space may range from simple (such as identifying the object category in an image) to more complex (such as recognizing the fine-grained category of a ship in the image when key visual features are missing).}
\item{$A$ represents the action space, which refers to the set of executable actions available to the sensing system. These actions include adjustments to the sensor's resolution, intrinsic parameters, sensing modalities, combinations of modalities, and changes in observation angles and orientations.}
\item{$G$ denotes the global workspace, which integrates information from $P$ (perception space), $K$ (knowledge space), and $T$ (task space) and outputs control instructions to $A$ (action space).}
\end{itemize}

In the cyclic interaction interpretation process, the global workspace encodes and integrates the visual representations from $P$, the remote sensing expert knowledge from $K$, and the interpretation requirements from $T$ into a unified representation. It first establishes associations between perceptual information and expert knowledge and performs reasoning based on these connections. Guided by the interpretation requirements, the global workspace then makes decisions based on the reasoning outcomes and finally executes the determined interactive actions in $A$.

\subsection{The Potential of Applying LLMs as the Global Workspace}

With the rapid development of LLMs, such as LLaMA\cite{touvron2023llama}, GPT\cite{brown2020language} and DeepSeek\cite{bi2024deepseek}, these models have been shown to effectively encode world knowledge and exhibit logical reasoning capabilities. This paper argues that LLMs possess strong potential to serve as the global workspace within remote sensing intelligent interpretation systems for the following reasons:

\paragraph{LLMs can construct unified representations across multimodal data} The Platonic Representation Hypothesis\cite{huh2024position} posits that data from different modalities are essentially expressions of the same underlying reality across different dimensions, exhibiting a tendency toward consistent convergence. Utilizing the language modality as the unified convergence point facilitates cross-modal information association and reasoning. In terms of visual modality representation, pioneering vision-language models such as CLIP\cite{radford2021learning}, ALIGN\cite{jia2021scaling}, and BLIP\cite{li2022blip} map visual and linguistic modalities into a shared semantic space through contrastive learning. These works effectively enhance the model's deep understanding of object relationships, confirming that visual and language modalities can achieve consistent convergence through shared semantics. With respect to multimodal representation, LanguageBind\cite{zhu2023languagebind} employs contrastive learning to align various modalities--including video, infrared, depth maps, and audio--with the language modality, hence achieving multimodal embedding into the latent space of LLM. Furthermore, AllSpark\cite{shao2025allspark} embeds 13 distinct modalities into a language-centered representation.

\paragraph{LLMs can associate perceptual information with existing knowledge} Through training on large-scale corpora, LLMs encode the logical relationships within language, as manifested in their strong logical reasoning capabilities\cite{anand2024mathify,ahn2024large}. This feature enables LLMs to first comprehend perceptual information in multimodal data and then match it with previously learned knowledge, thus driving advancements across multiple fields. In terms of information understanding, LLMs can not only understand tasks in algebra, calculus, and geometry by recognizing objectives and variable conditions, thereby constructing appropriate problem-solving logic but also handle geographic data mining and analysis tasks, gaining deep insights into task objectives such as spatial patterns, geographic entity relationships, and trend analysis. With respect to knowledge association, the knowledge linking ability of LLMs stems from the vast amount of world knowledge, linguistic patterns, and conceptual associations accumulated during their training process. This knowledge serves as the model's ``long-term memory", allowing the model to quickly integrate relevant information and provide explanations when faced with complex inputs. The PaLM model proposed by Chowdhery et al.\cite{chowdhery2023palm} demonstrates that the large-scale parameters of LLMs are crucial for enhancing retrieval capabilities. This characteristic allows the model to associate learned knowledge with minimal prompting in complex contexts, thus enabling it to make accurate responses.

\paragraph{LLMs can handle complex task reasoning and decision-making} Reasoning is essentially the process of drawing conclusions from known information or premises through logical inference chains \cite{shao2025asking,yang2025learning}. LLMs achieve reasoning and decision-making for complex tasks by constructing inference chains and progressively deducing them from the associated information. Wei et al.\cite{wei2022chain} explored the role of chain-of-thought prompting in large language models and demonstrated that by clearly defining reasoning links, LLMs can not only better understand the task itself but also efficiently solve complex tasks that require multiple reasoning steps. Moreover, reasoning and decision-making for complex tasks often require LLMs to dynamically adjust their reasoning chains based on task objectives. Relevant studies have shown that LLMs can optimize their decision-making process by adjusting their reasoning chains. An example of this capability is the DeepSeek-R1\cite{guo2025deepseek} model, which can adjust its reasoning chain and refine its reasoning strategy before responding. By incorporating feedback information, it can correct previous errors, thereby solving complex tasks. For instance, during the AIME 2024 test, DeepSeek-R1\cite{guo2025deepseek} successfully solved 79.8\% of high-difficulty problems by reflecting on and adjusting its reasoning steps.

\begin{figure*}[!t]
\centering
\includegraphics[width=0.9\textwidth]{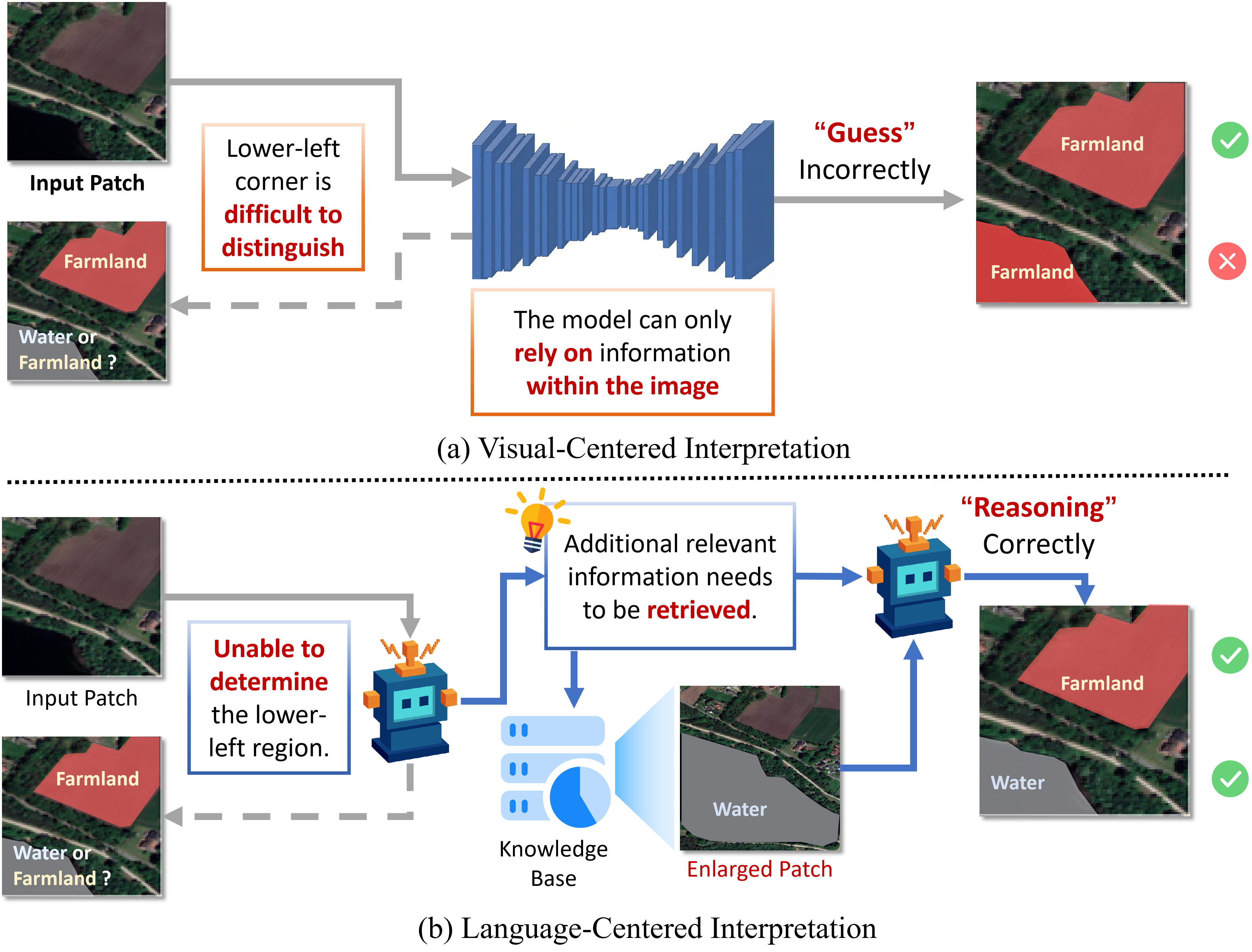}%
\caption{Paradigm Shift in Remote Sensing Image Interpretation: From Visual-Centered to Language-Centered}
\label{paradigm_shift}
\end{figure*}

\subsection{The Paradigm Shift in Remote Sensing Interpretation: From Visual-Centered to Language-Centered}
Currently, the mainstream approach to remote sensing intelligent interpretation remains dominated by a visual-centered passive perception paradigm, which executes tasks through end-to-end inference mechanisms. Although recent visual-based geospatial foundation models (GFMs)\cite{bountos2025fomo, hong2023spectralgpt, wu2025semantic} have significantly advanced feature extraction and heterogeneous alignment across multimodal data, they remain constrained by three inherent limitations as task complexity and reasoning depth increase:

\begin{itemize}
\item{\emph{Semantic Gap}: These models rely on data alignment rooted in physical consistency, making it difficult to bridge the semantic divide. Recent works such as DOFA\cite{xiong2024neural} and msGFM\cite{han2024bridging} essentially rely on spatiotemporal consistency to construct feature associations. Although effective at integrating heterogeneous visual signals such as optical, SAR, hyperspectral, and infrared imagery, this mechanism struggles to bridge the gap between 'pixel-level features' and 'abstract concepts.' Particularly when addressing abstract semantics that lack direct physical manifestations (e.g., 'factory abandonment due to economic recession' or 'complex tactical intents'), purely visual models are unable to transcend the cognitive barrier from low-level signals to high-level logic, primarily because of the lack of a symbolic semantic interface.}

\item{\emph{Closed Knowledge System}: It is challenging to incorporate open-ended external knowledge. The knowledge within the VFM is derived solely from the pixel patterns in the training data. This ``what you see is what you get" learning style constructs a closed system, preventing the direct utilization of vast external knowledge bases (e.g., historical observations, geographic laws, or expert expertise). In contrast, language models possess an innate ability to connect with world knowledge. Lacking such symbolic knowledge injection channels, visual models often show a deficit in generalization when handling complex tasks requiring background reasoning, such as long-tail object recognition or disaster causality analysis.}

\item{\emph{Passive Perception}: Limited by a static mapping paradigm. Regardless of their feature extraction strength, existing visual models exhibit unidirectional ``input image to output result" mapping. They lack agency; they cannot understand complex, unstructured intents such as an agent, nor can they perform reflection or proactive planning (such as invoking external tools or adjusting observation perspectives) when information is insufficient. This lack of interactivity renders visual-centered models passive data analysis tools rather than intelligent cognitive systems capable of autonomous problem solving.}

\end{itemize}

Consequently, the interpretation process of the visual-centered paradigm (as illustrated in Fig. \ref{paradigm_shift}(a) is restricted to judgments based on local, explicit clues within the current field of view. It lacks the capacity for cross-scale association, intent understanding, and information completion. When visible information is insufficient, the model cannot infer strategic actions such as ``further observation required" or ``larger-scale imagery needed," thus leading to erroneous results. As shown in the figure, owing to missing visual cues, the model misidentifies ``water" in the lower-left corner as ``farmland."

In contrast to passive perception, the language-centered interpretation paradigm emphasizes the organic integration of bottom-up perception and top-down semantic reasoning (Fig. \ref{paradigm_shift}(b)). In this framework, perception provides explicit clues, while language-led cognition infers results based on semantic structure and directs further observation. As illustrated, the model initially struggles to classify the lower-left area. The language reasoning module then proactively requests ``more perceptual information," prompting the system to retrieve an enlarged patch from the database. This additional visual context enables the model to accurately identify the true category.

In this language-centered paradigm, perception-driven visual processing and cognition-driven semantic reasoning are effectively integrated. This integration allows the system to move beyond basic image analysis to perform complex reasoning and information completion using language-organized knowledge. This ``perception (extracting image information), cognition (water or farmland?), decision (finding more information) and action (searching the knowledge base)" mechanism not only enhances generalization but also enables the system to robustly handle interpretation tasks with high structural complexity and rich semantics.

\subsection{Advantages of Language-Centered Intelligent Interpretation of Remote Sensing Imagery}

The language-centered intelligent interpretation paradigm mimics the fundamental human way of understanding the world. By integrating the spaces of perception, knowledge, task, and execution, it constructs a cyclic interpretation mode capable of effective interaction with the physical world. In this mode, language transcends its role as a mere communication tool, functioning instead as the central hub for knowledge integration, orchestration, and decision-making. Through this language-driven cyclic process, the model can address complex tasks in open environments, achieving transformation from passive environmental perception to active environmental interaction. The rapid advancement of LLMs has rendered this paradigm feasible. The capabilities of LLMs in unified representation, knowledge association, and reasoning-based decision-making overcome the inherent limitations of traditional visual models caused by the strong coupling of data, architectures, and tasks. Consequently, language-centered models demonstrate the following advantages in remote sensing interpretation:

\paragraph{Language-centered interpretation models demonstrate significant advantages in integrating multimodal perceptual data} Stemming from variations in intrinsic imaging mechanisms (e.g., spatial resolution and spectral characteristics) and extrinsic acquisition conditions (e.g., spatiotemporal coverage), multimodal data exhibit high heterogeneity at the structural level. The traditional visual-centered paradigm struggles to bridge this physical gap, often necessitating strict pixel-level spatiotemporal alignment to achieve effective fusion. In contrast, the language-centered paradigm effectively circumvents this bottleneck. Although remote sensing data, ranging from optical and infrared to SAR and LiDAR, diverge sharply in terms of physical attributes, from a language-centered perspective, they are essentially different representations of the same geo-object. Consequently, the language modality serves as a unified semantic anchor, enabling deep association and unified representation of multimodal data at the semantic level without the need for strict physical alignment.

\paragraph{Language-centered interpretation models possess superior open-environment learning capabilities, enabling flexible adaptation to unknown scenes and the rapid integration of new knowledge} In contrast, traditional visual models rely primarily on the mapping between ``massive imagery and limited labels" as their supervision signal. This paradigm suffers from fundamental theoretical limitations: discrete and finite semantic labels fail to encapsulate the rich information regarding ground objects and their spatiotemporal relationships. This shortcoming inevitably confines the model's cognitive boundaries within a predefined closed-set knowledge domain, resulting in severely restricted generalization capabilities when identifying unlabeled or novel scenes. Conversely, large language models (LLMs) associate visual perceptual information with expansive open-world knowledge systems. This feature allows models to effectively overcome the cognitive bottlenecks imposed by limited annotation categories, achieving zero-shot understanding of unseen concepts.

\paragraph{Language-centered interpretation models demonstrate significant advantages in handling complex cognitive tasks} The traditional visual-centered paradigm typically adopts a ``one-model-per-task" architecture, necessitating the construction of independent, specialized models for distinct tasks such as object detection and change detection. This approach not only incurs high development and maintenance costs but also creates severe semantic barriers between models, making it difficult to address complex interpretation requirements that demand multistep reasoning. In contrast, language-centered models transcend these rigid task boundaries. Leveraging the powerful logical reasoning capabilities and chain-of-thought (CoT)\cite{wei2022chain} mechanisms of LLMs, these models automatically decompose complex macrointerpretation tasks (e.g., ``analyzing port traffic") into executable atomic sub-tasks (e.g., ``identifying vessels," ``counting objects," and ``assessing status"). By mapping multimodal perceptual information into a unified linguistic space, the model can perform logical deduction and decision-making akin to the capabilities of human experts. This paradigm shift from ``task-specific mapping" to ``general-purpose logical planning" not only drastically reduces the complexity of multitask systems but also endows interpretation models with ``generalist" capabilities to flexibly handle diverse, nonstandard requirements in rapidly changing open environments.

In summary, compared with traditional visual-centered interpretation models, language-centered models demonstrate clear advantages. Leveraging their rich knowledge association capabilities and powerful reasoning abilities, language models can actively acquire and integrate multisource data, effectively addressing the challenges posed by open remote sensing scenarios and complex interpretation tasks. The shift from a visual-centered to a language-centered paradigm broadens the scope of intelligent remote sensing image interpretation and enhances the model's adaptability and interpretive accuracy in contexts characterized by multisource data, open environments, and complex task demands.

\begin{figure*}[!t]
\centering
\includegraphics[width=\textwidth]{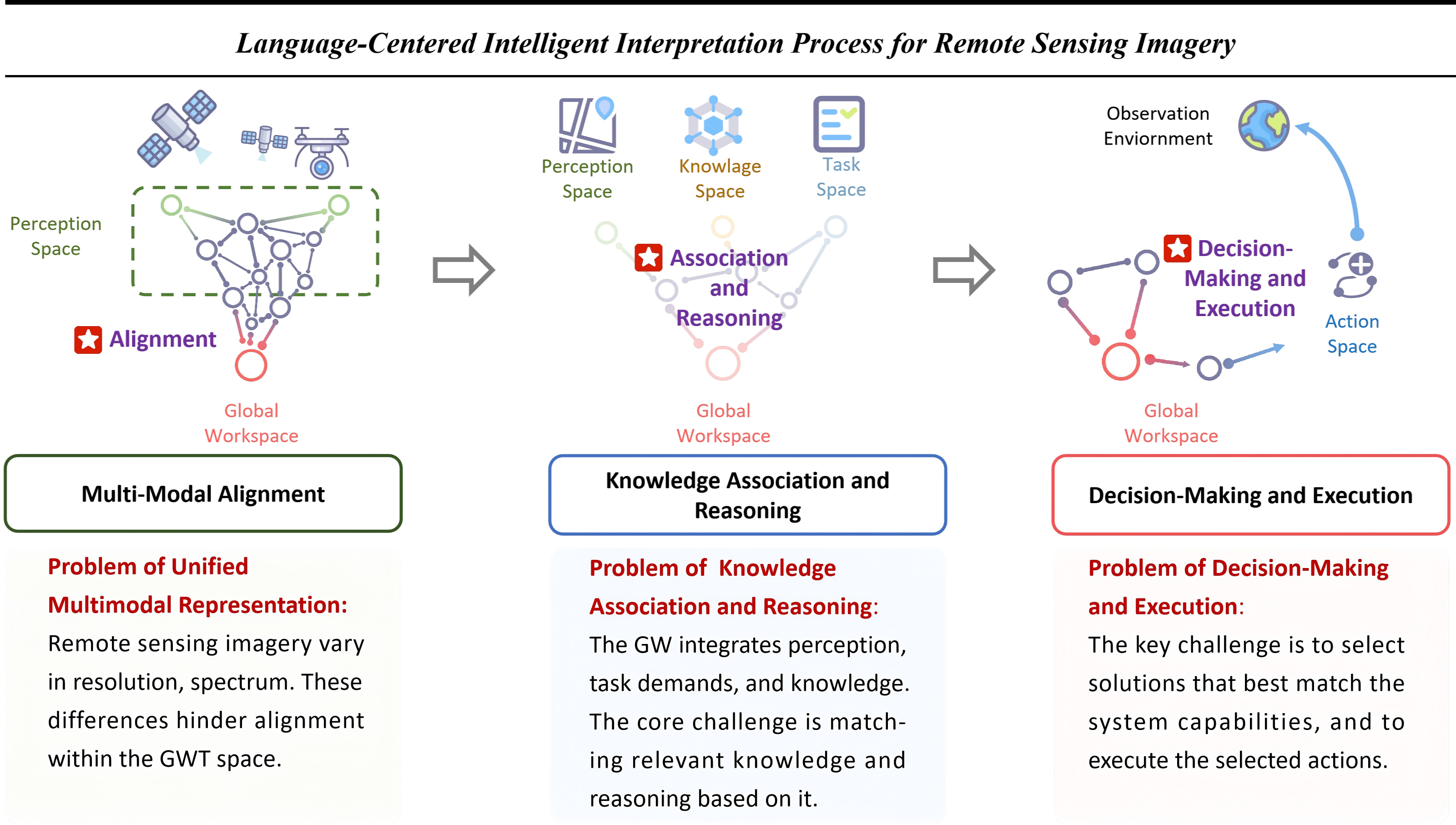}%
\caption{Challenges and key problems in building a language-centered framework for intelligent remote sensing image interpretation.}
\label{gwt_problems}
\end{figure*}

\section{A Language-Centered Framework for Intelligent Interpretation of Remote Sensing Imagery}

Drawing upon GWT, this section proposes a language-centered intelligent interpretation paradigm for remote sensing imagery, aiming to construct a complete interpretation closed loop spanning from multimodal perception to high-level cognition and, finally, to action feedback (as illustrated in Fig. \ref{gwt_problems}). This paradigm abstracts complex interpretation tasks into three progressive cognitive stages. First, heterogeneous multimodal perceptual information ($P$), such as optical and radar data, is mapped onto a unified global workspace through an alignment mechanism to establish a unified perceptual representation. Second, dynamic associations among perceptual information, remote sensing expert knowledge ($K$), and task requirements ($T$) are facilitated within the workspace to conduct deep semantic reasoning. Finally, decision-making is performed based on reasoning results and translated into specific executable actions ($A$) to advance the interpretation process.

However, while this paradigm offers a clear logical framework for the synergy of ``perception-cognition-action", its practical implementation faces critical challenges from physical alignment and cognitive reasoning to decision execution. Specifically, these challenges include overcoming the physical heterogeneity of multimodal data to achieve a unified representation; ensuring precise knowledge association and reasoning within complex knowledge bases; and selecting optimal solutions from ambiguous inferences for efficient decision-making and execution. In this section, these three core issues are explored in depth.

\subsection{Overview of Core Technical Challenges }

\subsubsection{Unified Multimodal Representation}

In a language-centered remote sensing intelligent interpretation system, the alignment of multimodal data representations is key to achieving effective cross-modal information fusion. The central challenge lies in how to map heterogeneous information from different sensing modalities into a unified global workspace for integration and analysis. However, aligning multimodal remote sensing data to a shared language space still faces several significant challenges:

\paragraph{Structural alignment across modalities} Nonlinguistic modalities differ significantly from language modalities in terms of data representation formats, making it difficult to embed them into a shared semantic space. For example, RGB, MSI, and SAR imagery are typically represented as dense 2D matrices, while LiDAR point cloud data are stored as sparse 3D point sets. These structural inconsistencies can lead to a loss of perceptual information when nonlinguistic modalities are mapped onto a 1D language-based semantic space, thereby affecting the effectiveness of multimodal fusion. Therefore, one of the fundamental challenges for language-centered unified representation is how to standardize the representation of nonlinguistic modalities in a way that preserves their perceptual content while aligning them with language-based representations.

\paragraph{Semantic alignment across modalities} Language modalities carry human abstract conceptual systems, whereas nonlanguage modalities are direct samplings of the physical world. This fundamental difference means that aligning their semantics typically requires additional human prior knowledge. However, in the field of remote sensing, nonlanguage modality data are vast and rich in spatiotemporal information, making it extremely costly and labor intensive to achieve precise alignment between language and nonlanguage modalities through manual annotation. Furthermore, there are often significant mismatches in granularity and informational content between the two modalities. For instance, the term ``ship" in natural language typically refers to the entire vessel, whereas in remote sensing imagery, the representation of a ship may consist of various individual visual components. Direct cross-modal alignment without proper abstraction may result in semantic ambiguity. Thus, a core challenge for language-centered multimodal fusion is how to establish accurate semantic alignment mechanisms while minimizing reliance on manual annotation.

\subsubsection{Knowledge Association and Reasoning}
The global workspace contains a complex composition of knowledge, and the ability to associate and integrate knowledge from different sources plays a critical role in solving complex remote sensing image interpretation tasks. Knowledge association is essential for enabling the system to make correct reasoning and decisions. Therefore, achieving accurate knowledge association is among the primary challenges in building a language-centered intelligent interpretation system for remote sensing imagery. Knowledge association is a complex process that requires the establishment of effective links among perceptual information, task requirements, and both internal and external knowledge sources. Specifically, it involves two key steps: task understanding and knowledge retrieval.

\paragraph{Understanding remote sensing interpretation tasks} Task understanding is a prerequisite for accurate knowledge association. A remote sensing interpretation system must deeply comprehend the task requirements and relate them to perceptual information to determine which pieces of information are most relevant. For example, in a fine-grained ship classification task, different types of ships within the same general category often differ only in specific subtle features. Therefore, the interpretation system must leverage internal knowledge to first identify the coarse category of the target ship and then analyze the image content to determine which visual features--such as the position or number of ship islands or the hull number--are most helpful for fine-grained classification. This approach enables the system to more efficiently guide the execution of subsequent recognition steps.

\paragraph{Knowledge retrieval in remote sensing image interpretation} Knowledge retrieval is key to accurately extracting relevant information. Remote sensing interpretation systems must not only utilize the knowledge encoded within the model but also possess the ability to extract task-relevant knowledge from external multimodal sources. Continuing with the example of fine-grained ship classification, once the system has fully understood the task requirements, it must retrieve and extract information related to ship subcategories and their distinctive features from both internal and external knowledge sources. For instance, the Queen Elizabeth-class aircraft carrier is equipped with two ship islands: this type of feature can serve as crucial evidence in subsequent reasoning and decision-making processes and directly affects the system's interpretation performance. In practical terms, knowledge retrieval involves two core technical challenges: efficient knowledge encoding and reliable knowledge retrieval. The former requires the model to embed various forms of data--such as structured data, unstructured text, or temporal information--into its internal parameters while ensuring that the information is well organized and meaningfully associated to avoid knowledge confusion. The latter demands that the model be able to rapidly locate and retrieve information highly relevant to the current task from a vast internal and external knowledge space by leveraging the complex associative relationships among knowledge elements.

\subsubsection{Decision-Making and Execution}
Accurate decision-making and execution are the critical links driving the cyclic interactive interpretation process and bear the responsibility of translating cognitive reasoning from the global workspace into actual physical or digital operations. At this stage, the interpretation system must schedule matching interpretation tools from a capability library or issue specific control instructions to the observation system based on the task intent generated by reasoning. This process is not a simple unidirectional transmission of instructions but a complex mapping facing the dual challenges of the semantic gap and environmental dynamics. Specifically, this phase confronts the following two core challenges:

\paragraph{Tool Decision-Making} The primary challenge facing the interpretation system is bridging the semantic gap between natural language reasoning and rigid tool interfaces. Under the language-centered paradigm, LLM reasoning results typically exist in abstract natural language forms (e.g., ``count the number of vessels in the port"), whereas existing remote sensing tools (such as object detection models, edge extraction algorithms, and change detection networks) rely on specific code invocation formats and parameter configurations. Enabling the system to accurately comprehend task intent and retrieve the most appropriate algorithm model from a massive, heterogeneous tool library is the core difficulty of tool decision-making. Furthermore, complex tasks often require a chained combination of multiple tools (e.g., performing ``cloud removal" prior to ``land-cover classification"), so the system must be able to plan a ``chain of tools." Therefore, designing a decision mechanism capable of precisely mapping unstructured language instructions to executable code or API calls is a key prerequisite for endowing the interpretation system with ``operational agency".

\paragraph{Action Execution} Upon determining ``which tool to use" or ``what observation to make," designing an efficient execution mechanism to cope with the dynamically changing physical environment constitutes another major challenge. Remote sensing observation environments are typically characterized by high dynamism and are subject to uncontrollable factors such as cloud cover, illumination changes, and target movement, which can render preset perception strategies ineffective during execution. For instance, in UAV inspection or dynamic satellite scheduling, the system cannot rely solely on static task planning for information acquisition but must possess the ability to respond to environmental feedback in real time. This requirement implies that the interpretation system must establish an ``input-feedback-correction" closed loop during action execution, dynamically adjusting sensor parameters (e.g., angle and resolution) or correcting flight paths based on the returned observation status. Consequently, constructing an adaptive execution mechanism that accommodates dynamic environmental constraints to ensure the robust acquisition of high-quality data serves as the physical foundation for realizing efficient reasoning and decision-making.

\subsection{Language-Centered Unified Multimodal Representation}
In language-centered remote sensing intelligent interpretation systems, the key to unified multimodal representation lies in the effective mapping of heterogeneous information from different modalities into the language space. This task requires not only the unification of data representation formats across modalities but also the accurate alignment of the semantics conveyed by each modality.

Therefore, the process of language-centered unified multimodal representation involves two core problems:

\begin{itemize}
\item{How to align the structural of different modalities?}
\item{How to align the semantics across modalities?}
\end{itemize}

\begin{figure*}[!t]
\centering
\includegraphics[width=0.9\textwidth]{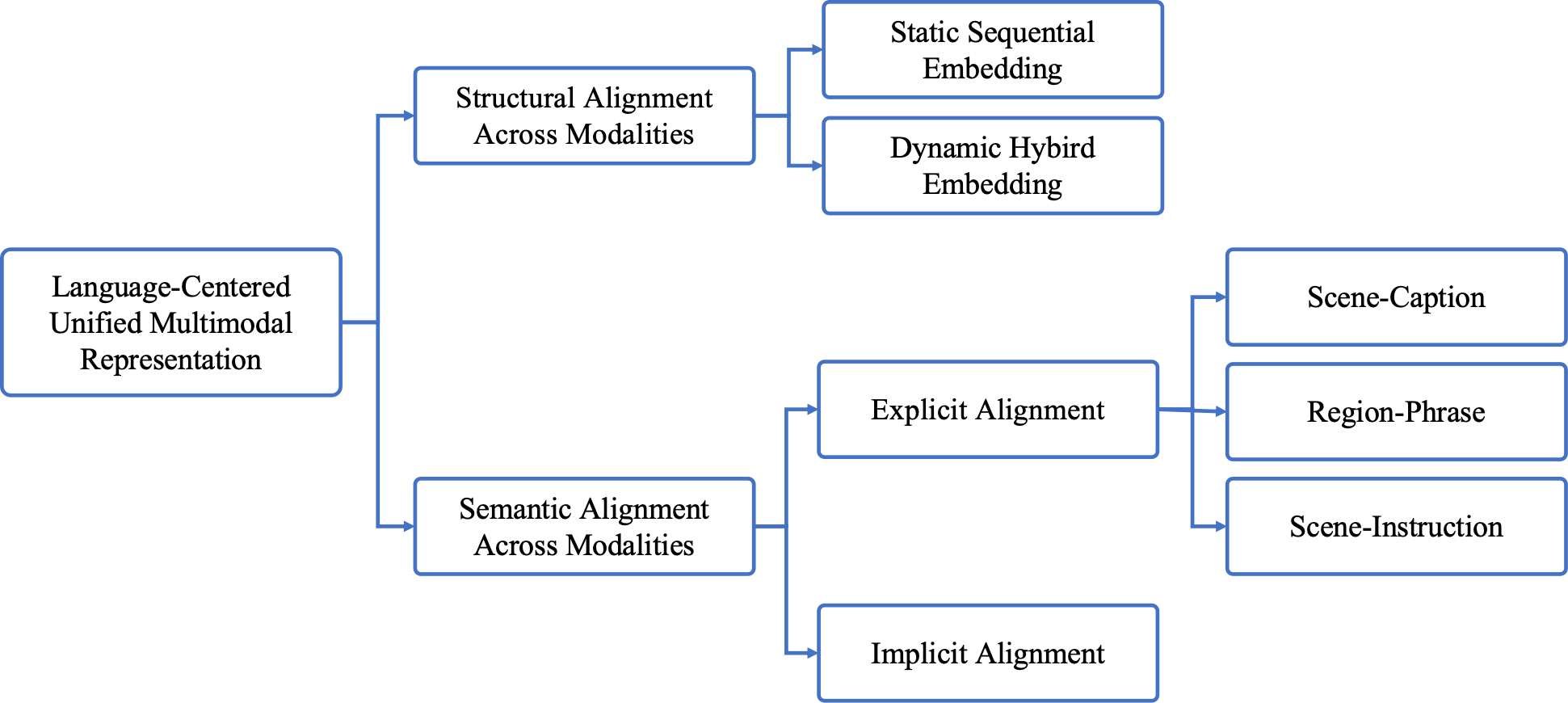}%
\caption{Structural organization of language-centered multimodal representation unification.}
\label{3_2content_structure}
\end{figure*}

The structure of this section is illustrated in Fig.\ref{3_2content_structure}, and representative methods corresponding to each component are summarized in Table \ref{tab:3_2}.

\begin{table*}[t]
\caption{Unified multimodal representation structure and representative methods.
\label{tab:3_2}}
\centering
\begin{tabular}{>{\raggedright\arraybackslash}p{3.5cm} 
                >{\raggedright\arraybackslash}p{4.2cm} 
                >{\raggedright\arraybackslash}p{7.5cm}}
\toprule
\multirow{1}{*}{Content structure} & Category & Representative algorithms \\

\midrule

\multirow{2}{*}{Structural Alignment} & Static Sequential Embedding & {VisualBERT\cite{li2019visualbert}, LXMERT\cite{tan2019lxmert}, UNITER\cite{chen2020uniter}, RSGPT\cite{hu2025rsgpt}, GeoChat\cite{kuckreja2024geochat}, RS-LLaVA\cite{bazi2024rs}} \\
\cmidrule(l){2-3}
                    & Dynamic Hybrid Embedding & {Oscar\cite{li2020oscar}, VIMA\cite{jiang2022vima}, PaLM-E\cite{driess2023palm}, TEOChat\cite{irvin2024teochat}, SkySenseGPT\cite{luo2024skysensegpt}} \\

\midrule

\multirow{7}{*}{Semantic Alignment} & Scene-Caption & CLIP\cite{radford2021learning}, UniCL\cite{li2024unicl}, K-LITE\cite{shen2022k}, LaCLIP\cite{fan2023improving}, SimVLM\cite{wang2021simvlm}, BLIP\cite{li2022blip}, Qu et al.\cite{qu2016deep}, RSICD\cite{lu2017exploring},  RSITMD\cite{yuan2022exploring}, RemoteCLIP\cite{liu2024remoteclip},  GeoCLIP\cite{vivanco2023geoclip} \\
                   \cmidrule(l){2-3}
                   &Region-Phrase & ViLBert\cite{lu2019vilbert}, RoIAlign\cite{he2017mask}, VL-BERT\cite{su2019vl}, ViLT\cite{kim2021vilt}, FGVP\cite{yang2023fine}, Alpha-CLIP\cite{sun2024alpha}, ASM\cite{wang2023all}, Zou and Shi\cite{shi2017can}, GeoVG\cite{sun2022visual}, RSVG\cite{zhan2023rsvg}, TEMO\cite{lu2023few} \\
                   \cmidrule(l){2-3}
                   &Scene-Instruction & LLaVA\cite{liu2023visual}, InstructBLIP\cite{dai2023instructblipgeneralpurposevisionlanguagemodels}, CAT\cite{wang2023caption}, SEAL\cite{wu2024v}, RSVQA\cite{lobry2020rsvqa}, VQA-TextRS\cite{al2022open}, Prompt-RSVQA\cite{chappuis2022prompt} \\
                   \cmidrule(l){2-3}
                   & Implicit alignment & Maniparambil et al.\cite{maniparambil2024vision}, LM4VisualEncoding\cite{pang2023frozen}, LAMP\cite{adeniji2023language}, GRAFT\cite{mall2023remote} \\
\bottomrule
\end{tabular}
\end{table*}

\subsubsection{Structural Alignment across Modalities}
Data from different modalities often exhibit significant differences in spatial resolution and temporal coverage. For example, the spatial resolution of remote sensing imagery may vary depending on the sensor type (e.g., RGB, MSI, or SAR), while temporal coverage is influenced by the frequency of data acquisition. Traditional approaches typically represent each modality in distinct formats, making cross-modal fusion difficult and lacking effective channels for information interaction. However, with the success of transformers\cite{vaswani2017attention} in natural language processing, an increasing number of studies have demonstrated that it is possible to achieve unified representation of multimodal data by constructing modality-specific embeddings. In this approach, the transformer processes data from different modalities at the embedding layer, effectively mapping them---despite spatial and temporal differences---into a shared semantic space, thus enabling seamless fusion and collaboration across modalities. This strategy can be categorized into two main types depending on how the embeddings from different modalities are combined: static sequential embedding and dynamic hybrid embedding.

\begin{figure*}[!t]
\centering
\includegraphics[width=0.9\textwidth]{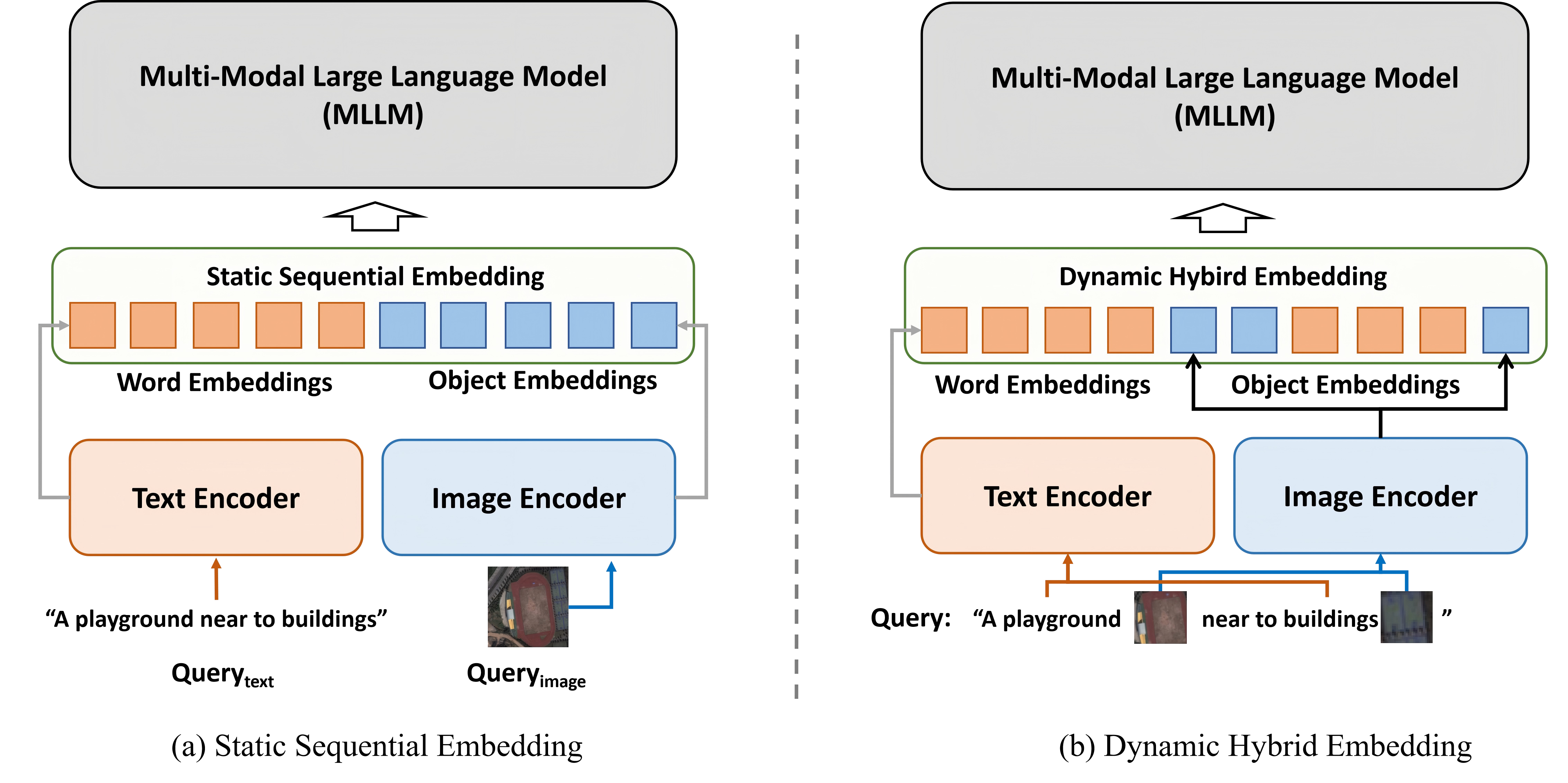}%
\caption{Illustration of unified multimodal representation embedding methods.}
\label{structural}
\end{figure*}

\paragraph{Static Sequential Embedding} As illustrated in Fig.\ref{structural}(a), the static sequential embedding approach combines embeddings from different modalities in a fixed order and length after passing each modality through its respective encoder. For example, in VisualBERT\cite{li2019visualbert}, the first N embeddings represent textual information, while the following K embeddings represent visual information separated by a [SEP] token. This approach allows the model to process and interpret the fused multimodal data jointly. Building upon this approach, LXMERT\cite{tan2019lxmert} enhances the model's capacity to extract both intramodal and intermodal features by incorporating self-attention and cross-attention mechanisms. UNITER\cite{chen2020uniter} further improves multimodal interaction learning by combining four training objectives: masked language modeling (MLM), masked region modeling (MRM), word-region alignment (WRA), and image-text matching (ITM). 

In the field of remote sensing, RSGPT\cite{hu2025rsgpt} integrates visual and language modalities using a Q-Former to fuse modality-specific embeddings. It introduces image-question-answer triplets based on remote sensing imagery interpretation as domain-specific knowledge and proposes RSIEval as a benchmark for evaluating the performance of remote sensing multimodal large language models (RSMLLMs). Building on this work, GeoChat\cite{kuckreja2024geochat} fine-tunes a pretrained LLaMA\cite{touvron2023llama} model using four types of visual-language understanding tasks. It distinguishes visual and language modalities using \texttt{\textless im\_start\textgreater} and \texttt{\textless im\_end\textgreater} tokens, significantly improving the model's understanding of remote sensing imagery. Similarly, RS-LLaVA\cite{bazi2024rs} adopts the multimodal architecture of LLaVA\cite{liu2023visual}, introducing a projection layer to map visual modalities. Remote sensing images and their descriptions are encoded as image tokens and text tokens, respectively, and the model is further optimized using LoRA\cite{hu2022lora} fine-tuning techniques.

\paragraph{Dynamic Hybrid Embedding} As shown in Fig.\ref{structural}(b), dynamic hybrid embedding refers to the approach in which data from different modalities are concatenated and directly input into the embedding layer for joint processing. Unlike the modality-tagging method, in dynamic hybrid embedding, the number and positions of the embeddings from each modality are dynamically variable. This dynamic approach allows the model to effectively capture interaction patterns across modalities, thereby enhancing its understanding of multimodal joint representations. For example, OSCAR\cite{li2020oscar} directly combines textual data, object labels, and object images into a single input, which is uniformly processed through the embedding layer. This approach significantly improves the model's ability to extract target information in query-based tasks. PaLM-E\cite{driess2023palm} builds on this work by integrating continuous data streams from the physical world, enabling the model to interact with real-world environments. VIMA\cite{jiang2022vima} focuses on the domain of embodied intelligence and directly combines text tokens, object tokens, and action tokens, allowing the model to learn how to generate appropriate action feedback in response to different instructions and goals, thus enabling robots to perform a wide variety of tasks.

In the field of remote sensing, TEOChat\cite{irvin2024teochat} employs a shared vision-language connector to achieve hybrid embedding of image tokens and text tokens, thereby further enhancing the fusion of multimodal information. To improve the model's understanding and reasoning ability with respect to geographic knowledge, SkySenseGPT\cite{luo2024skysensegpt} incorporates more detailed instructions into text tokens and introduces both region-level and image-level scene graph information. These features significantly strengthen the model's ability to understand both local and global relationships among geo-objects. Through this approach, remote sensing models are better equipped to capture spatial information and object relationships across different scales and levels, thereby improving their overall ability to intelligently interpret remote sensing imagery.

The shift from modality-tagging methods to joint embedding strategies has become a major development trend for improving the performance of language-centered multimodal models. The core idea is to increase the model's learning difficulty by disrupting the order of tokens from different modalities, thereby encouraging a deeper understanding of multimodal data. Modality-tagging methods assign fixed positions and lengths to different modalities, allowing the model to clearly distinguish the source of each token. However, this structured design can limit the model's ability to capture complex relationships across modalities. In contrast, joint embedding mixes and concatenates tokens from different modalities in a less constrained manner, compelling the model to learn intermodal interactions and complementary information within a more flexible and intricate input structure. As a result, the model must not only understand the individual characteristics of each modality but also learn how to integrate latent relationships across modalities, thus enhancing its overall ability to interpret multimodal data holistically.

\subsubsection{Semantic Alignment across Modalities}
Aligning representations from different modalities at the semantic level is a fundamental challenge in multimodal alignment research. Existing approaches to semantic alignment can be broadly categorized into two types: explicit alignment and implicit alignment. The key difference between them lies in whether the alignment relationships across modalities are explicitly introduced during training, such as through additional manual annotations or predefined alignment schemes.

\begin{figure*}[!t]
\centering
\includegraphics[width=0.9\textwidth]{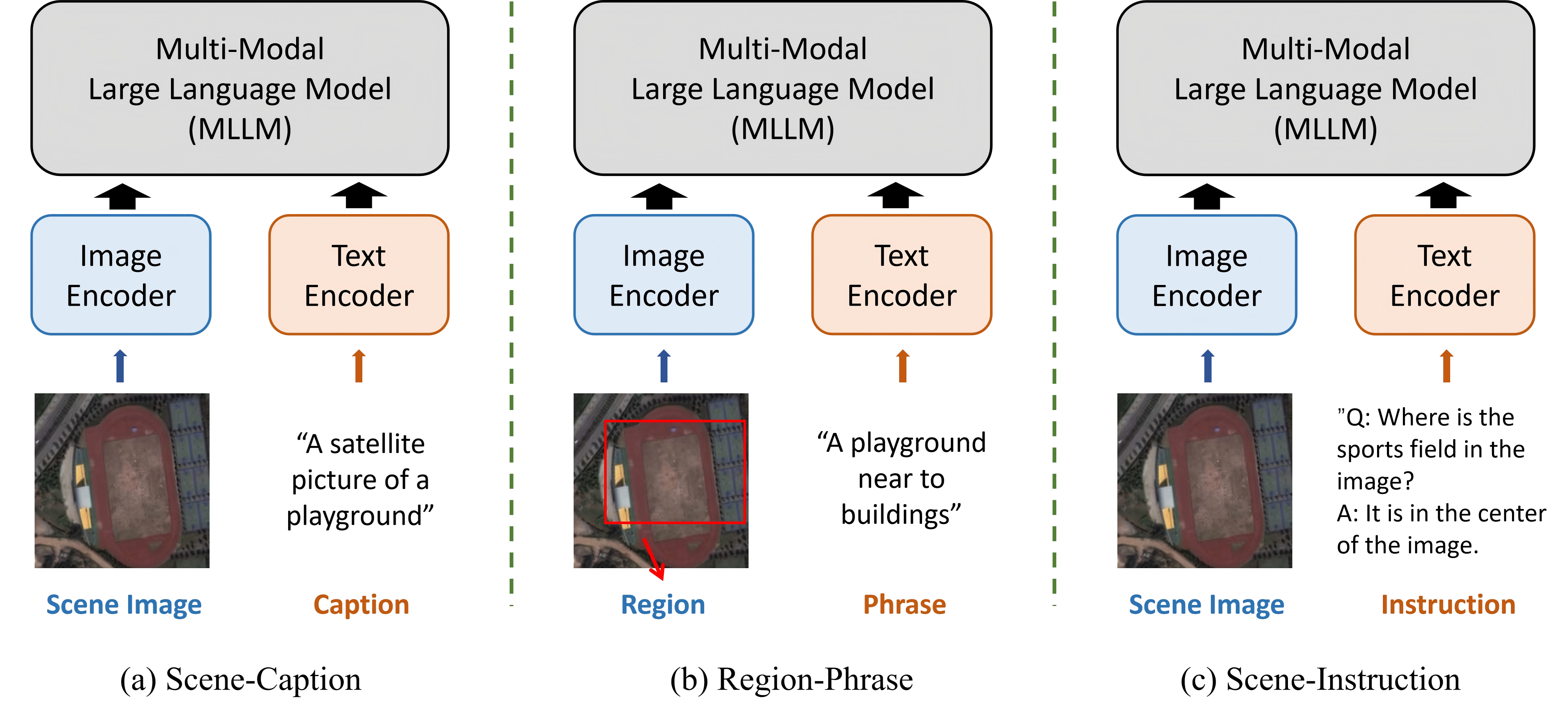}%
\caption{Illustration of explicit alignment methods.}
\label{semantic}
\end{figure*}

\paragraph{Explicit Alignment} To effectively embed different modalities into the language modality, a straightforward approach is to use paired data--such as image-text or point cloud-text datasets--to train the model. Within this training framework, the model learns to align modalities by modeling the pairing relationships between multimodal data and corresponding textual descriptions. Based on the granularity of alignment between nonlinguistic and linguistic modalities, current research efforts can be broadly classified into three main approaches: scene-caption alignment, region-phrase alignment, and scene-instruction alignment. The characteristics of each approach are illustrated in Fig.\ref{semantic}.

\begin{itemize}
\item{\textbf{Scene-Caption Alignment}: Scene-caption alignment methods focus on aligning an entire image with its corresponding textual description, aiming to help the model learn the overall association between visual content and language, as illustrated in Fig.\ref{semantic}(a). A representative work in this category is CLIP\cite{radford2021learning}, which leverages contrastive learning to achieve effective alignment and has inspired a series of follow-up methods, including UniCL\cite{li2024unicl}, K-LITE\cite{shen2022k}, and LaCLIP\cite{fan2023improving}. Owing to the task-agnostic nature of contrastive learning, these approaches are well suited as foundational models that can be quickly adapted to various downstream tasks via fine-tuning, few-shot, or even zero-shot learning. However, these methods face several limitations. First, they rely heavily on large-scale, high-quality image-text paired datasets, and their image understanding ability tends to be shallow. Second, such methods often struggle to handle both image understanding and text generation tasks simultaneously. 

\vspace{0.1cm}

To address these issues, SimVLM\cite{wang2021simvlm} introduced the PrefixLM pretraining objective, which maintains strong image understanding while enabling text generation capabilities. Similarly, BLIP\cite{li2022blip} performs well in terms of both image understanding and captioning, and its unique CapFilt mechanism effectively reduces noise in the training data. Despite these improvements, scene-caption-based methods generally lack the ability to deeply and finely understand image content. This limitation largely stems from the fact that simple image-text pairs fail to provide sufficiently rich guidance, thereby constraining the model's capacity to learn more complex vision-language relationships. 

\vspace{0.1cm}

\textbf{In remote sensing}, vision-language models have rapidly become a research focus because of their superior capabilities in image understanding and language processing. One of the early efforts in scene-caption alignment for remote sensing was conducted by Qu et al.\cite{qu2016deep}, who utilized the UCM-Captions dataset\cite{qu2016deep}. However, this dataset was relatively small in scale and contained simple descriptions, limiting its ability to represent complex content and diverse scenarios in remote sensing imagery. To address these limitations, subsequent works such as RSICD\cite{lu2017exploring} and RSITMD\cite{yuan2022exploring} constructed larger, more diverse, and semantically richer datasets while also developing task-specific models for various downstream applications. Building upon these foundations, RemoteCLIP\cite{liu2024remoteclip} and GeoCLIP\cite{vivanco2023geoclip} have further advanced the field. These studies introduced large-scale scene-caption datasets for remote sensing containing up to 1.6 million and 5 million images, respectively, and proposed two foundational models tailored specifically to the remote sensing domain. Compared with earlier efforts, these studies adopt transformer-based architectures and demonstrate a deeper understanding of remote sensing image content.}

\vspace{0.1cm}

\item{\textbf{Region-Phrase Alignment}: To enhance the model's fine-grained understanding of images, some studies have explored incorporating detailed correspondences between image regions and individual words or phrases, as illustrated in Fig.\ref{semantic}(b). Early work in this research direction--such as ViLBERT\cite{lu2019vilbert}-typically relied on external object detectors to identify regions of interest (ROIs), using techniques such as RoIAlign\cite{he2017mask} to extract features from these regions and align them with corresponding text tokens. However, this approach has several notable limitations. First, its dependence on pretrained external detectors restricts the range of detectable object categories and directly affects model performance. Second, the inclusion of an additional detection module increases model complexity and reduces training efficiency. 

\vspace{0.1cm}

To address these issues, VL-BERT\cite{su2019vl} integrates the detection module into the training process, while ViLT\cite{kim2021vilt} eliminates it altogether, instead using a pure transformer architecture to extract image features. These adjustments significantly improve efficiency while maintaining strong performance. Additionally, some researchers argue that relying solely on ROI features may lead to the loss of valuable background contextual information. In response, strategies such as background blurring, masking, and feature fusion have been proposed to preserve as much background information as possible while extracting regional features.

\vspace{0.1cm}

\textbf{In remote sensing}, region-phrase alignment has emerged as a compelling research direction, particularly within the context of scene-caption tasks. Early studies primarily adopted traditional encoder-decoder architectures and focused on extracting features from target regions for image caption generation. While these approaches proved effective, they often overlooked domain-specific knowledge unique to remote sensing. With the rapid advancement of vision-language models, concept-level region-phrase alignment has seen new breakthroughs in remote sensing. For instance, GeoVG\cite{sun2022visual} was the first to introduce the visual grounding task into the remote sensing domain. It employs a dual-stream architecture to separately extract visual features and geographic relational graphs, which are then integrated through a fusion module to effectively combine geospatial relationships with visual information. Building upon this work, RSVG\cite{zhan2023rsvg} introduced the MGVLF module, which enhances grounding accuracy by filtering out noise and reinforcing salient features. Additionally, few-shot object detection (FSOD) has become an emerging topic in remote sensing. A recent work, TEMO\cite{lu2023few}, has made innovative contributions in this direction by incorporating detailed descriptions of all object classes as additional features, effectively mitigating class confusion.}

\vspace{0.1cm}

\item{\textbf{Scene-Instruction Alignment}: While the previously discussed scene-caption and region-phrase alignment methods effectively achieve direct alignment between visual and language modalities, they often fall short in exploring the complex and abstract relationships among concepts expressed across modalities--an aspect that is crucial for achieving higher-level compositional generalization. Against this backdrop, scene-instruction-based alignment demonstrates its unique value. Its core lies in answering questions through logical reasoning, as illustrated in Fig.\ref{semantic}(c). This approach requires the model not only to understand individual concepts but also to grasp the intricate relationships between them. 

\vspace{0.1cm}

In recent years, LLMs have demonstrated powerful general knowledge understanding and reasoning capabilities, which has driven a new research direction that leverages LLMs to address multimodal tasks. By fine-tuning LLMs with scene-instruction datasets, as demonstrated in LLaVA\cite{liu2023visual} and InstructBLIP\cite{dai2023instructblipgeneralpurposevisionlanguagemodels}, models can further enhance their logical reasoning and instruction-following abilities. These works have also simplified the process of constructing high-quality instruction datasets. Moreover, recent studies have introduced novel instruction processing mechanisms; for example, CAT\cite{wang2023caption} expands narrow language instructions into vision-language instructions, while SEAL\cite{wu2024v} incorporates a visual search mechanism to help the model focus more precisely on key regions within an image.

\vspace{0.1cm}

\textbf{In remote sensing}, scene-instruction alignment is gradually emerging as a critical technique. It enables nonexpert users to interact with remote sensing imagery using natural language. In RSVQA\cite{lobry2020rsvqa}, researchers used OpenStreetMap data and manual annotation to create the first large-scale remote sensing visual question answering (VQA) dataset and proposed a CNN-RNN-based VQA model. Subsequent work advanced the field by developing VQA-TextRS\cite{al2022open}, a dataset supporting open-ended question answering, alongside a Transformer-based VQA model. In the same year, Prompt-RSVQA\cite{chappuis2022prompt} proposed a method for mapping images into the language space and feeding them, along with questions, into BERT\cite{koroteev2021bert} to generate answers. These studies indicate that remote sensing research is rapidly converging in the direction of vision-language models, laying the groundwork for future innovations in human-AI interactions and multimodal scene understanding.}

\end{itemize}

In summary, the progression from scene-caption to region-phrase and, more recently, to scene-instruction alignment reflects a gradual deepening of the semantic understanding of geo-objects within remote sensing intelligent interpretation. This evolution corresponds to a shift from coarse to fine grained image-text matching strategies. Specifically, scene-caption alignment focuses on matching entire images with holistic textual descriptions; region-phrase alignment refines the image side by dividing it into multiple regions and aligning each with a corresponding phrase; scene-instruction alignment further enhances the granularity of image-text alignment by using complex instructions to describe the relationships among geo-objects. The development of image-text alignment methods from coarse to fine granularity has accelerated the evolution of remote sensing interpretation models from shallow semantic understanding toward deep mechanistic comprehension.

\vspace{0.1cm}

\paragraph{Implicit Alignment} While explicit alignment methods have achieved remarkable success in multimodal learning, they also present several practical limitations. First, they rely on large-scale paired image-text datasets for model training, which poses significant challenges in terms of data collection and preprocessing. Second, achieving alignment between two different modalities typically requires separate training procedures for each modality, increasing both computational and time costs. More importantly, as the number of modalities increases, even when a language-centered alignment strategy is used, the alignment process becomes exceedingly complex and cumbersome. Moreover, when interaction is required between two nonlanguage modalities, using language as an intermediary can lead to issues in both efficiency and accuracy. If the alignment between one modality and language is imperfect, errors may accumulate during the conceptual transfer process, ultimately leading to suboptimal or incorrect interaction outcomes.

Despite the significant heterogeneity across different modalities, they ultimately describe the same physical world. This fundamental fact implies that, regardless of whether the alignment between visual and language modalities is explicitly defined, an intrinsic condition for unified convergence should exist. In support of this condition, Minyoung Huh et al.\cite{huh2024position} experimentally demonstrated that visual models with different architectures and training objectives exhibit a high degree of representational consistency and that this consistency continues to improve as the model size and training data scale increase. This pattern also holds for cross-modal vision-language models. Similarly, Maniparambil\cite{maniparambil2024vision} reported that it is possible to perform subtitle matching or retrieval tasks using unaligned visual and language encoders, thereby further validating the existence of an inherent consistency between visual and language representation spaces. These findings raise a fundamental question: Is there an implicit, inherently defined alignment between language and vision? If so, then it may be possible to achieve concept-level alignment between vision and language even when training on single-modal data---this is the central idea behind implicit alignment, which has recently garnered increasing research interest. 

\vspace{0.1cm}

Notably, many of these studies choose to align visual representations to the language space rather than the other way around. This observation leads to a second key question: Why is language often chosen as the convergence point for vision-language alignment? This paper argues that, as the primary medium of human communication, language most accurately reflects human information cognition and logical reasoning mechanisms. Furthermore, language is inherently abstract, generalizable, and adaptable, making it well suited as a global workspace for efficiently organizing a conceptual space that encompasses visual and other nonlinguistic modalities. Building on this view, this paper refers to such research efforts as \emph{language-centered implicit multimodal alignment}.

Language-centered implicit multimodal alignment aims to explore how nonlinguistic modality information can be directly aligned with the language representation space in the absence of prior modality-language alignment pairs. In this line of research, the work LM4VisualEncoding\cite{pang2023frozen} revealed that even when trained solely on textual data, the encoder of a large language model demonstrates surprisingly strong performance on both pure visual tasks and multimodal tasks. This finding offers new insights into the use of language models for visual tasks, highlighting their ability to encode and process visual-language information through shared parameters.
Furthermore, a recent LAMP\cite{adeniji2023language} study confirms the potential of pretrained language models in handling 3D point cloud data, providing additional evidence of the broad adaptability of language models in interpreting and processing nontraditional textual data. Collectively, these research efforts have pushed the boundaries of applying large language models to computer vision tasks and have expanded our understanding of their latent capabilities in processing unimodal visual data.

\begin{figure*}[!t]
\centering
\includegraphics[width=0.9\textwidth]{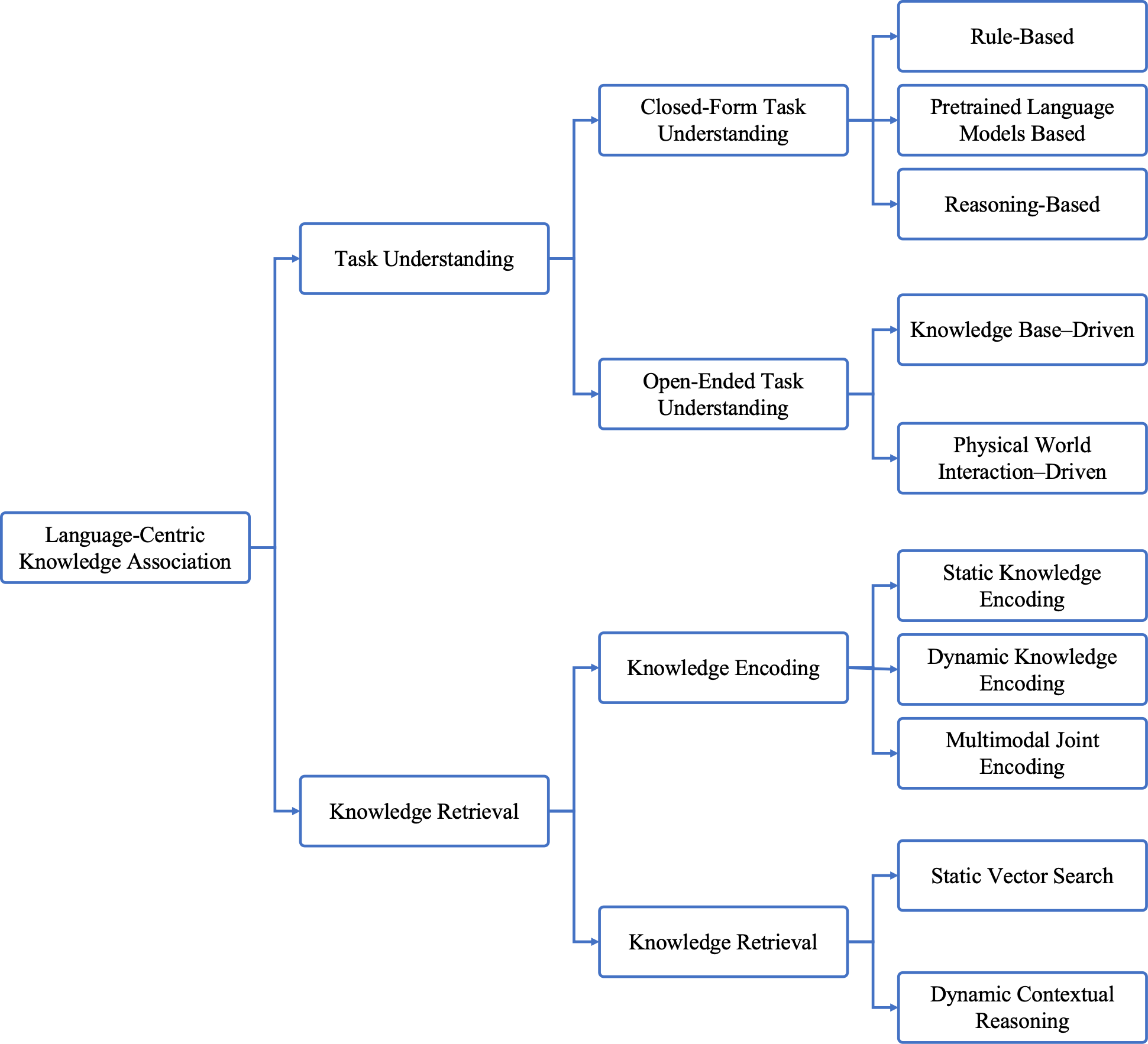}%
\caption{Structural organization of language-centered knowledge association.}
\label{3_3Knowledge}
\end{figure*}

With the rapid advancement of remote sensing and related technologies, vast amounts of remote sensing data are being generated daily. However, constructing large-scale image-text paired datasets in the remote sensing domain remains a significant challenge. Therefore, introducing implicit alignment into remote sensing is both necessary and timely. Recently, several studies have begun to explore this direction. For example, GRAFT\cite{mall2023remote} leverages internet images captured from the same ground locations as remote sensing images to serve as an intermediary between remote sensing imagery and language. It proposes a method to train a remote sensing vision-language model without relying on any textual annotations. Such attempts to introduce implicit alignment into the remote sensing field provide promising new directions for addressing the shortage of large-scale multimodal image-text pairs. Nonetheless, this line of research is still in its early stages and requires further exploration.

\subsection{Language-Centered Knowledge Association and Reasoning}
Alignment enables the convergence of information from different modalities into a shared language embedding space, whereas the core of knowledge association lies in establishing associations between perceptual information and internal/external knowledge based on the interpretation task requirements. Accordingly, the knowledge association process primarily involves two key challenges: 

\begin{itemize}
\item{How to understand the interpretation task?}
\item{How to identify the relevant knowledge?}
\end{itemize}

The structure of this section is illustrated in Fig.\ref{3_3Knowledge}, and representative methods for each component are summarized in Table \ref{tab:3_3}.

\begin{table*}[t]
\caption{Language-centered knowledge association structure and representative methods.
\label{tab:3_3}}
\centering
\begin{tabular}{>{\raggedright\arraybackslash}p{3.5cm} 
                >{\raggedright\arraybackslash}p{4.2cm} 
                >{\raggedright\arraybackslash}p{7.5cm}}
\toprule
\multirow{1}{*}{Content structure} & Category & Representative algorithms \\

\midrule

\multirow{5}{*}{Task Understanding} 
                    & Rule-based & {APLICOT\cite{mizoguchi1983prolog}, ACT-R\cite{anderson1997act}, RS-DFM\cite{wang2024rs}, Feng et al.\cite{feng2023self}, Semantic-CD\cite{zhu2025semantic}} \\
\cmidrule(l){2-3}
                    & Pretrained Language Models based & {Llama3\cite{grattafiori2024llama}, GPT-3\cite{brown2020language}, DeepSeek-V3\cite{liu2024deepseek}, LiT-4-RSVQA\cite{hackel2023lit}, GeoGPT\cite{zhang2023geogpt}, GeoChat\cite{kuckreja2024geochat}, SkyEyeGPT\cite{zhan2025skyeyegpt}, EarthGPT\cite{zhang2024earthgpt}} \\
\cmidrule(l){2-3}
                    & Reasoning-based & {CoT\cite{wei2022chain}, Self-Reflection\cite{renze2024self}, Self-Refine\cite{madaan2023self}, OpenAI-O1\cite{jaech2024openai}, DeepSeek-R1\cite{guo2025deepseek}, Remote Sensing ChatGPT\cite{guo2024remote}, CEOS\cite{wagner2024generative}} \\
\cmidrule(l){2-3}
                    & Knowledge Base Driven & {Auto-CoT\cite{zhang2022automatic}, MemoryBank\cite{zhong2024memorybank}} \\
\cmidrule(l){2-3}
                    & Physical World Interaction Driven & {ReAct\cite{yao2023react}, ReSpAct\cite{dongre2024respact}, RS-Agent\cite{xu2024rs}} \\

\midrule

\multirow{5}{*}{Knowledge Retrieval} 
                    & Static Knowledge Encoding & {Word2Vec\cite{church2017word2vec}, GloVe\cite{pennington2014glove}, FastText\cite{joulin2016fasttext}} \\
\cmidrule(l){2-3}
                    & Dynamic Knowledge Encoding & {ELMo\cite{sarzynska2021detecting}, BERT\cite{koroteev2021bert}, GPT\cite{radford2018improving}} \\
\cmidrule(l){2-3}
                    & Multimodal Joint Encoding & {MKBE\cite{pezeshkpour2018embedding}, MULE\cite{kim2020mule}, E5-V\cite{jiang2024e5}, EarthMarker\cite{zhang2024earthmarker}} \\
\cmidrule(l){2-3}
                    & Static Vector Search & {TF-IDF\cite{aizawa2003information}, k-NN\cite{cover1967nearest}, ANN\cite{arya1998optimal}} \\
\cmidrule(l){2-3}
                    & Dynamic Contextual Reasoning & {Self-attention\cite{shaw2018self}, Cross-attention\cite{hou2019cross}, Block-Attention\cite{sun2024block}, KTIR\cite{mi2024knowledge}} \\
\bottomrule
\end{tabular}
\end{table*}

\vspace{0.1cm}

\subsubsection{Task Understanding} The depth of task understanding in LLMs is reflected in their ability to decompose complex tasks. For example, a ship motion monitoring task requires an interpretation system to detect and recognize all ships in a specified maritime region, identify their fine-grained categories, and determine dynamic changes over time through temporal analysis. Owing to the highly composite nature of this task, a single model would struggle to handle it effectively. Instead, an LLM is capable of decomposing it into four relatively independent yet sequentially connected subtasks, namely, detection, coarse classification, fine-grained recognition, and temporal analysis, each of which can be delegated to a specialized expert model. In this process, the LLM effectively restructures the complex task of ``ship motion monitoring" into a series of simpler subtasks, each with a clear knowledge objective and a corresponding expert model. This operation demonstrates the LLM's comprehension of the task and reduces the overall task complexity, thereby facilitating more efficient task execution. From this perspective, based on whether the task environment, including the knowledge space and the external environment, changes during task understanding, we categorize task understanding into two types: closed-form task understanding and open-ended task understanding.

\paragraph{Closed-Form Task Understanding} In closed-form task understanding scenarios, both the knowledge space and the environment space are static and fixed. Task understanding is achieved primarily through existing knowledge, without involving interaction with the external environment. There are three main approaches to task decomposition in this setting: rule-based task decomposition, pretrained language model-based task decomposition, and chain-of-thought-based task understanding.

\begin{figure*}[!t]
\centering
\includegraphics[width=0.9\textwidth]{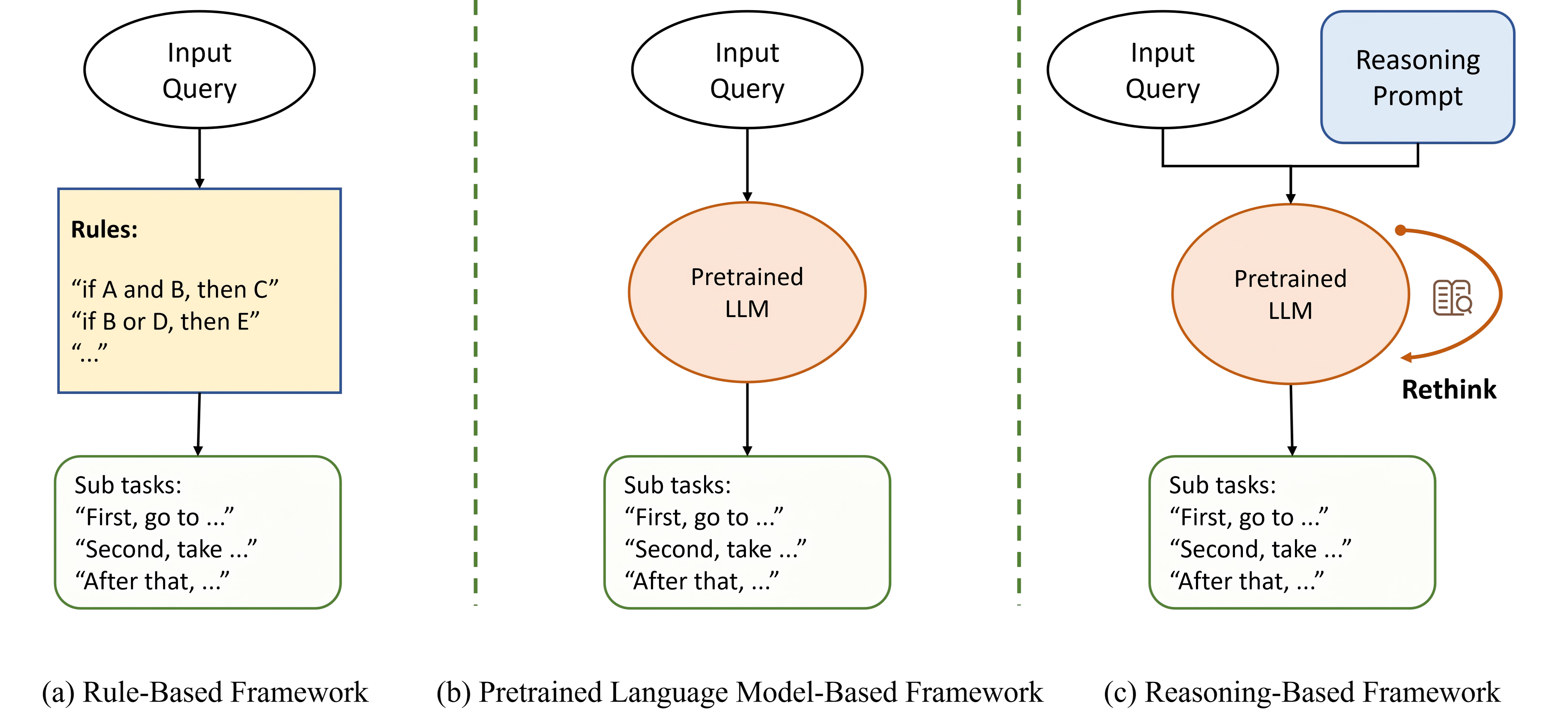}%
\caption{Illustration of closed-form task understanding methods.}
\label{close_forme}
\end{figure*}

\begin{itemize}

\item{\textbf{Rule-Based Task Decomposition}: Rule-based task decomposition relies on predefined rules and fixed logical procedures to break down and execute tasks, as illustrated in Fig.\ref{close_forme}(a). These rules are typically handcrafted for specific task requirements. Once the model receives a task input, it follows the established rules step by step to reason and complete the task. This approach offers strong interpretability, as each reasoning step and decision-making rule is explicit and fixed, allowing the process by which the model derives its output from the input to be clearly traced, understood, and verified. In the field of natural language processing, early rule-based systems such as APLICOT\cite{mizoguchi1983prolog} (based on Prolog) used logical rules (e.g., A :- B, C, meaning ``IF B AND C, THEN A") to connect multiple rule units and perform reasoning in complex tasks. However, this approach suffered from limited context understanding, as it could only build logical relationships based on local information. To address this shortcoming, ACT-R\cite{anderson1997act} extended APLICOT\cite{mizoguchi1983prolog} by introducing a cognitive framework that combines fixed logic rules with memory units, enabling higher-level cognitive processing and improving flexibility and efficiency in task decomposition. Although rule-based methods can effectively support task understanding and decomposition under controllable logical chains, their reliance on explicitly defined rules has become a significant limitation in the era of big data. When dealing with complex tasks, the rigidity of rules often leads to poor performance. As a result, with the rise of language model-based methods, rule-based decomposition has been gradually phased out, particularly because of language models' superior flexibility and adaptability in handling diverse task scenarios.

\vspace{0.1cm}

In the field of remote sensing, complex tasks often rely on expert prior knowledge, with predefined rules used to decompose tasks into subcomponents, which are then processed step by step in conjunction with interpretation models. For example, in the area of target recognition, RS-DFM\cite{wang2024rs} addresses 3D object detection under multiplatform observation conditions by decomposing the task into four fixed subtasks to improve detection accuracy and robustness: feature extraction, BEV (bird's-eye view) generation, high- and low-frequency feature decoupling and calibration, and feature fusion. In multimodal interpretation, Feng et al.\cite{feng2023self} (2023) break down cross-modal interpretation into two subtasks: multiscale spatial feature extraction and multimodal feature fusion. In change detection, semantic-CD\cite{zhu2025semantic} decomposes the task into semantic segmentation and change feature extraction. It uses an adapted vision encoder to extract semantic features from bitemporal images, followed by a semantic change detection decoder to identify areas of semantic variation. These methods demonstrate that rule-based task decomposition plays an important role in remote sensing interpretation, as it not only simplifies complex tasks but also enhances accuracy and operational stability. However, such approaches are heavily dependent on expert-designed priors, with fixed task decomposition structures, making them poorly suited for open and dynamic environments.
}

\vspace{0.1cm}

\item{\textbf{Pretrained Language Model-based Task Decomposition}: As illustrated in Fig.\ref{close_forme}(b), task decomposition based on pretrained language models leverages large-scale models such as Llama3\cite{grattafiori2024llama}, GPT-3\cite{brown2020language}, and DeepSeek-V3\cite{liu2024deepseek}. These models are trained on massive text corpora and can understand and flexibly handle a wide range of tasks. Task decomposition in this paradigm is typically performed by interpreting task descriptions or instructions provided as input and generating output based on the model's pretrained knowledge and contextual understanding. The primary advantage of this approach lies in its flexibility: pretrained language models can handle a wide variety of tasks and adapt to diverse input types without the need for manually designed rules or constraints. Owing to their strong language understanding capabilities, these models can provide reasonable solutions across different scenarios and show superior adaptability to changing task requirements. However, this flexibility comes at a cost--particularly in terms of interpretability. Since the decision-making process of these models is based on internal parameters and massive training data, the reasoning steps are often opaque and difficult to verify externally. As a result, the outputs may be nontraceable and lack guaranteed correctness.

\vspace{0.1cm}

In the remote sensing domain, as new tasks continue to emerge---such as remote sensing visual question answering (RSVQA), remote sensing visual captioning (RSVC), and remote sensing visual grounding (RSVG)---interpretation tasks are being increasingly expressed in natural language. Rule-based task decomposition methods struggle to adapt to this flexible task format, prompting researchers to explore how to more efficiently ``understand" remote sensing interpretation tasks. To this end, several studies have introduced LLMs to analyze task requirements. For example, LiT-4-RSVQA\cite{hackel2023lit} uses the text encoder of a VLM to interpret task instructions, while GeoGPT\cite{zhang2023geogpt} employs an LLM to analyze and decompose interpretation tasks. Moreover, works such as GeoChat\cite{kuckreja2024geochat}, SkyEyeGPT\cite{zhan2025skyeyegpt}, and EarthGPT\cite{zhang2024earthgpt} further integrate LLMs into interpretation systems to enhance their understanding and decomposition capabilities for complex tasks. While LLM-based task decomposition offers significantly greater flexibility, allowing the system to automatically adapt to different task demands and intelligently break down complex tasks, the lack of transparency in the reasoning process introduces uncertainty regarding the trustworthiness of the decomposition results. This issue limits the practical applicability of such approaches in remote sensing scenarios that require high reliability.
}

\vspace{0.1cm}

\item{\textbf{Reasoning-Based Task Decomposition}: Reasoning-based task decomposition methods combine the flexibility of language models with the interpretability of rule-based systems by guiding the model to perform step-by-step logical reasoning, as illustrated in Fig.\ref{close_forme}(c). In this approach, reasoning instructions are provided to the model in the form of prompts, alongside the input query. During the generation of subtasks, the model continuously rethinks and refines its outputs, effectively decomposing the task through a progressive reasoning process. This method encourages the model to self-check at each step, ensuring that task execution is both structured and logically sound. The earliest representative of this approach is the chain-of-thought (CoT)\cite{wei2022chain} prompting method, which uses simple cues like ``Let's think step by step" to guide the model in generating intermediate reasoning steps and clearly showcasing the decomposition process in the final result. However, the reasoning logic in CoT\cite{wei2022chain} can be difficult to control, leading to unstable outcomes. To address this issue, methods such as self-reflection\cite{renze2024self} and self-refinement\cite{madaan2023self} introduce reflection mechanisms, enabling the model to self-evaluate the logic and accuracy of its reasoning, thereby improving reliability. Nevertheless, these methods rely on manually crafted prompt templates, and their performance on more complex tasks remains limited. To overcome these challenges, OpenAI-O1\cite{jaech2024openai} and DeepSeek-R1\cite{guo2025deepseek} apply reinforcement learning to fine-tune the reasoning capabilities of language models.

\vspace{0.1cm}

In the remote sensing domain, similar approaches have been explored in remote sensing ChatGPT\cite{guo2024remote}, which interprets user instructions in natural language and combines them with domain-specific geoscientific knowledge to decompose complex interpretation tasks into subtasks. These subtasks are then handled sequentially by invoking corresponding expert models. This strategy not only enhances the automation of remote sensing interpretation but also reduces the dependence on domain experts. Furthermore, CEOS\cite{wagner2024generative}, which was developed based on the LangChain\cite{pandya2023automating} framework, serves as an Earth observation analysis platform capable of automatically executing tasks such as vegetation analysis.
}

\end{itemize}

In summary, while closed-task understanding methods offer certain advantages in terms of task decomposition, they also face inherent limitations. Owing to the static nature of the knowledge and environment spaces involved in these methods, models are heavily reliant on predefined rules or pretrained knowledge, which often fall short when confronted with dynamic environments or unseen task requirements. This limitation is particularly evident when dealing with tasks that fall outside the coverage of the model's internal knowledge base, where LLMs often struggle to deliver satisfactory results. Although rule-based and pretrained language model-based methods can effectively handle well-defined tasks and achieve a high level of automation, their adaptability and flexibility are constrained by the static nature of their knowledge and the lack of interpretability in their reasoning processes. Furthermore, while reasoning-based task decomposition methods attempt to improve reasoning robustness through chain-of-thought prompting and reflection mechanisms, they still encounter difficulties when facing knowledge gaps or complex task scenarios. As a result, closed-task understanding methods continue to face significant challenges in terms of performance and reliability when applied to dynamic and complex real-world tasks.

\paragraph{Open-Ended Task Understanding} In open-ended task understanding scenarios, both the knowledge space and the environment space can be dynamically adjusted according to actual task requirements. To overcome the inherent limitations of closed-task understanding, researchers have begun to explore the integration of external information into the task understanding process. Based on the source of external information, open-task understanding can be further divided into two main pathways: knowledge base-driven task understanding and physical world interaction-driven task understanding.

\begin{figure*}[!t]
\centering
\includegraphics[width=0.9\textwidth]{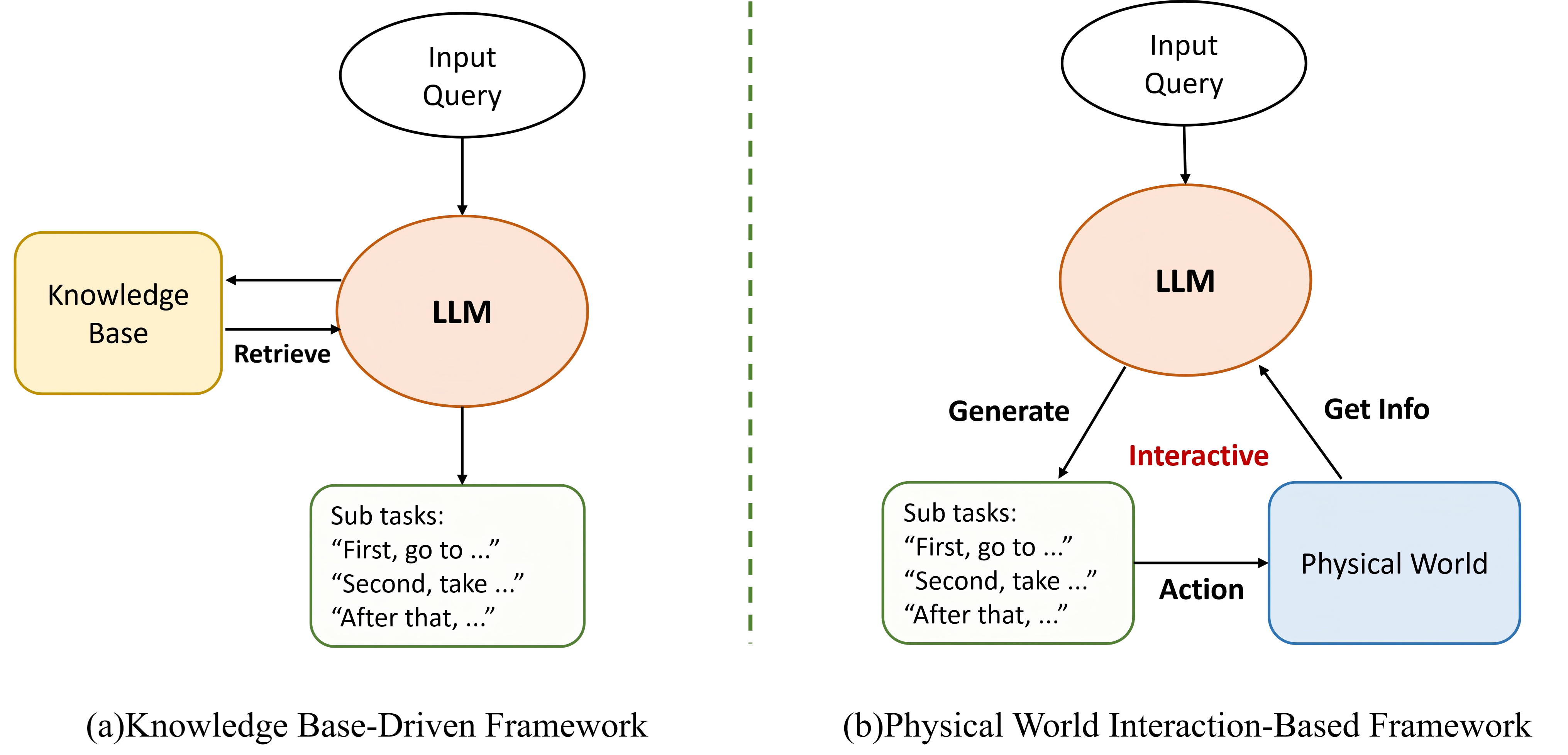}%
\caption{Illustration of Open-Ended Task Understanding Methods.}
\label{open_ended}
\end{figure*}

\begin{itemize}

    \item{\textbf{Knowledge Base-Driven Task Decomposition}: As illustrated in Fig. \ref{open_ended}(a), knowledge base-driven open-task understanding enhances a language model's comprehension by enabling interaction with external knowledge bases. The model retrieves information from these sources that is similar or relevant to the input query, thereby injecting additional knowledge into the task decomposition process. Representative works include Auto-CoT\cite{zhang2022automatic}, which retrieves task decomposition examples similar to the input instruction and guides the LLM to imitate these examples for more reasonable task breakdowns. Another example is MemoryBank\cite{zhong2024memorybank}, which leverages retrieval-augmented generation (RAG) to extract task-relevant knowledge from an external database and supports the decomposition process accordingly. Because external knowledge bases can be updated in real time, this method allows LLMs to continuously access the latest information and fill knowledge gaps as needed. However, the relevance and usefulness of the retrieved knowledge still depend heavily on the accuracy of the retrieval algorithm, which determines whether the acquired information truly aligns with the task requirements.}

\vspace{0.1cm}
    
    \item{\textbf{Physical World Interaction-Driven Task Decomposition}: As illustrated in Fig.\ref{open_ended}(b), physical world interaction-driven open-task understanding acquires the required information through feedback from interactions with the physical environment, enabling the task decomposition process to unfold progressively. The language model initially generates a set of subtasks based on the input query; as actions associated with these subtasks affect the physical world, feedback from the environment is used to refine and optimize the subsequent decomposition steps. A representative work in this direction is ReAct\cite{yao2023react}, which combines task decomposition with action execution. By directly interacting with the physical world, the model determines the next steps of task decomposition based on both environmental feedback and the results of previous task planning. ReSpAct\cite{dongre2024respact} extends this framework by introducing user interaction feedback, allowing the LLM to seek advice, clarify ambiguities, or understand user preferences through dialog, thereby updating its task decomposition plan in a more responsive manner. However, a key challenge in this line of research remains how to design efficient and reliable environmental interaction mechanisms.}
\end{itemize}

In the field of remote sensing, open-task understanding is still in its early stages of development. One of the most recent relevant studies is RS-Agent\cite{xu2024rs}. RS-Agent\cite{xu2024rs} builds upon a pretrained LLM and implements five specialized agents that are individually responsible for reasoning, decision-making, memory, knowledge-enhanced retrieval, and decision-enhanced retrieval. It also establishes integration channels with conventional remote sensing interpretation tools such as object detection models, superpixel reconstruction models, scene classification models, and change detection models.

\begin{figure*}[!t]
\centering
\includegraphics[width=0.9\textwidth]{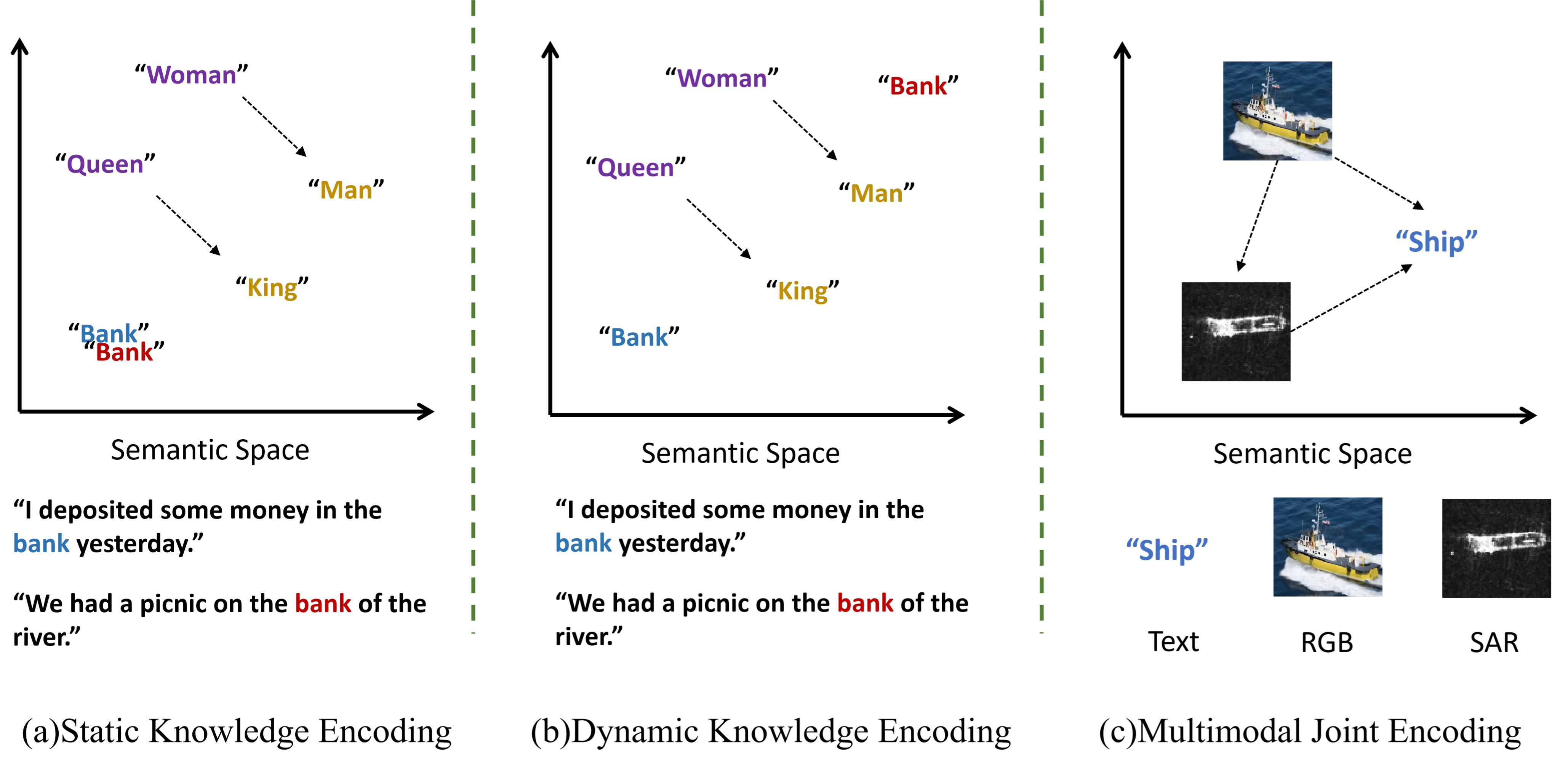}%
\caption{Illustration of knowledge encoding methods.}
\label{knowledge_encoding}
\end{figure*}

This architecture enables the model to retrieve task-relevant knowledge from a broader external knowledge base, thereby improving its capacity for task understanding. However, RS-Agent\cite{xu2024rs} relies solely on external knowledge bases for information retrieval and lacks mechanisms for interacting with and receiving feedback from the real physical environment. As a result, the model struggles to assess whether task decomposition is accurate, appropriate, and efficient under dynamic and open-world conditions. Thus, enabling effective interaction and feedback with the physical environment will be a critical challenge for the future development of RS-Agent\cite{xu2024rs} and similar research efforts.

\subsubsection{Knowledge Retrieval}

Once the task requirements have been parsed and clarified during the task understanding process, the core challenge shifts to how to extract highly relevant information from internal and external knowledge bases to support subsequent reasoning and decision-making. Two fundamental components of knowledge retrieval are knowledge encoding and knowledge querying. Knowledge encoding focuses on transforming external knowledge into representations that the model can understand and utilize effectively in later stages. Knowledge retrieval, on the other hand, is guided by the specific content of the task requirements and aims to selectively extract information that is highly relevant from both internal and external sources. This component ensures that the model has access to accurate and timely knowledge, which is critical for effective reasoning and task execution. Through precise knowledge retrieval, the model can compensate for its internal knowledge gaps and leverage existing knowledge bases to further improve its ability and accuracy in handling complex tasks.

\paragraph{Knowledge Encoding} External knowledge is typically stored in the form of data. The process of knowledge encoding essentially involves transforming this multimodal data into vector representations and mapping it into a unified vector space to facilitate subsequent knowledge association and retrieval. Within this space, semantically related knowledge should be located close together, while irrelevant knowledge should be distant. The core objective of this process is to enable the model to efficiently search, access, and integrate relevant knowledge. With the advancement of knowledge retrieval technologies, knowledge encoding has evolved through three key stages: static encoding, dynamic encoding, and multimodal joint encoding.

\begin{itemize}
    \item{\textbf{Static Knowledge Encoding}: As illustrated in Fig.\ref{knowledge_encoding}(a), static knowledge encoding maps different words into a shared semantic space where semantic relationships between words are preserved---for example, between ``Woman" and ``Queen", or ``Queen" and ``King". One of the pioneering methods in this domain is word2vec\cite{church2017word2vec}, which effectively captures both semantic and syntactic relationships between words. It uses two architectures, CBOW (continuous bag of words) and skip-gram, to project high-dimensional word representations into a lower-dimensional vector space, ensuring that semantically similar words are positioned closer together. However, a major limitation of word2vec\cite{church2017word2vec} is its inability to capture the dynamic semantic shifts of words across different contexts. To address this shortcoming, GloVe\cite{pennington2014glove} proposed a matrix factorization approach based on word co-occurrence statistics, allowing it to encode global semantic relationships more effectively. Building on this work, FastText\cite{joulin2016fasttext} refined word representation further by breaking words down into subword units. This capability enables the model to learn internal word structures, thereby improving its ability to represent low-frequency or out-of-vocabulary words. Nevertheless, static knowledge encoding fails to reflect contextual semantic variation. As shown in Fig.\ref{knowledge_encoding}(a), methods such as word2vec\cite{church2017word2vec} tend to map the word ``bank"---regardless of whether it refers to a financial institution or a riverbank---to the same point in the semantic space, leading to semantic ambiguity.}

    \vspace{0.1cm}

    \item{\textbf{Dynamic Knowledge Encoding}: Unlike static encoding methods, which assign a fixed vector representation to each word, dynamic knowledge encoding generates context-dependent word representations, enabling more accurate semantic interpretation, especially in the presence of contextual variability and polysemy. As illustrated in Fig.\ref{knowledge_encoding}(b), the word ``bank" can be mapped to different semantic regions depending on its surrounding context-'whether it refers to a financial institution or a riverbank. A representative method in this category is ELMo\cite{sarzynska2021detecting}, which uses a pretrained bidirectional language model to produce dynamic word embeddings based on the surrounding context, effectively capturing the context-specific meanings of words. BERT\cite{koroteev2021bert} takes this approach a step further by employing the bidirectional self-attention mechanism of the transformer architecture, thus allowing it to simultaneously consider both the left and right context of a word to generate highly context-aware embeddings. In contrast to BERT\cite{koroteev2021bert}, GPT\cite{radford2018improving} is a unidirectional language model trained to predict the next token in a sequence. This training strategy enables GPT\cite{radford2018improving} to compress semantic information from the input corpus efficiently, despite not incorporating the full bidirectional context.}

    \vspace{0.1cm}

    \item{\textbf{Multimodal Joint Encoding}: While dynamic knowledge encoding has made significant progress in handling language-based knowledge representation, it struggles to effectively process multimodal data. In contrast, multimodal knowledge encoding offers a more comprehensive and unified solution for integrating information across diverse modalities. As illustrated in Fig.\ref{knowledge_encoding}(c), this approach enables the joint representation of vision, language, and other modalities to describe a shared semantic concept, laying a solid foundation for downstream knowledge retrieval and reasoning. In this field, MKBE\cite{pezeshkpour2018embedding} constructs a multimodal knowledge base by unifying the representations of visual and language modalities. During the construction process, MKBE\cite{pezeshkpour2018embedding} focuses on preserving the temporal characteristics of each modality and the relationships between entities expressed across modalities. For example, for the visual modality, MKBE\cite{pezeshkpour2018embedding} uses convolutional neural networks (CNNs) to extract spatial and semantic features from images. For the language modality, LSTM\cite{hochreiter1997long} networks are employed to capture sequential dependencies between words. At the entity-relation level, MKBE\cite{pezeshkpour2018embedding} explicitly expresses cross-modal relationships by designing subject-relation pairs as training ground truth and uses a cross-entropy loss function to train the model to recognize entity relationships. Building on MKBE\cite{pezeshkpour2018embedding}, MULE\cite{kim2020mule} further enriches the language modality by supporting multilingual representations, thereby improving the model's compatibility with multiple languages. Although these methods have made notable advances in multimodal joint encoding, they still require dedicated embedding modules for each individual modality. With the rapid development of LLMs, newer approaches such as E5-V\cite{jiang2024e5} leverage MLLMs as unified encoders, allowing different modalities to be directly mapped into a shared embedding space without the need for modality-specific components. This approach significantly simplifies the implementation of multimodal joint encoding and improves the efficiency of cross-modal data integration.

    \vspace{0.1cm}

    In the field of remote sensing, the unified representation and understanding of multimodal data remains a central challenge for intelligent interpretation systems. The main difficulty lies in effectively fusing information from different modalities. With the rapid advancement of LLMs, a language-centered paradigm for multimodal joint embedding has gained increasing attention. EarthMarker\cite{zhang2024earthmarker} applies this paradigm to remote sensing by embedding visual modality data into the representational space of language models, thereby constructing a modality-unified semantic space. Within this framework, EarthMarker\cite{zhang2024earthmarker} introduces MoV (Mixture of Visual Experts) and a modality alignment module to integrate remote sensing imagery, visual object prompts, and textual instructions. This combination enables a unified understanding and representation of heterogeneous multimodal information. This approach not only enhances the interpretability of remote sensing data but also offers a novel pathway for intelligent processing in multimodal tasks.
    }

\end{itemize}

\begin{figure*}[!t]
\centering
\includegraphics[width=0.9\textwidth]{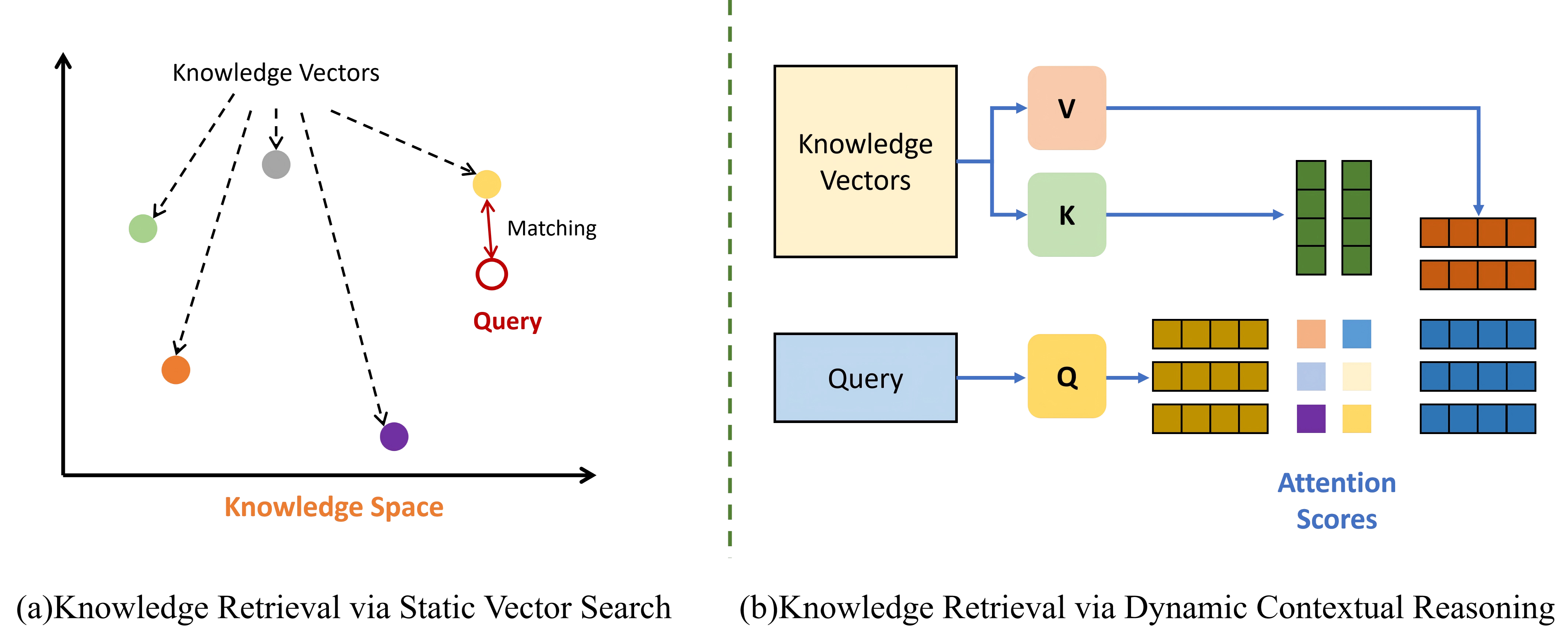}%
\caption{Illustration of knowledge retrieval methods.}
\label{knowledge_retrieval}
\end{figure*}

In summary, knowledge encoding transforms external knowledge into vector representations that can be understood by models, laying the foundation for subsequent knowledge retrieval. The development of knowledge encoding has gone through three major stages---static encoding, dynamic encoding, and multimodal joint encoding----each enhancing the expressive power of knowledge in different ways. Static encoding methods, such as word2vec\cite{church2017word2vec}, GloVe\cite{pennington2014glove}, and FastText\cite{joulin2016fasttext}, focus on capturing semantic relationships between words. However, they are limited by their inability to dynamically adapt to changes in context. Dynamic encoding methods, such as ELMo\cite{pezeshkpour2018embedding}, BERT\cite{koroteev2021bert}, and GPT\cite{radford2018improving}, can adjust word embeddings based on contextual information, allowing more flexible and accurate semantic interpretation. Multimodal joint encoding goes a step further by integrating information from multiple modalities---such as vision and language---thus providing robust support for cross-modal knowledge retrieval and reasoning.

\paragraph{Knowledge Retrieval} Knowledge retrieval is a crucial step following knowledge encoding. It aims to extract information that is highly relevant to the current task from large-scale knowledge bases or databases. This information is then used to support subsequent reasoning and decision-making processes. With the advancement of technology, knowledge retrieval has evolved from traditional keyword-based search methods to more intelligent approaches, such as static vector-based retrieval and dynamic context-aware reasoning. These modern methods enable models to identify the most valuable and relevant knowledge from massive information repositories, even in complex task scenarios, thereby enhancing task comprehension and decision-making performance.

\begin{itemize}
    \item{\textbf{Static Vector Search}: Static vector-based knowledge retrieval relies on precomputed knowledge embeddings. As illustrated in Fig.\ref{knowledge_retrieval}(a), the system retrieves relevant knowledge by calculating the similarity between the query vector and stored knowledge vectors. Common techniques include TF-IDF\cite{aizawa2003information}, k-nearest neighbors (k-NN)\cite{cover1967nearest}, approximate nearest neighbor (ANN)\cite{arya1998optimal} search, and cosine similarity. TF-IDF\cite{aizawa2003information} evaluates the importance of keywords by measuring their frequency in a document relative to their prevalence across the entire corpus. However, it ignores semantic relationships between words. The k-NN\cite{cover1967nearest} algorithm finds the most similar knowledge vectors by computing the distances between the query and the stored vectors. To improve efficiency, the ANN\cite{arya1998optimal} uses optimized indexing structures to perform approximate retrieval, thus reducing computational cost while maintaining acceptable accuracy. These methods are effective for static vector retrieval but lack adaptability and often struggle with complex semantic understanding tasks, where deeper context and reasoning are required.}

    \vspace{0.1cm}

    \item{\textbf{Dynamic Contextual Reasoning}: Unlike static vector search, dynamic context-aware knowledge retrieval (as illustrated in Fig.\ref{knowledge_retrieval}(b)) leverages attention mechanisms\cite{vaswani2017attention, hou2019cross} to establish relationships between pieces of knowledge within a broader context, significantly improving retrieval precision in complex semantic understanding tasks. The attention mechanism operates by computing attention scores through interactions among queries, keys, and values, enabling the model to identify the most relevant knowledge across the global context. In traditional attention frameworks, the query is matched against all key-value pairs in the knowledge base. However, as the number of knowledge vectors increases, the computational cost of this matching process increases dramatically, leading to long retrieval times. To address this issue, block attention\cite{sun2024block} introduces a blockwise strategy, dividing large query sequences into smaller chunks and performing attention calculations within each block independently. This approach significantly reduces the computational overhead. For example, with a prompt length of 32K, block attention reduces the TTFT-block latency by 98.7\%. By optimizing the attention-based retrieval process, this method improves the efficiency of dynamic knowledge retrieval in large-scale knowledge bases.

    \vspace{0.1cm}

    In the remote sensing domain, effective text-to-image retrieval is a key challenge for Earth observation tasks. Owing to the vast volume of remote sensing imagery and the minimal variance among image embeddings, traditional text retrieval techniques often struggle to match images accurately. To address this shortcoming, the KTIR\cite{mi2024knowledge} (knowledge-to-image retrieval) framework proposes a cross-modal retrieval method. It attributes poor retrieval performance to the lack of differentiation in textual descriptions. KTIR\cite{mi2024knowledge} addresses this issue by converting structured knowledge within remote sensing images into textual expressions and then applying a cross-attention\cite{hou2019cross} mechanism that uses these expressions as queries for retrieval. This approach significantly improves the retrieval accuracy between remote sensing images and textual descriptions.
    }
\end{itemize}

In summary, static vector-based retrieval methods perform information retrieval by calculating similarity scores between query and knowledge vectors, but they face significant limitations in capturing contextual variations. To overcome this challenge, dynamic context-aware knowledge retrieval methods leverage attention mechanisms to establish relationships across a broader context, thereby enhancing the accuracy of knowledge retrieval in complex semantic scenarios.

\begin{table*}[!t]
\caption{Corresponding Representative methods of Language-Centered Reasoning and Decision-Making.
\label{tab:3_4}}
\centering
\begin{tabular}{>{\raggedright\arraybackslash}p{2.5cm} 
                >{\raggedright\arraybackslash}p{2.5cm} 
                >{\raggedright\arraybackslash}p{7.5cm}}
\toprule
\multirow{1}{*}{Content structure} & Category & Representative algorithms \\

\midrule

\multirow{7}{*}{Decision-Making} 
                    & Tool Decision & {ReACT\cite{yao2023react}, DFSDT\cite{qin2023toolllm}, Tool-bench\cite{qin2023toolllm}, ToolLLaMA\cite{touvron2023llama}, Levine et al.\cite{levine2018learning}, LangChain\cite{pandya2023automating}, AutoGPT\cite{yang2023auto}, CLIPort\cite{shridhar2022cliport}, Toolformer\cite{schick2023toolformer}, Re-Invoke\cite{chen2024re}, WebGPT\cite{nakano2021webgpt}, WebCPM\cite{qin2023webcpm}} \\

\cmidrule(l){2-3}
                    & Action Execution & {Hu and Pagilla\cite{hu2022view}, Namiko\cite{saito2021select}, PID\cite{johnson2005pid}, LQR\cite{guardeno2019mimo}, DeepMPC\cite{amos2018differentiable}, DeepGait\cite{tsounis2020deepgait}, GDQ\cite{marchesini2021centralizing}, RobotPilot\cite{jin2023robot}, PIBOT\cite{min2024toward}} \\
\bottomrule
\end{tabular}
\end{table*}

\subsection{Language-Centered Decision-Making and Execution}

Within the closed loop of remote sensing interpretation, the core task of decision-making and execution is to formulate subsequent actionable strategies that advance the interpretation process by synthesizing task requirements, domain knowledge, and the agent's capability boundaries.

However, the critical bottleneck constraining interpretation quality often stems from perceptual limitations, which manifest at two primary levels:

\begin{itemize}

    \item Functional Level: The agent lacks the capacity to directly invoke existing tools or algorithms to satisfy complex interpretation needs;
    
    \item Informational Level: The current observed imagery lacks the essential visual information required to complete the reasoning process.
    
\end{itemize}

To address these two levels of challenges, this section explores methods related to tool-based decision-making and action execution. Representative methods are summarized in Table \ref{tab:3_4}.

\subsubsection{Decision-Making and Execution}
Once the gap between the current interpretation process and the task objective is clearly identified, the core challenge shifts to determining how to make informed decisions concerning the next interpretation actions, using task-relevant knowledge and the agent's capabilities, to narrow that gap. This gap is caused primarily by insufficient perceptual information, which arises from two dimensions: the data level and the functional level. To address these issues, it is necessary to adopt action-level decision-making methods and tool-level decision-making strategies.

\paragraph{Tool decision} challenges at the functional level arise primarily from the semantic gap between an agent's natural language decision-making and rigid remote sensing interpretation tools. While an agent's reasoning process is represented in natural language, interpretation tools (e.g., object detection algorithms and change detection models) rely on specific code interfaces and parameter configurations. Therefore, establishing a precise mapping from linguistic instructions to executable code is the core problem of tool decision-making. There are two main categories of solutions:

\begin{itemize}

    \item Tuning-free methods, such as ReACT\cite{yao2023react} and DFSDT\cite{qin2023toolllm}, which guide LLMs via meticulously designed prompts to align inferred subtask descriptions with tool function descriptions.

    \item Fine-tuning methods, represented by studies such as Tool-bench\cite{qin2023toolllm}, have shown that tuning-free methods have limited precision in complex matching scenarios. Consequently, ToolLLaMA\cite{touvron2023llama} employs instruction tuning to increase the model's accuracy in terms of tool selection and invocation.
    
\end{itemize}

In the remote sensing domain, tool decision-making is manifested by UAVs or satellite platforms flexibly invoking onboard equipment or analysis algorithms based on task requirements. Early methods relied heavily on hand-crafted rules or geometric model-based strategies, completing tool operations via predefined task plans. However, these approaches are highly scene-dependent and lack generality. With the advent of deep learning, deep reinforcement learning (DRL) was introduced to grant agents autonomous learning capabilities. For instance, the end-to-end framework proposed by Levine et al.\cite{levine2018learning} enables robots to optimize tool-use strategies through environmental feedback. Nonetheless, DRL methods remain computationally expensive and show weak adaptability during task migration.

Recently, LLM-based agent research has emerged as a new paradigm for remote sensing tool decision-making. Frameworks such as LangChain\cite{pandya2023automating} and AutoGPT\cite{yang2023auto} utilize LLMs to decompose tasks and generate invocation commands. CLIPort\cite{shridhar2022cliport} integrates vision-language models with manipulation control modules to achieve efficient tool operation in multitask scenarios. Toolformer\cite{schick2023toolformer} enables language models to master external API calls through self-supervised learning, while Re-Invoke\cite{chen2024re} uses LLMs to extract user intent and retrieves the most relevant tools via multiview similarity ranking. Additionally, works such as WebGPT\cite{nakano2021webgpt} and WebCPM\cite{qin2023webcpm} integrate search and browsing tools directly into LLMs, achieving an integrated pipeline from task analysis to tool invocation.

\paragraph{Action execution} challenges at the informational level manifest as a lack of critical visual information in collected imagery, which is typically caused by physical factors such as restricted observation angles, poor image quality, or occlusions. Action execution serves as an active sensing mechanism to address this problem. Its core lies in dynamically adjusting the physical state of sensors (e.g., position, angle, modality) based on current perceptual feedback to proactively supplement or improve data quality. In general robotics, Hu and Pagilla\cite{hu2022view} regulated robotic arms using perceptual feedback to control sensor orientation and obtained higher-quality point clouds for precise pose estimation. Namiko et al.\cite{saito2021select} combined multimodal sensor feedback with robotic motion states to achieve accurate tool-object-action associations.

In the remote sensing domain, action execution methods are primarily applied to the dynamic adjustment of the flight attitude, altitude, speed, and observation angles of UAVs or satellites. Traditional methods based on classical control theory, such as PID\cite{johnson2005pid} and LQR\cite{guardeno2019mimo}, perform excellently in stability control but face limitations in nonlinear, multivariable coupled dynamic environments. As technology has evolved, DeepMPC\cite{amos2018differentiable} combines reinforcement learning with model predictive control for real-time attitude adjustment. DeepGait\cite{tsounis2020deepgait} optimizes terrain adaptation in quadruped robots by combining visual input with global localization. GDQ \cite{marchesini2021centralizing} uses global value networks to enhance multiagent coordination in mapless navigation. Furthermore, imitation learning has been used in applications, such as RobotPilot\cite{jin2023robot}, which achieves high-precision task-oriented control by mimicking human operation trajectories. Recent studies have further integrated VLM; for instance, PIBOT\cite{min2024toward} combines VLMs with pose estimation to enable autonomous flight control for humanoid pilots, thus demonstrating the effective guidance of language-driven high-level cognition over low-level action execution.

\section{Training Datasets and Evaluation Benchmarks}

Data and evaluation benchmarks serve not only as the fuel driving the performance of deep learning models but also as a direct mirror reflecting the evolution of interpretation paradigms \cite{sun2017revisiting,kaplan2020scaling,bommasani2021opportunities}. 
In the transition of remote sensing interpretation from a ``visual-centered'' to a ``language-centered'' paradigm, we observe distinct staged characteristics in data modalities: a gradual transition from early ``image--label'' pairs to ``image--caption'' descriptions and, finally, to ``image--instruction'' corpora that support the cognitive capabilities of general foundation models (GFMs) \cite{kuckreja2024geochat,luo2024skysensegpt,shu2025earthmind,liu2024rsunivlm}. 
This enrichment of data dimensionality has not only shattered the closed semantic boundaries of traditional visual tasks but also enabled higher-level reasoning and interactive capabilities, which have been increasingly emphasized in recent multimodal instruction-tuned models. 
This section provides a systematic review of the representative training datasets and evaluation benchmarks within this evolving ecosystem.

\subsection{Training Datasets}

In the developmental trajectory of remote sensing intelligent interpretation, representative datasets are summarized in the TABLE \ref{tab:rs_vl_datasets}. Within the language-centered interpretation framework, different types of datasets play distinct roles. Based on their functional positioning in model construction, we categorize existing representative datasets into four groups: visual representation learning, cross-modal semantic alignment, large-scale foundation pretraining, and instruction tuning.

\begin{table*}[t]
\centering
\caption{Summary of Representative Training Datasets}
\label{tab:rs_vl_datasets}
\renewcommand{\arraystretch}{1.15}
\setlength{\tabcolsep}{5pt}
\begin{tabular}{l c c c c c p{4.4cm}}
\toprule
\multirow{2}{*}{\textbf{Dataset}} &
\multirow{2}{*}{\textbf{Year}} &
\multicolumn{2}{c}{\textbf{Image}} &
\multicolumn{2}{c}{\textbf{Text}} &
\multirow{2}{*}{\textbf{Primary Purpose}} \\
\cmidrule(lr){3-4} \cmidrule(lr){5-6}
& & \textbf{Num} & \textbf{Resolution} & \textbf{Num} & \textbf{Form} & \\
\midrule
UCM\cite{yang2010bag} & 2010 & 2,100 & $256\times256$ & -- & -- &
Visual Representation Learning \\

UCM-Caption\cite{qu2016deep} & 2016 & 2,100 & $256\times256$ & 2,100 & Sentence &
Cross-Modal Semantic Alignment \\

AID\cite{xia2017aid} & 2017 & 10,000 & $600\times600$ & -- & -- &
Visual Representation Learning \\

NWPU-RESISC45\cite{cheng2017remote} & 2017 & 31,500 & $256\times256$ & -- & -- &
Visual Representation Learning \\

DOTA\cite{dota} & 2018 & 2,806 & -- & -- & -- &
Visual Representation Learning \\

iSAID\cite{waqas2019isaid} & 2019 & 2,806 & -- & -- & -- &
Visual Representation Learning \\

RSVQA-LR\cite{rsvqa} & 2020 & 772 & $256\times256$ & 77,232 & Question + Options &
Instruction Tuning \\

RSVQA-HR\cite{rsvqa} & 2020 & 10,659 & $512\times512$ & 1,066,316 & Question + Options &
Instruction Tuning \\

MillionAID\cite{million-aid} & 2021 & $\sim$1,000,000 & -- & -- & -- &
Large-Scale Foundation Pretraining \\

NWPU-Caption\cite{nwpu-caption} & 2022 & 31,500 & $256\times256$ & 157,500 & Sentence &
Cross-Modal Semantic Alignment \\

RSVGD\cite{rsvgd} & 2022 & 55,722 & $800\times800$ & 38,320 & Phrase &
Cross-Modal Semantic Alignment \\

SkyScript\cite{skyscript} & 2023 & $\sim$5,200,000 & -- & 2,600,000 & Phrase + Sentence &
Large-Scale Foundation Pretraining \\

RS5M\cite{rs5m} & 2023 & $\sim$5,000,000 & -- & 3,000,000 & Phrase &
Large-Scale Foundation Pretraining \\

RRSIS-D\cite{yuan2023rrsis} & 2024 & 17,402 & - & 17,402 & Phrase & Cross-Modal Semantic Alignment \\

VRSBench\cite{vrsbench} & 2024 & 29,614 & -- & 205,307 & Phrase + Sentence + QA &
Instruction Tuning \\

GEOBench-VLM\cite{geobench} & 2024 & -- & -- & 10,000 & Sentence + QA + Options &
Instruction Tuning \\

CHOICE\cite{an2024choice} & 2025 & 2,000 & $512\times512$ & 10,507 & Question + Options &
Instruction Tuning \\
\bottomrule
\end{tabular}
\end{table*}

\begin{itemize}
    \item {\bf{Visual Representation Learning}}: This category of datasets is primarily used to provide models with feature extraction capabilities and serves as a control group (or counterpart) for evaluating the zero-shot classification accuracy of multimodal models. Scene classification datasets, represented by NWPU-RESISC45\cite{cheng2017remote} and MillionAID\cite{million-aid}, define rigorous categorical systems. Although they do not provide complex semantic narratives, their high-resolution imagery and clear class boundaries serve as essential benchmarks to test whether VLMs retain fine-grained discriminative power for remote sensing objects. Furthermore, DOTA\cite{dota} provides annotations for oriented bounding boxes (OBBs) in dense and arbitrary orientations, catering to the unique perspectives of RS imagery. Within a language-centered framework, it is frequently used to validate a model's foundational performance in geometric perception tasks, such as spatial localization and object counting.

    \item {\bf{Cross-Modal Semantic Alignment}}: The core role of these datasets is to bridge the gap between pixel features and high-level semantics by introducing natural language; they form the functional basis for training models to output descriptive language. Datasets such as UCM-Caption\cite{qu2016deep} and NWPU-Caption\cite{nwpu-caption} utilize human-written sentences to describe image content. Unlike discrete labels, these descriptions provide supervisory signals for syntactic structures and attribute associations. Their emergence enables models to learn how to ``translate" visual features into natural language descriptions, successfully achieving alignment between visual and linguistic modalities.

    \item {\bf{Large-Scale Foundation Pretraining}}: Addressing the demand of RS Large Foundation Models for ``open-world knowledge," these datasets provide universal cognitive capabilities through their massive scale and open semantic space. SkyScript\cite{skyscript} and RS5M\cite{rs5m} are currently the most representative pretraining corpora. By leveraging geographic information associations or large-scale filtering, they have constructed millions of pairs consisting of high-resolution RS images and their corresponding fine-grained semantic descriptions. The immense scale of these datasets supports CLIP-style contrastive pretraining, covering long-tail concepts and geospatial knowledge. They are the critical source for a model's zero-shot generalization and open-world understanding.

    \item {\bf{Instruction Tuning}}: As model capabilities expand toward higher-order cognition, traditional single-task testing is no longer sufficient to guide or measure comprehensive intelligence. The new generation of datasets focuses on evaluating instruction following, multitask generalization, and complex logical reasoning. VRSBench\cite{vrsbench} and GEOBench-VLM\cite{geobench} are standardized multitask evaluation datasets; the former focuses on quantifying transfer learning capabilities across tasks, while the latter validates unified multitask performance under complex instructions. The more recent CHOICE\cite{an2024choice} dataset utilizes a multiple-choice format to rigorously examine geospatial-temporal reasoning and logical analysis. It requires models to perform multistep reasoning to evaluate their trustworthiness and hallucination levels, marking a shift in evaluation dimensions deep into the human cognitive level.
    
\end{itemize}

\begin{table*}[t]
\centering
\caption{Classification Accuracy (\%) Comparison on Scene Classification Tasks}
\label{tab:rs_classification_comparison}
\renewcommand{\arraystretch}{1.2}
\setlength{\tabcolsep}{8pt}
\begin{tabular}{l l l l c c} 
\toprule
& 
\multirow{2}{*}{\textbf{Method}} & 
\multirow{2}{*}{\textbf{Type}} & 
\multirow{2}{*}{\textbf{Publication Year}} & \multicolumn{2}{c}{\textbf{Dataset}} \\ \cmidrule(lr){5-6}
&  &  &  & AID\cite{xia2017aid} & NWPU-RESISC45\cite{cheng2017remote} \\ 
\midrule

\multirow{5}{*}{visual-centered} 
& CLIP\cite{radford2021learning} & \multirow{5}{*}{VLM} & ICML'2021 & 69.25 & 66.70 \\
& RemoteCLIP\cite{liu2024remoteclip} &  & TGRS 2024 & 74.30 & 70.85 \\
& SkyScript\cite{skyscript} &  & AAAI 2024 & 70.94 & 71.70 \\
& GeoRSCLIP\cite{vivanco2023geoclip} &  & TGRS 2024 & 76.27 & 73.90 \\
& SkySense\cite{zhu2025skysense} &  & CVPR 2025 & 82.93 & 83.28 \\

\midrule

\multirow{11}{*}{Language-Centered} 
& GeoChat\cite{kuckreja2024geochat} & \multirow{8}{*}{MLLM} & CVPR 2024 & 72.03 & - \\
& EarthGPT\cite{zhang2024earthgpt} &  & TGRS 2024 & 77.37 & 74.72 \\
& SkySenseGPT\cite{luo2024skysensegpt} &  & arXiv 2024 & 75.50 & 83.33 \\
& EarthDial\cite{soni2025earthdial} &  & CVPR 2025 & 67.33 & 76.67 \\
& EarthMarker\cite{zhang2024earthmarker} &  & TGRS 2024 & 77.97 & - \\
& TEOChat\cite{irvin2024teochat} &  & ICLR 2025 & 80.90 & - \\
& LHRS-Bot\cite{muhtar2024lhrs} &  & ECCV2024 & 91.26 & 83.94 \\
& VHM\cite{pang2025vhm} &  & AAAI 2025 & 91.70 & 95.80 \\ \cmidrule(lr){2-6}
& RingMo-Agent\cite{hu2025ringmo} & \multirow{3}{*}{LLM-Based Agent} & arXiv 2025 & 92.43 & 93.96 \\
& Earth-Agent\cite{feng2025earth} &  & arXiv 2025 & 93.42 & 96.12 \\
& RSThinker\cite{liu2025towards} &  & arXiv 2025 & \textbf{98.17} & \textbf{96.89} \\

\bottomrule
\end{tabular}
\end{table*}

\subsection{Performance Evaluation and Analysis on Downstream Tasks}
To systematically validate the effectiveness of the language-centered interpretation paradigm across multi-level remote sensing tasks, this paper establishes a comprehensive evaluation framework encompassing both perception and cognition. Specifically, the experimental tasks are categorized into two complementary domains: 

\begin{itemize}

    \item Visual Perception Tasks: Designed to evaluate the model's capability in the fine-grained extraction of geospatial elements, primarily serving applications such as Land Use and Land Cover (LULC) analysis \cite{kuckreja2024geochat,liu2024rsunivlm}.
    
    \item Semantic Understanding and Reasoning Tasks: Built upon the foundation of visual perception, these tasks address more complex interactive requirements, include image caption, visual question answering and complex reasoning \cite{liu2023visual,shu2025earthmind,openai_gpt4_2023}.
    
\end{itemize}

\subsubsection{Visual Perception Tasks} Scene classification and semantic segmentation constitute the cornerstone of LULC analysis. Under the visual-centered paradigm, both tasks are fundamentally treated as end-to-end visual pattern matching within closed sets, functioning as global feature mapping and dense pixel classification, respectively. Although efficient on specific datasets, this approach suffers from limited generalization in open-world scenarios or complex environments characterized by spectral confusion (i.e, same objects with different spectra), primarily due to a lack of explicit understanding of geospatial attributes.

Conversely, the language-centered paradigm reframes these tasks as knowledge-driven concept reasoning and semantic grounding. This approach leverages linguistic priors to parse the logical composition of visual elements, guiding precise pixel alignment within an open semantic space. This paradigm shift transcends the limitations of low-level feature matching; by incorporating high-level semantic constraints, it significantly boosts the universality and robustness of interpretation systems across unseen classes and cross-domain settings.

\begin{itemize}

    \item Experiments on Scene Classification: TABLE \ref{tab:rs_classification_comparison}  presents a comparative analysis of zero-shot classification performance across different interpretation paradigms on the AID\cite{xia2017aid} and NWPU-RESISC45\cite{cheng2017remote} datasets. Examining the accuracy distribution across models reveals distinct phase-wise differences in generalization performance within open scenarios as the field evolves from the visual-centered paradigm to the language-centered paradigm.

    \vspace{0.1cm}

    Regarding the visual-centered paradigm, experimental data indicates that performance gains are gradually plateauing. Early general VLMs (e.g., CLIP\cite{radford2021learning}) achieved an accuracy of only 66.70\% on NWPU-RESISC45\cite{cheng2017remote} due to significant domain shifts. Although subsequent works (e.g., RemoteCLIP\cite{liu2024remoteclip}, GeoRSCLIP\cite{rs5m}) raised accuracy to approximately 74\% by introducing domain-specific image-text alignment training, the magnitude of improvement has progressively narrowed. Notably, SkySense-O\cite{zhu2025skysense}, the state of the art (SOTA) model within this paradigm, pushed the accuracy further to 83.28\% through large-scale vision-language contrastive learning and hard negative mining. These results suggest that while optimizing implicit alignment strategies within the feature space can enhance discriminative power to a certain extent, diminishing marginal returns have begun to emerge when confronting the complex semantics of unseen categories.

    \vspace{0.1cm}

    In contrast, the language-centered paradigm, particularly models incorporating Agent architectures, has demonstrated a significantly higher performance ceiling. Early MLLMs (e.g., EarthGPT\cite{zhang2024earthgpt}, GeoChat\cite{kuckreja2024geochat}) primarily established cross-modal associations through basic instruction tuning. Constrained by the limitations of single-step reasoning in capturing fine-grained visual features, their zero-shot accuracy generally stabilized within the 72\%-77\% range. However, with the introduction of chain-of-thought reasoning and tool invocation mechanisms, agent-based models have exhibited substantial advantages. Both RSThinker\cite{liu2025towards} and Earth-Agent\cite{feng2025earth} achieved results exceeding 93\% across the two datasets, with RSThinker\cite{liu2025towards} reaching 98.17\% on AID\cite{xia2017aid}.

    \vspace{0.1cm}

\begin{table*}[t]
\centering
\caption{Semantic Segmentation Comparison on RRSIS-D\cite{chen2025rsrefseg} Dataset (\%)}
\label{tab:rs_semanticseg_comparison}
\renewcommand{\arraystretch}{1.2}
\setlength{\tabcolsep}{8pt}
\begin{tabular}{l l l l l c c} 
\toprule
& 
\multirow{2}{*}{\textbf{Method}} & 
\multicolumn{2}{c}{\textbf{Type}}  & 
\multirow{2}{*}{\textbf{Publication Year}} & \multicolumn{2}{c}{\textbf{RRSIS-D\cite{yuan2023rrsis}}} \\ \cmidrule(lr){3-4} \cmidrule(lr){6-7}
&  & Vision-Encoder & Language-Encoder &  & cIoU & gIoU \\ 
\midrule

\multirow{5}{*}{visual-centered} 
& RMISN & ViT\cite{liu2024rotated} & BERT\cite{koroteev2021bert} & CVPR'2024 & 76.50 & 62.27 \\
& LGCE\cite{yuan2023rrsis}  & ViT\cite{liu2024rotated} & BERT\cite{koroteev2021bert} & TGRS'2024 & 76.24 & 61.02 \\
& FIANet\cite{lei2024exploring}  & Swin\cite{liu2021swin} & BERT\cite{koroteev2021bert} & TGRS'2024 & 76.91 & 64.01 \\
& CrossVLT\cite{cho2023cross}  & Swin\cite{liu2021swin} & BERT\cite{koroteev2021bert} & TMM'2023 & 76.32 & 61.00 \\
& RSRefSeg\cite{chen2025rsrefseg} & SAM\cite{kirillov2023segment} & CLIP-TE\cite{radford2021learning} & arXiv'2025 & 77.24 & 64.67 \\

\midrule

\multirow{3}{*}{Language-Centered} 
& Geopixel & CLIP-VE\cite{radford2021learning} & InternLM2\cite{cai2024internlm2} & ICML'2025 & \textbf{83.33} & 67.30 \\
& RemoteSAM\cite{yao2025remotesam} & Swin\cite{liu2021swin} & BERT\cite{koroteev2021bert} & ACM'2025 & 80.04 & 71.75 \\
& UniGeoSeg\cite{ni2025unigeoseg} & Swin\cite{liu2021swin} & Phi-1.5 & arXiv'2025 & 81.35 & \textbf{72.23} \\

\bottomrule
\end{tabular}
\end{table*}

    \item Experiments on Semantic Segmentation: The rapid advancement of VFMs has significantly enhanced pixel-level perception capabilities regarding the boundaries of geospatial objects. This has triggered a qualitative shift in the core challenge of semantic segmentation: the bottleneck is no longer the low-level clustering of homologous pixels, but rather the construction of high-level semantic referents. Fundamentally, the focus of the task has transformed into leveraging the logical constraints of linguistic instructions to guide class-agnostic VFMs in mapping generic visual boundaries to specific semantic categories.

    \vspace{0.1cm}

    TABLE \ref{tab:rs_semanticseg_comparison} presents a performance comparison of different interpretation paradigms on the referring remote sensing image segmentation (RRSIS-D\cite{chen2025rsrefseg}) task. Based on whether a chain-of-thought or an explicit logical reasoning process is incorporated, models are categorized into visual-centered and language-centered paradigms to reveal the mechanistic evolution from ``feature matching" to ``semantic reasoning."

    \vspace{0.1cm}

    Experimental data indicates that although visual-centered models, represented by RSRefSeg\cite{chen2025rsrefseg}, have significantly improved boundary perception by integrating SAM\cite{kirillov2023segment} (achieving a gIoU of 64.67\%), their performance ceiling remains constrained by the static, shallow alignment between visual and linguistic features. In contrast, the language-centered paradigm successfully breaks this bottleneck by reconstructing the interpretation mechanism from ``passive feature comparison" to ``explicit spatial-semantic logical deduction." Specifically, under the premise of employing similar visual foundation models, RemoteSAM\cite{yao2025remotesam}, which places language at the core, leverages deeper semantic instruction parsing to significantly boost gIoU to 71.75\% (+7.08\%). Furthermore, the latest UniGeoSeg\cite{ni2025unigeoseg} pushes the gIoU even higher to 72.23\% (+7.56\%).
    
\end{itemize}

\begin{figure*}[!t]
    \centering
    \includegraphics[width=\textwidth]{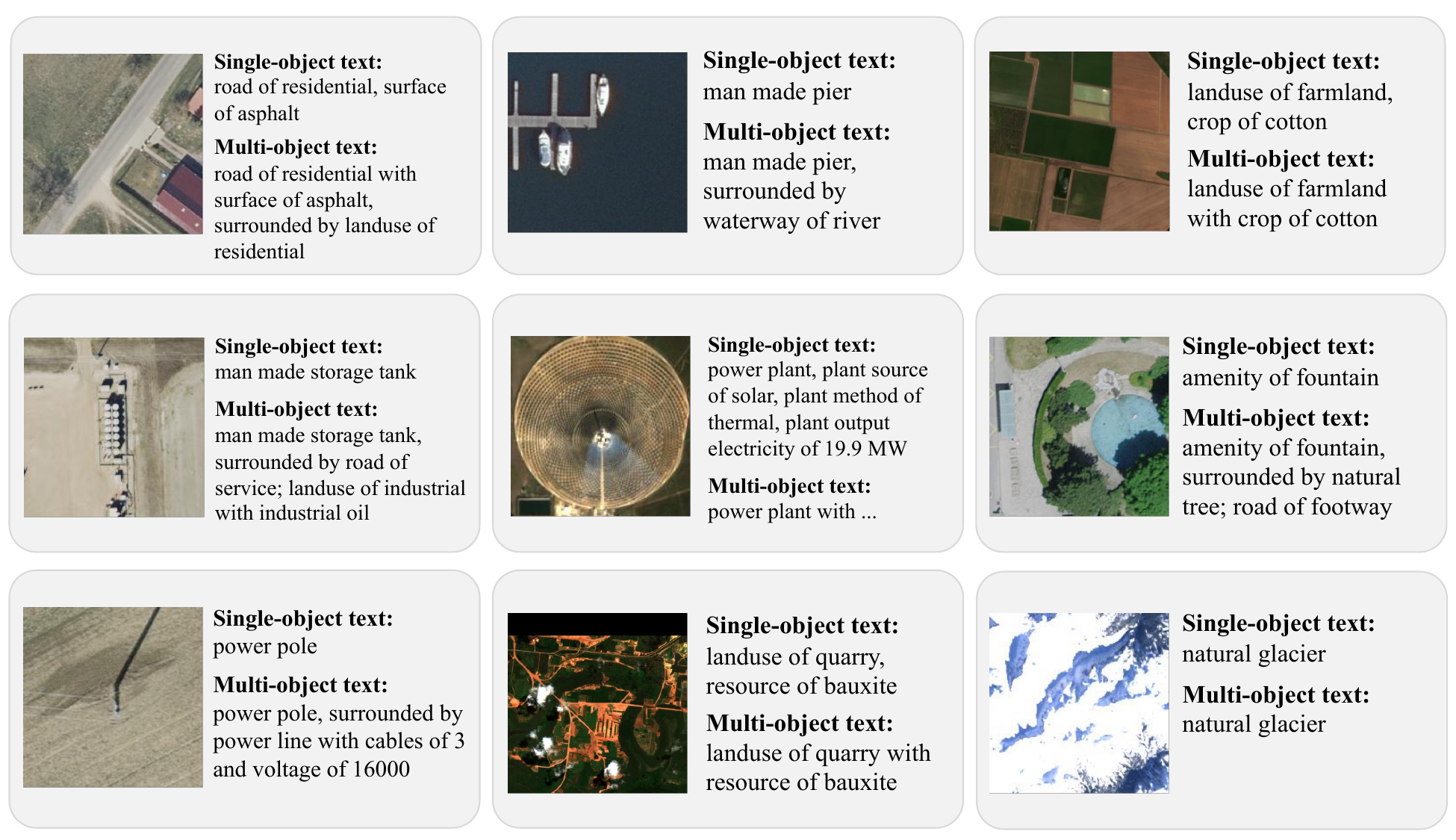}%
    \caption{Example of image captioning data sourced from \cite{wang2024skyscript}}
    \label{image_caption}
\end{figure*}

\begin{table*}[t]
\centering
\caption{Performance Comparison across Image Captioning Methods, data sourced from \cite{lin2025rs}}
\label{tab:captioning_comparison}
\renewcommand{\arraystretch}{1.15}
\setlength{\tabcolsep}{5pt}
\begin{tabular}{l c c c c c c c}
\toprule
\textbf{Method} &
\textbf{BLEU-1} &
\textbf{BLEU-2} &
\textbf{BLEU-3} &
\textbf{BLEU-4} &
\textbf{METEOR} &
\textbf{ROUGE\_L} &
\textbf{CIDEr} \\
\midrule
VLAD + RNN~\cite{lu2017exploring}        & 63.11 & 51.93 & 46.06 & 42.09 & 29.71 & 58.78 & 200.66 \\
VLAD + LSTM~\cite{lu2017exploring}     & 70.16 & 60.85 & 54.96 & 50.30 & 34.64 & 65.20 & 231.31 \\
mRNN~\cite{qu2016deep}                 & 60.10 & 50.70 & 32.80 & 20.80 & 19.30 & --    & 214.00 \\
mLSTM~\cite{qu2016deep}               & 63.50 & 53.20 & 37.50 & 21.30 & 20.30 & --    & 222.50 \\
mGRU~\cite{li2018multi}                 & 42.56 & 29.99 & 22.91 & 17.98 & 19.41 & 37.97 & 124.82 \\
mGRU embedword~\cite{li2018multi}       & 75.74 & 69.83 & 64.51 & 59.98 & 36.85 & 66.74 & 279.24 \\
CSMLF~\cite{wang2019semantic}               & 37.71 & 14.85 & 7.63  & 5.05  & 9.44  & 29.86 & 13.51 \\
SAA~\cite{lu2019sound}                   & 79.62 & 74.01 & 69.09 & 64.77 & 38.59 & 69.42 & 294.51 \\
Soft-attention~\cite{xu2015show}    & 74.54 & 65.45 & 58.55 & 52.50 & 38.86 & 72.37 & 261.24 \\
Hard-attention~\cite{xu2015show}    & 81.57 & 73.12 & 67.02 & 61.82 & 42.63 & 76.98 & 299.47 \\
SD-RSIC~\cite{sumbul2020sd}             & 74.80 & 66.40 & 59.80 & 53.80 & 39.00 & 69.50 & 213.20 \\
RTRMN (semantic)~\cite{wang2020retrieval}    & 55.26 & 45.15 & 39.62 & 35.87 & 25.98 & 55.38 & 180.25 \\
RTRMN (statistical)~\cite{wang2020retrieval} & 80.28 & 73.22 & 68.21 & 63.93 & 42.58 & 77.26 & 312.70 \\
SVM-D BOW~\cite{hoxha2021novel}            & 76.35 & 66.64 & 58.69 & 51.95 & 36.54 & 68.01 & 271.42 \\
SVM-D CONC~\cite{hoxha2021novel}           & 76.53 & 69.47 & 64.17 & 59.42 & 37.02 & 68.77 & 292.28 \\
Postprocessing~\cite{hoxha2023improving}  & 79.73 & 72.98 & 67.44 & 62.62 & 40.80 & 74.06 & 309.64 \\
RSGPT-13B~\cite{hu2025rsgpt}            & 86.12 & 79.14 & 72.31 & 65.74 & 42.21 & 78.34 & 333.23 \\
SkyEyeGPT-7B~\cite{zhan2025skyeyegpt}     & 90.71 & 85.69 & \textbf{81.56} & \textbf{78.41} & 46.24 & 79.49 & 236.75 \\
RS-LLaVA-13B~\cite{bazi2024rs}       & 90.00 & 84.88 & 80.30 & 76.03 & 49.21 & \textbf{85.78} & 355.61 \\
RS-CapRet-7B~\cite{silva2024large}      & 84.30 & 77.90 & 72.20 & 67.00 & 47.20 & 81.70 & 354.80 \\
\midrule
\textbf{RS-MoE-7B\cite{lin2025rs}} &
\textbf{94.81} & \textbf{87.09} & 79.57 & 72.34 &
\textbf{66.97} & 62.74 & \textbf{396.46} \\
\bottomrule
\end{tabular}
\end{table*}

\begin{table*}[t]
\centering
\setlength{\tabcolsep}{6pt}
\caption{Performance comparison on presence, comparison, and reasoning tasks (\%), data sourced from \cite{soni2025earthdial}}
\label{tab:presence_comp}

\begin{tabular}{lcccc|lccc}
\toprule
\textbf{Model} & \textbf{Presence} & \textbf{Comp} & \textbf{R/U} & \textbf{Avg.} &
\textbf{Model} & \textbf{Presence} & \textbf{Comp} & \textbf{Avg.} \\
\midrule
MiniGPTv2\cite{chen2023minigpt}        & 55.16 & 55.22 & 39.00 & 54.96 &
MiniGPTv2\cite{chen2023minigpt}        & 40.79 & 50.91 & 46.46 \\
Qwen-VL \cite{bai2023qwen}          & 38.57 & 67.59 & 61.00 & 55.35 &
Qwen-VL \cite{bai2023qwen}          & 66.44 & 60.41 & 63.06 \\
InternVL2-8B \cite{team2024internvl2} & 58.54 & 72.28 & 71.00 & 66.51 &
InternVL2-8B \cite{team2024internvl2} & \textbf{67.35} & 76.91 & \textbf{72.70} \\
GeoChat \cite{kuckreja2024geochat}        & 91.09 & 90.33 & 94.00 & 90.70 &
GeoChat \cite{kuckreja2024geochat}        & 58.45 & \textbf{83.19} & 72.30 \\
LHRS-Bot \cite{muhtar2024lhrs}       & 88.51 & 90.00 & 89.07 & 89.19 &
EarthGPT \cite{zhang2024earthgpt}      & 62.77 & 79.53 & 72.06 \\
\midrule
\textbf{EarthDial\cite{soni2025earthdial}} & \textbf{92.58} & \textbf{92.75} & \textbf{94.00} & \textbf{92.70} &
\textbf{EarthDial\cite{soni2025earthdial}} & 58.89 & 83.11 & 72.45 \\
\bottomrule
\end{tabular}
\end{table*}

\subsubsection{Image Caption}

Image captioning, which serves as a bridge between visual perception and language generation, aims to automatically generate natural language sentences that elaborate on object attributes, spatial relationships, and scene contexts within remote sensing imagery (as shown in Fig. \ref{image_caption}). Unlike traditional scene classification tasks, which can only output discrete, predefined closed-set labels (such as ``airport" or ``forest"), image captioning requires models to possess higher-order semantic compositionality. This feature allows the model to understand and express fine-grained information, such as ``an airplane taking off from a runway" or ``sparse vegetation covered by snow."

The shift toward the language-centered paradigm endows interpretation systems with zero-shot and open-set adaptation capabilities, enabling them to leverage the compositionality of language to generate accurate descriptions in unseen scenarios. Evaluations of various image captioning methods (as shown in TABLE \ref{tab:captioning_comparison}) on the UCM-Captions\cite{qu2016deep} dataset, as conducted by RS-MoE\cite{lin2025rs}, revealed that early mainstream approaches predominantly adopted the classic ``CNN+RNN" architecture (e.g., VLAD+RNN~\cite{lu2017exploring}, mLSTM\cite{krause2016multiplicative}) or incorporated soft/hard-attention mechanisms to enhance visual feature extraction. While these models could generate basic descriptive sentences, they often encountered bottlenecks in metrics based on n-gram overlap, such as BLEU and ROUGE\_L, struggling to capture complex semantic logic. With the rise of the language-centered paradigm, a new generation of models, represented by RSGPT\cite{hu2025rsgpt}, RS-LLaVA\cite{bazi2024rs}, and RS-MoE\cite{lin2025rs}, has gradually come to dominate the field. These methods are no longer limited to simple sequence generation; instead, they utilize LLMs as universal semantic decoders, leveraging their vast knowledge reserves and reasoning capabilities to understand remote sensing imagery. In terms of evolutionary trends, LLM-driven models have achieved marked qualitative improvement across all metrics. This improvement is particularly evident in the CIDEr metric, which better reflects semantic consistency and human perception, where these models demonstrate a significant performance advantage over traditional methods.

\subsubsection{Visual Question Answering}

\begin{figure*}[t]
    \centering
    \includegraphics[width=0.9\textwidth]{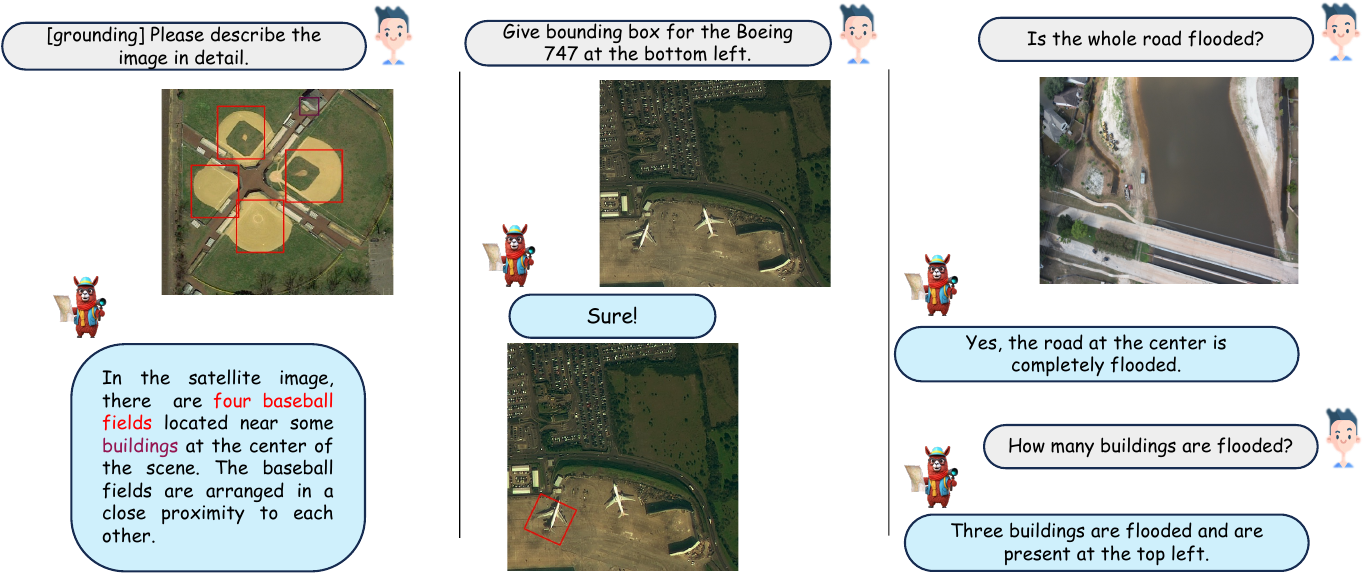}%
    \caption{Example of visual question answer data sourced from \cite{kuckreja2024geochat}}
    \label{vqa_example}
\end{figure*}

VQA further advances remote sensing interpretation from unidirectional, static descriptions to bidirectional, dynamic interactions. It requires models to perform precise information retrieval and reasoning within imagery based on specific questions posed in natural language. As remote sensing application scenarios become increasingly complex, user requirements have shifted from general scene descriptions (captioning) toward target counting, fine-grained attribute analysis, and spatial relationship judgment (as shown in Fig. \ref{vqa_example}). The diversity of VQA tasks intuitively demonstrates the potential of the linguistic modality to integrate complex interpretation tasks. Whether it involves fine-grained grounding of land-cover layouts, precise reference to specific targets based on directional instructions (e.g., ``bottom-left corner"), or logical reasoning and counting regarding ``disaster severity," these taskswhich have traditionally requiring separate specialized modelscan now be unified into a natural language question-answering format. In the face of such demands, relying solely on global feature alignment or simple image-text matching mechanisms (such as the early CLIP\cite{radford2021learning} paradigm) has proven inadequate. The deepening of the language-centered paradigm has propelled a transformation from ``matching-based" models to ``reasoning-based" MLLMs. In these new architectures (e.g., GeoChat\cite{kuckreja2024geochat} and RSGPT\cite{hu2025rsgpt}), language is no longer merely the format of the output but serves as the core carrier of cognition.

\begin{table*}[!t]
\centering
\setlength{\tabcolsep}{3.5pt}
\caption{Evaluation results under the RGB, SAR, and RGB--SAR fused modalities, data sourced from \cite{shu2025earthmind}}
\label{tab:rgb_sar_fused_eval}

\resizebox{\textwidth}{!}{%

\begin{tabular}{l c c c c c c c c c c c c c}
\toprule
\textbf{Model} &
\textbf{Size} &
\textbf{M-Avg} &
\textbf{O-Avg} &

\makecell{\textbf{Scene}\\\textbf{Class.}} &
\makecell{\textbf{Object}\\\textbf{Exist.}} &
\makecell{\textbf{Halluci.}\\\textbf{Detect.}} &
\makecell{\textbf{Object}\\\textbf{Count.}} &
\makecell{\textbf{Spatial}\\\textbf{Relation.}} &
\makecell{\textbf{Referring}\\\textbf{Segmen.}} &

\makecell{\textbf{Image}\\\textbf{Caption}} &
\makecell{\textbf{Disaster}\\\textbf{Forecast.}} &
\makecell{\textbf{Route}\\\textbf{Plann.}} &
\makecell{\textbf{Urban}\\\textbf{Assess.}} \\

\midrule
\textit{Full mark} & -- & 100 & 5 &
100 & 100 & 100 & 100 & 100 & 100 &
5 & 5 & 5 & 5 \\
\midrule
\multicolumn{14}{c}{\textbf{Evaluation on the RGB modality}} \\
\midrule
GPT-4o$^\dagger$ \cite{openai_gpt4o_2024}        & -- & \textbf{70.0} & 2.63 & \textbf{97.3} & 79.9 & \textbf{86.4} & 34.0 & \textbf{52.3} & -- & \textbf{4.58} & 1.75 & \textbf{2.01} & 2.18 \\
GPT-4V$^\dagger$ \cite{openai_gpt4_2023}        & -- & 61.5 & 2.12 & 90.7 & 72.9 & 75.9 & 39.0 & 29.2 & -- & 3.28 & 1.54 & 1.82 & 1.86 \\
GeoChat \cite{kuckreja2024geochat}               & 7B & 41.6 & 1.78 & 71.3 & 51.8 & 46.8 & 18.9 & 19.0 & -- & 1.92 & 1.73 & 1.33 & 2.14 \\
LHRS-bot \cite{muhtar2024lhrs}              & 7B & 47.4 & 1.84 & 76.0 & 58.3 & 58.3 & 25.2 & 19.4 & -- & 2.56 & 1.75 & 1.55 & 1.50 \\
Skysensegpt \cite{luo2024skysensegpt}         & 7B & 45.7 & 1.45 & 77.1 & 56.8 & 56.7 & 26.8 & 11.1 & -- & 1.68 & 1.40 & 1.18 & 1.55 \\
GeoPixel \cite{ou2025geopix}             & 7B & 59.4 & 2.06 & 87.2 & 67.8 & 73.6 & 33.5 & 34.7 & 46.8 & 2.80 & 1.80 & 1.68 & 1.95 \\
EarthMind\cite{shu2025earthmind} & 4B & 69.0 & \textbf{2.82} & 96.5 & \textbf{81.2} & 81.6 & \textbf{47.3} & 38.4 & \textbf{54.0} & 3.35 & \textbf{3.37} & \textbf{2.01} & \textbf{2.55} \\

\midrule
\multicolumn{14}{c}{\textbf{Evaluation on the SAR modality}} \\
\midrule
GPT-4o$^\dagger$ \cite{openai_gpt4o_2024}        & -- & 55.9 & 2.40 & 75.6 & 71.4 & 72.9 & 22.8 & 36.6 & -- & 2.89 & 3.05 & 1.65 &  2.04 \\
GPT-4V$^\dagger$ \cite{openai_gpt4_2023}        & -- & 50.5 & 2.22 & 61.2 & 73.3 & 67.1 & 31.0 & 19.9 & -- & 2.63 & 2.98 & 1.40 & 1.85 \\
GeoChat \cite{kuckreja2024geochat}               & 7B & 40.2 & 1.45 & 58.1 & 49.8 & 46.8 & 27.9 & 18.5 & -- & 1.78 & 1.65 & 1.25 & 1.56 \\
LHRS-bot \cite{muhtar2024lhrs}              & 7B & 41.0 & 1.71 & 58.5 & 56.3 & 48.5 & 23.5 & 18.1 & -- & 1.86 & 1.70 & 1.45 & 1.83 \\
Skysensegpt \cite{luo2024skysensegpt}         & 7B & 39.9 & 1.55 & 51.9 & 53.8 & 49.2 & 33.8 & 10.7 & -- & 1.78 & 1.70 & 1.35 & 1.38 \\
GeoPixel \cite{ou2025geopix}             & 7B & 53.0 & 1.80 & 76.8 & 59.0 & 65.7 & 30.5 & 32.8 & 35.9 & 2.08 & 1.97 & 1.45 & 1.68 \\
EarthMind \cite{shu2025earthmind}                           & 4B & \textbf{67.5} & \textbf{2.64} & \textbf{95.4} & \textbf{77.4} & \textbf{74.6} & \textbf{46.8} & \textbf{43.1} & \textbf{53.0} & \textbf{3.10} & \textbf{3.25} & \textbf{1.89} & \textbf{2.30} \\

\midrule
\multicolumn{14}{c}{\textbf{Evaluation on the RGB--SAR fused modality}} \\
\midrule
GPT-4o$^\dagger$ \cite{openai_gpt4o_2024}        & -- & 67.7 & 2.28 & \textbf{97.7} & 79.6 & \textbf{86.2} & 31.6 & \textbf{43.5} & -- & 3.68 & 1.59 & 1.82 & 2.03 \\
GPT-4V$^\dagger$ \cite{openai_gpt4_2023}        & -- & 58.2 & 1.93 & 91.1 & 64.8 & 62.4 & 32.8 & 39.8 & -- & 2.89 & 1.48 & 1.57 & 1.79 \\
EarthMind \cite{shu2025earthmind} & 4B & \textbf{70.6} & \textbf{3.02} & \textbf{97.7} & \textbf{82.4} & 85.4 & \textbf{47.3} & 40.3 & \textbf{54.5} & \textbf{3.80} & \textbf{3.37} & \textbf{2.21} & \textbf{2.70} \\

\bottomrule
\end{tabular}

}
\end{table*}

The core of this language-centered design philosophy lies in utilizing LLMs as a universal cognitive interface, unifying interpretation requirements across different dimensions into a single sequence generation task. Based on this unified paradigm, language-centered models demonstrate the potential to surpass traditional models in two key dimensions: the expansion of multitask formats and the unification of heterogeneous data (multimodal/multiresolution).

First, with respect to the expansion of multitask formats, the linguistic modality empowers models to handle highly heterogeneous tasks through a single interface. Traditional interpretation systems typically require independent classification or regression heads for ``presence detection," ``attribute classification," or ``comparative counting." In contrast, language-centered models normalize these tasks into a question-answering format via instructions. The experimental results from EarthDial\cite{soni2025earthdial} (as shown in TABLE \ref{tab:presence_comp}) provide strong empirical support for this difference. In the RSVQA-LR\cite{rsvqa} benchmark, which covers diverse question types, general-purpose models (such as MiniGPTv2\cite{chen2023minigpt}) achieved an average accuracy of only 54.96\% because of a lack of specialized task adaptation, performing particularly poorly on the fine-grained rural/urban (R/U) classification task (39.00\%). Conversely, by constructing multitask instruction-tuning data, EarthDial\cite{soni2025earthdial} successfully activated the adaptability of the LLM to different tasks. It not only increased the average accuracy to 92.70\% but also achieved balanced high performance (above 92\%) across three vastly different subtasks: presence, comparison, and R/U classification. This finding demonstrates that the language-centered training mode effectively shatters barriers between tasks, endowing the model with ``versatile" capabilities to switch flexibly among various interpretation requirements.

Second, with respect to the unification of heterogeneous data, language demonstrates its dual value as both a ``semantic anchor" and a ``computational controller," enabling the deep integration of cross-modal data and cross-scale spatial resolution.

\begin{itemize}
    \item {\bf{Unification of Multimodal Data}}: The remote sensing field faces significant challenges because of the stark physical disparities between modalities such as optical (RGB) and synthetic aperture radar (SAR), which traditional methods struggle to integrate. EarthMind\cite{shu2025earthmind} demonstrates the pivotal role of language in overcoming this barrier. By leveraging language as a unified semantic representation layer, EarthMind\cite{shu2025earthmind} maps features from diverse modalities into a shared space, effectively bridging the modality gap. The experimental results (as shown in TABLE \ref{tab:rgb_sar_fused_eval}) revealed that while unaligned VLMs (e.g., GeoChat\cite{kuckreja2024geochat}) achieved a mean average score of only 40.2\% on SAR data, EarthMind\cite{shu2025earthmind}'s language-centered approach increased to 67.5\%, even outperforming GPT-4o\cite{hurst2024gpt} in RGB-SAR fusion tasks (70.6\% vs. 67.7\%). These findings confirm that this paradigm enables systems to transcend single-modality vision, transforming heterogeneous perceptual information into unified linguistic descriptions for the integrated understanding and reasoning of multisource data.

\begin{table*}[t]
\centering
\setlength{\tabcolsep}{5pt}
\caption{Comparison of vision--language models on LRS benchmarks, data sourced from \cite{luo2025large}}
\label{tab:lrs_leaderboard}

\begin{tabular}{
l
@{\hspace{4pt}}
c
@{\hspace{6pt}}
c
@{\hspace{6pt}}
c
@{\hspace{8pt}}
c
@{\hspace{4pt}}
c
@{\hspace{4pt}}
c
@{\hspace{4pt}}
c
@{\hspace{4pt}}
c
@{\hspace{4pt}}
c
}

\toprule
\multicolumn{1}{c}{\textbf{Leaderboard}} &
\multicolumn{1}{c}{\textbf{Data}} &
\multicolumn{1}{c}{\textbf{Vis. Encoder}} &
\multicolumn{1}{c}{\textbf{LLM}} &
\multicolumn{1}{c}{\textbf{Max Size}} &
\multicolumn{1}{c}{\makecell{\textbf{MME-}\\\textbf{RW-RS}}} &
\multicolumn{1}{c}{\makecell{\textbf{LRS-}\\\textbf{FAIR}}} &
\multicolumn{1}{c}{\makecell{\textbf{LRS-}\\\textbf{Bridge}}} &
\multicolumn{1}{c}{\makecell{\textbf{LRS-}\\\textbf{STAR}}} &
\multicolumn{1}{c}{\makecell{\textbf{Avg.}\\\textbf{Acc.}}} \\
\midrule

Qwen2-VL \cite{wang2024qwen2}      & --    & QwenViT\cite{wang2024qwen2} & Qwen2-7B\cite{bai2023qwen}   & 3,333$\times$3,333 & 49.73 & 23.80 & 38.12 & 27.87 & 34.88 \\
LLaVA-OV \cite{li2024llavaOnvision}      & 4.8M  & SigLIP\cite{zhai2023sigmoid}  & Qwen2-7B   & 2,304$\times$2,304 & 53.53 & 20.61 & 35.11 & 26.08 & 33.83 \\
IXC-2.5 \cite{IXC-2.5}         & --    & CLIP-L14\cite{radford2021learning} & InternLM2-7B\cite{cai2024internlm2} & 4,096$\times$4,096 & 36.12 & 25.25 & 38.41 & 27.30 & 31.77 \\
LLaVA-UHD-v2 \cite{LLaVA-UHD} & 1.42M & CLIP-L14+JBU & Vicuna-1.5-7B\cite{peng2023instruction} & 6,721$\times$1,008 & 40.77 & 22.82 & 32.57 & 26.08 & 30.56 \\
LLaVA-FlexAttn \cite{li2024flexattention} & 1.22M & CLIP-L14 & Vicuna-1.5-7B & 1,008$\times$1,008 & 37.75 & 19.57 & 29.99 & 22.76 & 27.52 \\
SEAL\cite{wu2024v}              & 0.95M & CLIP-L14 & Vicuna-1.5-7B & --                & 30.55 & 21.29 & 34.75 & 21.29 & 26.97 \\
MGM-HD\cite{li2024mini}           & 3.0M  & Mixed    & Vicuna-1.5-7B & 1,536$\times$1,536 & 31.51 & 17.90 & 35.92 & 20.13 & 26.36 \\
SliME\cite{zhang2024beyond}            & 2.0M  & CLIP-L14 & Vicuna-1.5-7B & 672$\times$1,008   & 28.54 & 17.11 & 32.09 & 22.99 & 25.18 \\
LLaVA-1.5\cite{liu2024LLaVA1.5}     & 1.22M & CLIP-L14 & Vicuna-1.5-7B & 336$\times$336     & 26.38 & 18.76 & 30.70 & 22.63 & 24.62 \\
\midrule
RSUni-VLM \cite{liu2024rsunivlm}  & 1.2M & SigLIP     & Qwen2.0-5B   & 336$\times$336 & 25.39 & 21.02 & 32.61 & 24.72 & 25.93 \\
GeoChat \cite{kuckreja2024geochat}    & 1.53M& CLIP-L14   & Vicuna-1.5-7B & 504$\times$504 & 28.62 & 20.18 & 24.54 & 13.75 & 21.77 \\
\midrule
GPT-4o\cite{openai_gpt4o_2024}              & -- & -- & -- & -- & 28.92 & 22.15 & 31.84 & 27.40 & 27.58 \\
GPT-4o-mini\cite{openai_gpt4o_2024}     & -- & -- & -- & -- & 6.69  & 18.67 & 31.99 & 25.85 & 20.80 \\
Claude-3.5-Sonnet\cite{anthropic2024claude} & -- & -- & -- & -- & 25.74 & 12.95 & 26.69 & 13.29 & 19.67 \\
\midrule
\midrule
\multicolumn{10}{l}{\textbf{Comparison}} \\
\midrule
LLaVA-1.5*      & 1.04M    & CLIP-L14     & Vicuna-1.5-7B & 336$\times$336 & 34.40 & 18.24 & 32.28 & 24.17 & 27.30 \\
LLaVA-1.5-p4*   & 1.04M    & CLIP-L14     & Vicuna-1.5-7B & 672$\times$672 & 38.20 & 19.18 & 35.43 & 26.50 & 29.83 \\
GeoChat*       & 1.04M & CLIP-L14     & Vicuna-1.5-7B & 504$\times$504 & 33.68 & 21.51 & 35.97 & 25.86 & 29.25 \\
SliME*           & 1.04M    & CLIP-L14     & Vicuna-1.5-7B & 672$\times$1,008 & 34.56 & 21.98 & 33.20 & 25.10 & 28.71 \\
LLaVA-UHD-v2* & 1.04M    & CLIP-L14+JBU  & Vicuna-1.5-7B & 672$\times$1,008 & 38.55 & 20.77 & 36.57 & 25.79 & 30.42 \\
DIP\cite{luo2025large} (LLaVA-1.5)           & 1.04M    & CLIP-L14      & Vicuna-1.5-7B            & Dynamic        & \textbf{39.04} & \textbf{22.97} & \textbf{36.89} & \textbf{27.48} & \textbf{31.59} \\
\midrule
LLaVA-Next-p4*\cite{li2024llavanext-strong}     & 1.04M    & CLIP-L14 & Qwen2-7B & 672$\times$672   & 39.12 & 21.06 & 36.27 & 25.80 & 30.56 \\
LLaVA-Next-p8*     & 1.04M & CLIP-L14 & Qwen2-7B & 1,008$\times$1,008 & 40.35 & 21.14 & 37.25 & 26.10 & 31.21 \\
LLaVA-Next-p2.5*   & 1.04M    & CLIP-L14 & Qwen2-7B & 1,680$\times$1,680 & 39.65 & 20.99 & 36.38 & 26.18 & 30.80 \\
DIP\cite{luo2025large}(LLaVA-Next)  & 1.04M    & CLIP-L14  & Qwen2-7B       & Dynamic          & \textbf{41.89} & \textbf{21.85} & \textbf{38.24} & \textbf{26.67} & \textbf{32.16} \\

\bottomrule
\end{tabular}
\end{table*}

    \item {\bf{Unification of Spatial Resolution}}: Research trends are shifting from passive ``static pyramids" toward active ``language-guided dynamic perception." In addressing the vast-scale variations inherent in remote sensing imagery, traditional visual-centered models often rely on fixed sliding windows or static image pyramids. While increasing input resolution can result in marginal gains, as shown in TABLE \ref{tab:lrs_leaderboard}, scaling LLaVA-1.5\cite{liu2024LLaVA1.5} from 336$\times$336 to a fixed 672$\times$672 resolution improves the average score from 27.30\% to 29.83\%. Such ``passive" processing results in significant computational redundancy and still risks losing fine-grained targets. In contrast, the introduction of the language modality empowers models to act as ``computational controllers." Recent innovations, such as the integration of text-guided token pruning with a dynamic image pyramid (DIP)\cite{luo2025large}, utilize instructions (e.g., ``detect small vessels") as prior signals to guide the visual encoder. By performing coarse-to-fine dynamic pruning, this method retains only visual tokens relevant to the textual intent. The experimental results demonstrate the efficacy of this paradigm: DIP\cite{luo2025large} (LLaVA-1.5\cite{liu2024LLaVA1.5}) achieves an average accuracy of 31.59\%, significantly outperforming both the fixed high-resolution baseline (29.83\%) and specialized high-res models such as LLaVA-UHD-v2\cite{LLaVA-UHD} (30.42\%). Furthermore, when applied to stronger backbones such as LLaVA-Next\cite{li2024llavanext-strong}, the dynamic approach reaches a peak performance of 32.16\%, surpassing the static 1,008$\times$1,008 resolution setting (31.21\%). This mechanism not only reduces computational complexity while preserving critical details but also reflects a fundamental evolution from a ``brute-force" mode of processing all pixels to an ``attention-on-demand" active perception paradigm.
    
\end{itemize}

\begin{figure*}[!t]
    \centering
    \includegraphics[width=\textwidth]{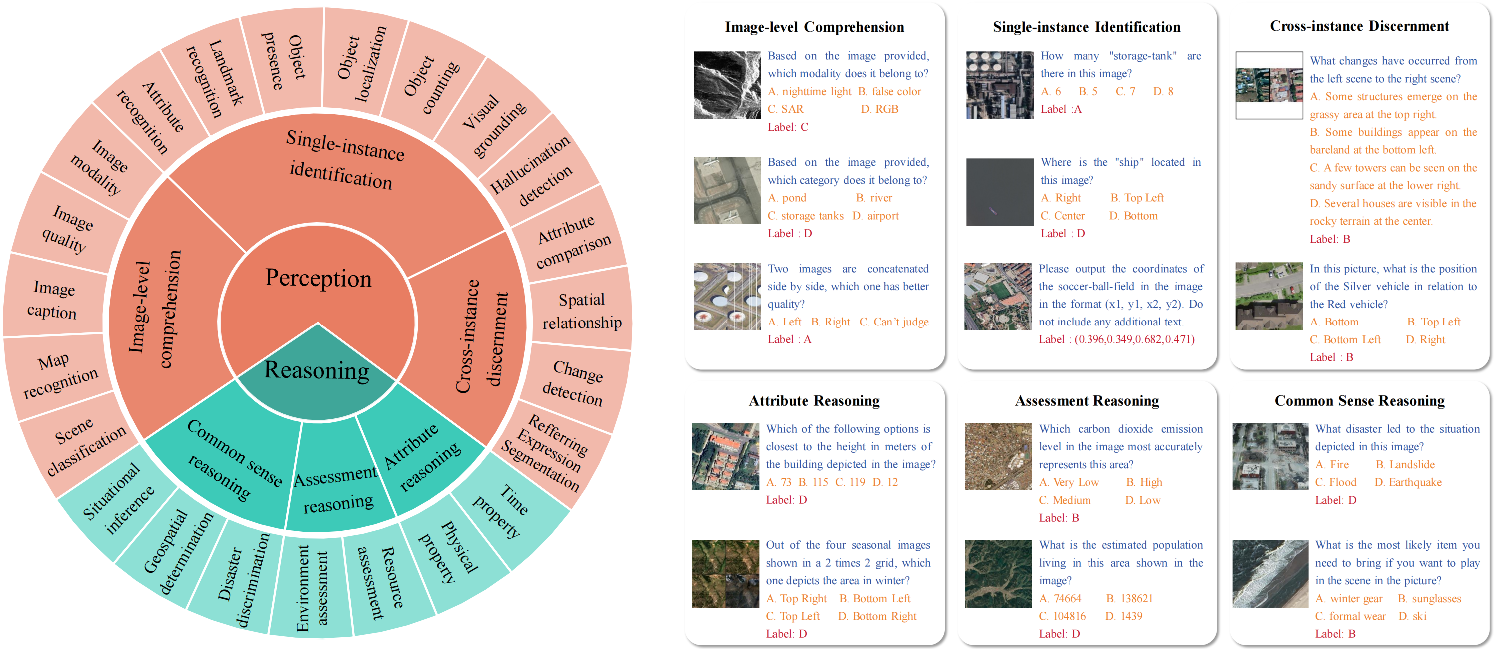}%
    \caption{Example of complex reasoning dataset design, sourced from \cite{an2024choice}}
    \label{cr_example}
\end{figure*}

In summary, the language-centered training paradigm renders ``general-purpose remote sensing" interpretation a reality. By providing a unified task interface, it eliminates the problem of model fragmentation inherent in task-specific designs. Furthermore, through a unified semantic space, it overcomes the dual heterogeneity of remote sensing data in terms of both sensor modalities and spatial resolution. Ultimately, this paradigm establishes a robust foundation for building the next generation of remote sensing interpretation systems characterized by multitask versatility and all-data adaptability.

\subsubsection{Complex Reasoning}
Complex reasoning represents the highest-order task in the cognitive leap of remote sensing interpretation from perception to cognition. Unlike VQA, which relies primarily on direct matching of internal visual cues (such as color, shape, and quantity), complex reasoning tasks are often open-ended and nonintuitive. Taking the problem settings in the CHOICE \cite{an2024choice} benchmark as an example (as shown in Fig. \ref{cr_example}), the answers to such questions typically cannot be retrieved directly from images. Instead, they require the integration of visual features with external domain knowledge, such as geographic principles, physical attributes, and spatiotemporal contexts, followed by multistep logical deduction. MLLMs face severe challenges when addressing such tasks. Without the support of external knowledge bases and rigorous logical verification mechanisms, these models are highly prone to hallucinations generating semantically fluent but factually incorrect conclusions that violate geographic realities. To overcome this bottleneck, the deepening of the language-centered paradigm is further propelling the evolution of interpretation systems toward LLM-based agents.

In this new architecture, the LLM is no longer merely a generator but has been upgraded to a central controller. By utilizing language as CoT\cite{wei2022chain}, it autonomously plans reasoning paths and dynamically invokes external tools (e.g., search engines, physical simulators, and specialized knowledge bases) for verification. Experimental results demonstrate that this agent-based interpretation mode not only compensates for the specialized knowledge deficiencies of standalone MLLMs but also significantly enhances the logical completeness and trustworthiness of the model when solving complex real-world problems through a ``perception-planning-action" closed loop.

As shown in the TABLE \ref{tab:rs_agent}, in the RSVQA-LR\cite{rsvqa} benchmark, we observe that general-purpose MLLMs (e.g., LLaVA-1.5\cite{liu2024LLaVA1.5} and MiniGPTv2\cite{chen2023minigpt}) are constrained by their static parametric knowledge, making it difficult for them to handle domain-specific tasks requiring precise judgment. Their average accuracy is approximately 60\% (59.8\% for MiniGPTv2). In contrast, by employing the LLM as a controller to dynamically invoke perception tools, RS-Agent\cite{xu2024rs} achieved an average accuracy of 90.9\%, significantly surpassing that of all general-purpose models. Notably, in the highly challenging ``rural/urban" (R/U) classification task, RS-Agent reached a remarkable accuracy of 97.0\%, while the contemporary general model Qwen-VL\cite{wang2024qwen2} achieved only 62.0\% accuracy. This massive performance gap underscores the necessity of transforming the LLM from a ``standalone interpreter" into a ``tool-using agent" to achieve precise anchoring of complex geographic attributes.

\begin{table*}[!t]
\centering

\caption{Results on the RS-Agent benchmark (\%), data sourced from \cite{xu2024rs}.}
\label{tab:rs_agent}
\begin{tabular}{lcccc}
\toprule
\textbf{Method} & \textbf{Rural/Urban} & \textbf{Presence} & \textbf{Compare} & \textbf{Avg.} \\
\midrule
LLaVA-15 \cite{liu2024LLaVA1.5}        & 59.2 & 73.2 & 65.2 & 65.9 \\
MiniGPTv2 \cite{chen2023minigpt}     & 60.0 & 51.6 & 67.6 & 59.8 \\
InstructBLIP \cite{li2022blip}   & 62.6 & 48.8 & 63.9 & 59.1 \\
mPLUG-Owl2 \cite{dai2023instructblip}      & 58.0 & 74.0 & 65.0 & 65.7\% \\
QWen-VL-Chat \cite{bai2023qwen}   & 62.0 & 47.7 & 54.6 & 58.7 \\
\midrule
RSVQA \cite{lobry2020rsvqa}        & 90.0 & 87.5 & 81.5 & 86.3 \\
SkyEyeGPT \cite{zhan2025skyeyegpt}     & 88.9 & 88.6 & 75.0 & 84.2 \\
LHRS-Det \cite{muhtar2024lhrs}    & 89.1 & 88.5 & 90.0 & 89.2 \\
GeoChat \cite{kuckreja2024geochat}   & 94.0 & 91.1 & 90.3 & 90.7 \\
\midrule
\textbf{RS-Agent\cite{xu2024rs}}          & \textbf{97.0} & \textbf{91.1} & \textbf{90.6} & \textbf{90.9} \\
\bottomrule
\end{tabular}
\end{table*}

\begin{table*}[!t]
\centering
\setlength{\tabcolsep}{4pt}
\caption{Performance comparison on geospatial agent benchmarks, data sourced from \cite{stamoulis2025geo}.}
\label{tab:geo_agent}

\begin{tabular}{l c c c c c c c c c c}
\toprule
\textbf{Model} &
\makecell{\textbf{Geospatial}\\\textbf{Agent}} &
\makecell{\textbf{Chat}\\\textbf{API}} &
\makecell{\textbf{Correct}\\\textbf{Rate}\%} &
\makecell{\textbf{Success}\\\textbf{Rate}\%} &
\makecell{\textbf{Tokens}\\\textbf{Avg (k)}} &

\makecell{\textbf{Agro}\\$\boldsymbol{\epsilon}\%$} &
\makecell{\textbf{Climate}\\$\boldsymbol{\epsilon}\%$} &
\makecell{\textbf{Urban}\\$\boldsymbol{\epsilon}\%$} &
\makecell{\textbf{Forest}\\$\boldsymbol{\epsilon}\%$} &
\makecell{\textbf{Vision}\\\textbf{R}\%} \\
\midrule

GPT-o1\cite{jaech2024openai} & GeoLLM-Squad\cite{lee2025multi} & $\checkmark$ & 90.97 & 97.2 & 77.36 & 0.14 & 0.58 & 1.39 & 0.56 & 98.71 \\
\midrule

\multirow{6}{*}{GPT-4o\cite{hurst2024gpt}}
 & GeoLLM-QA\cite{singh2024evaluating}     & $\checkmark$ & 52.49 & 62.3 & 22.78  & 5.34 & 3.97 & 6.33 & 6.73 & 72.80 \\
 & GeoLLM-Engine\cite{singh2024geollm} & $\checkmark$ & 74.11 & 82.6 & 22.25  & 5.05 & 2.26 & 8.31 & 7.85 & 87.71 \\
 & Chameleon\cite{lu2023chameleon}     & $\checkmark$ & 21.87 & 55.1 & 36.19 & 12.51& 9.14 & 9.62 & 10.77& 50.42 \\
 & Magentic\cite{fourney2024magentic}      & $\checkmark$ & 25.40 & 79.7 & 117.65 & 83.89& 80.63& 80.86& 83.13& 97.25 \\
 & GeoLLM-Squad\cite{lee2025multi}  & $\checkmark$ & 85.32 & 96.2 & 77.49 & 3.70 & 2.02 & 1.99 & 3.33 & 97.59 \\
 \cmidrule(lr){2-11}
 & Geo-OLMs\cite{stamoulis2025geo}      & $\checkmark$ & 86.03 & 95.1 & 30.44 & 3.44 & 1.17 & 4.40 & 2.51 & 96.26 \\
\midrule

\multirow{6}{*}{GPT-4o-mini\cite{hurst2024gpt}}
 & GeoLLM-QA\cite{singh2024evaluating}     & $\checkmark$ & 37.41 & 53.1 & 21.55 & 13.05& 10.29& 14.61& 12.82& 69.16 \\
 & GeoLLM-Engine\cite{singh2024geollm} & $\checkmark$ & 53.56 & 82.8 & 20.09 & 11.45& 8.69 & 8.88 & 8.59 & 60.33 \\
 & Chameleon\cite{lu2023chameleon}     & $\checkmark$ & 34.63 & 68.7 & 60.10 & 12.25& 8.05 & 8.85 & 8.84 & 67.42 \\
 & Magentic\cite{fourney2024magentic}      & $\checkmark$ & 18.85 & 73.1 & 187.45 & 13.44& 11.45& 11.54& 12.29& 75.51 \\
 & GeoLLM-Squad\cite{lee2025multi}  & $\checkmark$ & 59.46 & 87.7 & 72.98 & 16.42& 9.70 & 10.22& 10.19& 70.58 \\
 \cmidrule(lr){2-11}
 & Geo-OLMs \cite{stamoulis2025geo}     & $\checkmark$ & 58.75 & 85.4 & 28.49 & 15.66& 10.45& 9.75 & 8.63 & 75.49 \\
\bottomrule
\end{tabular}
\end{table*}

As research deepens, this trend is further evolving from ``single-agent tool invocation" toward multiagent collaboration, with the LLM serving as the cognitive hub. When interdisciplinary geographic issues involving agriculture, climate, and urban planning are addressed, a single agent often struggles to encompass expert knowledge from all fields. The experimental data in TABLE \ref{tab:geo_agent} highlight the immense potential of this ``cognitive hub" paradigm. A comparison of a GPT-4o-based\cite{hurst2024gpt} single agent (GeoLLM-QA\cite{singh2024evaluating}) with a multiagent squad (GeoLLM-Squad\cite{lee2025multi}) revealed that the single agent, possessing only QA capabilities, achieved a correctness rate of only 52.49\%. Conversely, when the LLM was used as a commander to orchestrate agents with different specialties (e.g., Agro, Climate, and Urban experts), the correctness rate of the GeoLLM-Squad\cite{lee2025multi}surged to 85.32\%, with a task success rate as high as 96.2\%. Furthermore, when the more reasoning-capable GPT-o1\cite{jaech2024openai} was introduced as the cognitive core, the accuracy of the GeoLLM-Squad\cite{lee2025multi} further exceeded 90.97\%.

These progressive experimental results reveal the future trend of remote sensing interpretation: language models are evolving into a universal cognitive hub. By flexibly associating external knowledge bases, professional tools, and even other agents, they enable collaborative and deep reasoning for complex geographic problems in the open world.

\section{Future Directions and Challenges}

\begin{figure*}[!t]
    \centering
    \includegraphics[width=0.85\textwidth]{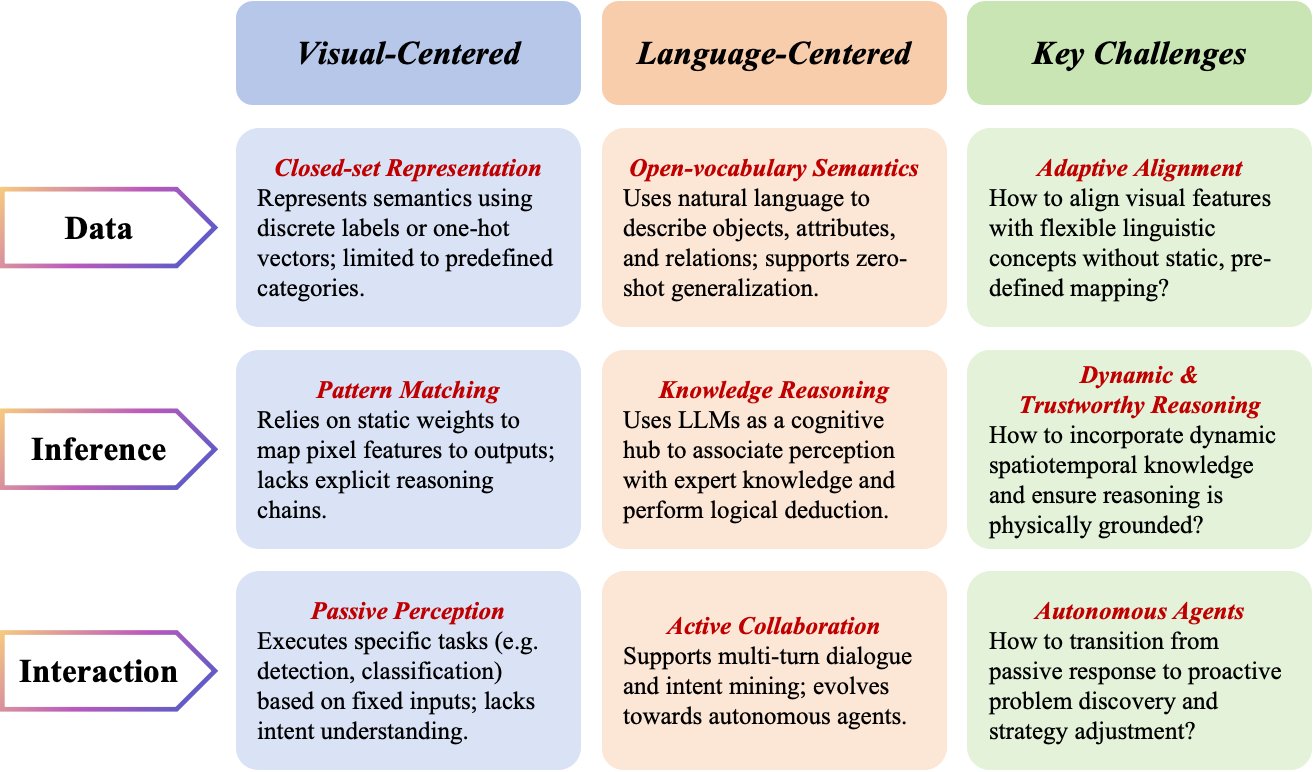}%
    \caption{Comparative analysis of visual-centered and language-centered paradigms}
    \label{compare_v2l}
\end{figure*}

To clarify the evolutionary significance of the language-centered remote sensing interpretation paradigm and to delineate its future trajectory, this section begins with a systematic comparison between the ``visual-centered" and ``language-centered" paradigms across three dimensions: data, model, and interaction (as summarized in Fig. \ref{compare_v2l}). Based on the distinctions between these two paradigms, we subsequently provide an outlook on the core future research directions.

\subsection{Comparative Analysis of Paradigms: From Vision to Language}

Currently, remote sensing interpretation is in a pivotal stage of transitioning from a visual-centered to a language-centered paradigm. Examining this shift reveals that the language-centered paradigm has reconstructed the traditional vision paradigm at its fundamental logic level:

\begin{itemize}

    \item Data Perspective: The visual-centered paradigm relies on closed-set discrete labels, which constrain the model's ability to represent long-tail concepts and fine-grained attributes in the open world. In contrast, the language-centered paradigm introduces continuous, semantically rich open-vocabulary semantics, enabling interpretation systems to handle unseen categories and complex spatial relationships. A challenge for future research lies in achieving adaptive alignment between visual features and linguistic semantics, moving beyond reliance on static annotation mappings.

    \item Inference Mechanism: Vision models are typically task specific; they rely on static weights for pattern matching but lack the capacity to invoke external knowledge. Conversely, the language-centered paradigm uses the LLM as a cognitive hub, possessing universal reasoning and knowledge association capabilities. However, this design introduces new challenges: how to integrate static parametric knowledge with dynamic spatiotemporal environments and how to ensure that the reasoning process is trustworthy and compliant with physical laws.

    \item Interaction Perspective: Traditional paradigms are passive and merely output results based on fixed instructions. The language-centered paradigm achieves human-machine alignment through natural language interfaces and is further evolving into autonomous agents with planning capabilities. This evolution requires future systems to not only understand instructions but also proactively identify problems and make autonomous decisions.
    
\end{itemize}

\subsection{Future Research Directions}

Based on the aforementioned comparative analysis, while the language-centered paradigm demonstrates immense potential, significant challenges remain in achieving true ``cognition-driven" interpretation. The following four directions merit focused exploration.

\subsubsection{Adaptive Alignment for Multimodal Remote Sensing Data}Although existing models incorporate the language modality, the alignment of nonlanguage modalities (e.g., SAR, hyperspectral) to the language space typically relies on preset static mappings or large-scale manual annotations. This rigid alignment limits the model's flexibility when facing multigranular semantic requirements. Human cognition suggests an adaptive ability to infer from examples: humans do not need to exhaust all vision-language pairs but can dynamically match visual targets in different contexts based on abstract concepts. Future research should strive to endow models with similar reasoning-driven alignment, utilizing the powerful semantic reasoning of LLMs to back-derive possible visual feature combinations from abstract conceptual spaces and perform dynamic matching with perceptual information. This shift from ``static mapping" to ``reasoning alignment" will enable models to adapt more flexibly to complex, open-world interpretation tasks.

\subsubsection{Task Understanding under Spatiotemporal Dynamics and Knowledge Constraints} From a model perspective, most existing language-centered methods are based on static spatiotemporal assumptions and overlook the inherently dynamic nature of remote sensing tasks. As time progresses and spatial scales shift, the characteristics and interrelationships of ground objects change significantly. However, the parametric knowledge within LLMs is often static, making it difficult to adapt to such environmental evolutions in real time and leading to ``knowledge obsolescence" or reasoning biases. A future breakthrough will concern  the introduction of a human-like ``dynamic memory mechanism." Inspired by the memory mechanisms of the hippocampus, researchers should explore the construction of dynamically updated knowledge graphs or external memory modules. This approach would allow interpretation systems to continuously integrate new observational data and expert feedback during interaction, achieving iterative knowledge updates. Such an approach will moreover resolve the contradiction between the static knowledge of large models and the dynamic environments of remote sensing, enhancing the accuracy of long-term monitoring tasks.

\subsubsection{Trustworthy Reasoning and Adaptive Decision-Making for Complex Tasks} While MLLMs enhance reasoning capabilities, existing models still face ``hallucination" risks in high-stakes fields, where reasoning processes are logically fluent but violate geophysiological facts. Furthermore, current reasoning is largely internal self-validation, lacking a feedback loop with the real physical environment, which may result in nonexecutable decisions. Future research must establish a ``physically trustworthy" reasoning framework. On the one hand, remote sensing physical mechanisms should be injected as hard constraints into the language model's reasoning chain to reduce factual errors. On the other hand, models should be endowed with adaptive decision-making capabilities, allowing them to dynamically adjust their perception strategies (Actions) based on external environmental feedback (e.g., cloud cover, sensor availability) rather than remaining confined to text-level reasoning.

\subsubsection{Autonomous Interactive Interpretation Agents} In the interaction dimension, current systems remain largely in a ``passive response" stage, relying on explicit user instructions. This situation represents a fundamental gap from the workflow of professional analysts, who can proactively scrutinize imagery, identify potential anomalies (e.g., unrequested landslide precursors), and autonomously formulate analysis plans. The ultimate form of remote sensing interpretation is the construction of autonomous remote sensing agents. Future systems should not merely be ``question-answering machines" but should possess ``proactivity": the ability to autonomously mine potential interpretation needs based on environmental context and domain knowledge, propose hypotheses, and invoke tools for verification. This paradigm shift from ``passive instruction execution" to ``autonomous problem discovery" will be a critical step toward artificial general intelligence (AGI) in remote sensing.

\section{Conclusion}
\begin{figure}[t]
    \centering
    \includegraphics[width=0.45\textwidth]{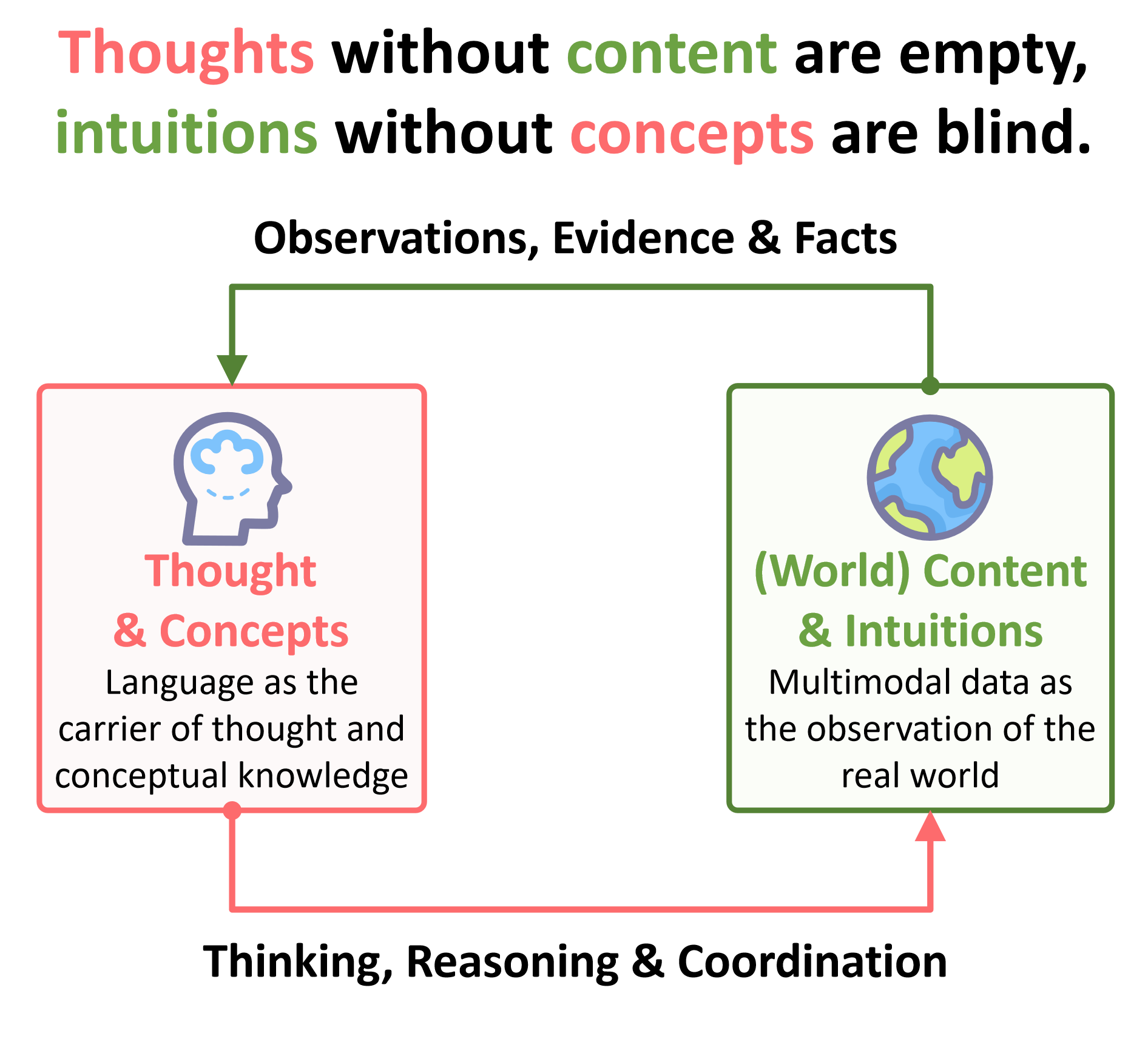}%
    \caption{Schematic Illustration of the Relationship between Intuition and Concepts in Kant Epistemology.}
    \label{conclusion}
\end{figure}

 Fig.\ref{conclusion} illustrates the epistemological relationship described by Kant \cite{preussischen1902akademie}, where visual perception (``intuitions") requires the illumination of linguistic concepts to avoid being ``blind." Based on this premise, remote sensing image interpretation technology is undergoing a paradigm shift from ``visual-centered perception" to ``language-centered cognition." This paper traces the dialectical relationship between language and vision in the human cognitive system, revealing the inherent limitations of the current visual-centered paradigm in terms of semantic abstraction, dynamic feedback, and closed-loop cognition. 
 
 Grounded in the GWT of cognitive science, we propose a novel interpretation paradigm that positions language as the central hub for multimodal representation. This framework not only reconstructs the full perception--cognition--decision loop but also empowers interpretation systems with cross-modal semantic association and interpretable reasoning through the referential and logical predicate functions of language. Compared with the visual-centered paradigm, the language-centered framework presents three transformative advantages: The natural language interface for task understanding eliminates barriers in human-machine interaction. The integration of dynamic knowledge graphs overcomes the limitations of perception constrained by ``what is seen is what is known." The cyclic interaction mechanism of the cognitive workspace enables a qualitative leap from static interpretation to autonomous decision-making. While current technologies still face challenges in multimodal representation alignment and spatiotemporal knowledge evolution modeling, the advancement of a language-centered cyclic interpretation paradigm holds promise for overcoming the instrumental limitations of existing systems, steering remote sensing interpretation toward becoming a cognitive agent capable of self-evolution. Future research should focus on deep coupling between language, vision, and physical space, constructing dynamically evolving cognitive schemata under spatiotemporal constraints. Ultimately, this effort will enable a leap from data-driven to emergent cognitive intelligence in remote sensing, creating a new paradigm of cognitive computing where ``language reconstructs the world."

\bibliographystyle{IEEEtran}
\bibliography{references}

@article{jay1988rise,
  title={The rise of hermeneutics and the crisis of ocularcentrism},
  author={Jay, Martin},
  journal={Poetics Today},
  volume={9},
  number={2},
  pages={307--326},
  year={1988},
  publisher={JSTOR}
}

@article{preussischen1902akademie,
  title={Akademie der Wissenschaften},
  author={Preussischen, K{\"o}niglich},
  journal={Kant’s gesammelte Schriften},
  year={1902}
}

@inproceedings{sun2017revisiting,
  title={Revisiting unreasonable effectiveness of data in deep learning era},
  author={Sun, Chen and Shrivastava, Abhinav and Singh, Saurabh and Gupta, Abhinav},
  booktitle={Proceedings of the IEEE international conference on computer vision},
  pages={843--852},
  year={2017}
}

@article{kaplan2020scaling,
  title={Scaling laws for neural language models},
  author={Kaplan, Jared and McCandlish, Sam and Henighan, Tom and Brown, Tom B and Chess, Benjamin and Child, Rewon and Gray, Scott and Radford, Alec and Wu, Jeffrey and Amodei, Dario},
  journal={arXiv preprint arXiv:2001.08361},
  year={2020}
}

@inproceedings{radford2021learning,
  title={Learning transferable visual models from natural language supervision},
  author={Radford, Alec and Kim, Jong Wook and Hallacy, Chris and Ramesh, Aditya and Goh, Gabriel and Agarwal, Sandhini and Sastry, Girish and Askell, Amanda and Mishkin, Pamela and Clark, Jack and others},
  booktitle={International conference on machine learning},
  pages={8748--8763},
  year={2021},
  organization={PmLR}
}

@article{bazi2024rs,
  title={Rs-llava: A large vision-language model for joint captioning and question answering in remote sensing imagery},
  author={Bazi, Yakoub and Bashmal, Laila and Al Rahhal, Mohamad Mahmoud and Ricci, Riccardo and Melgani, Farid},
  journal={Remote Sensing},
  volume={16},
  number={9},
  pages={1477},
  year={2024},
  publisher={MDPI}
}

@article{baars2005global,
  title={Global workspace theory of consciousness: toward a cognitive neuroscience of human experience},
  author={Baars, Bernard J},
  journal={Progress in brain research},
  volume={150},
  pages={45--53},
  year={2005},
  publisher={Elsevier}
}

@article{touvron2023llama,
  title={Llama: Open and efficient foundation language models},
  author={Touvron, Hugo and Lavril, Thibaut and Izacard, Gautier and Martinet, Xavier and Lachaux, Marie-Anne and Lacroix, Timoth{\'e}e and Rozi{\`e}re, Baptiste and Goyal, Naman and Hambro, Eric and Azhar, Faisal and others},
  journal={arXiv preprint arXiv:2302.13971},
  year={2023}
}

@article{bi2024deepseek,
  title={Deepseek llm: Scaling open-source language models with longtermism},
  author={Bi, Xiao and Chen, Deli and Chen, Guanting and Chen, Shanhuang and Dai, Damai and Deng, Chengqi and Ding, Honghui and Dong, Kai and Du, Qiushi and Fu, Zhe and others},
  journal={arXiv preprint arXiv:2401.02954},
  year={2024}
}

@article{liu2024rsunivlm,
  title={Rsunivlm: A unified vision language model for remote sensing via granularity-oriented mixture of experts},
  author={Liu, Xu and Lian, Zhouhui},
  journal={arXiv preprint arXiv:2412.05679},
  year={2024}
}

@article{wang2024qwen2,
  title={Qwen2-vl: Enhancing vision-language model's perception of the world at any resolution},
  author={Wang, Peng and Bai, Shuai and Tan, Sinan and Wang, Shijie and Fan, Zhihao and Bai, Jinze and Chen, Keqin and Liu, Xuejing and Wang, Jialin and Ge, Wenbin and others},
  journal={arXiv preprint arXiv:2409.12191},
  year={2024}
}

@article{li2024llavaOnvision,
  title={LLaVA-OneVision: Easy Visual Task Transfer},
  author={Li, Bo and Zhang, Yuanhan and Guo, Dong and Zhang, Renrui and Li, Feng and Zhang, Hao and Zhang, Kaichen and Li, Yanwei and Liu, Ziwei and Li, Chunyuan},
  journal={CoRR},
  year={2024}
}

@inproceedings{li2024flexattention,
  title={Flexattention for efficient high-resolution vision-language models},
  author={Li, Junyan and Chen, Delin and Cai, Tianle and Chen, Peihao and Hong, Yining and Chen, Zhenfang and Shen, Yikang and Gan, Chuang},
  booktitle={European Conference on Computer Vision},
  pages={286--302},
  year={2024},
  organization={Springer}
}

@article{LLaVA-UHD,
  title={LLaVA-UHD v2: an MLLM Integrating High-Resolution Feature Pyramid via Hierarchical Window Transformer},
  author={Zhang, Yipeng and Liu, Yifan and Guo, Zonghao and Zhang, Yidan and Yang, Xuesong and Chen, Chi and Song, Jun and Zheng, Bo and Yao, Yuan and Liu, Zhiyuan and others},
  journal={CoRR},
  year={2024}
}

@article{IXC-2.5,
  title={InternLM-XComposer-2.5: A Versatile Large Vision Language Model Supporting Long-Contextual Input and Output},
  author={Zhang, Pan and Dong, Xiaoyi and Zang, Yuhang and Cao, Yuhang and Qian, Rui and Chen, Lin and Guo, Qipeng and Duan, Haodong and Wang, Bin and Ouyang, Linke and others},
  journal={CoRR},
  year={2024}
}

@article{ou2025geopix,
  title={GeoPix: A multimodal large language model for pixel-level image understanding in remote sensing},
  author={Ou, Ruizhe and Hu, Yuan and Zhang, Fan and Chen, Jiaxin and Liu, Yu},
  journal={IEEE Geoscience and Remote Sensing Magazine},
  year={2025},
  publisher={IEEE}
}

@misc{team2024internvl2,
  title={Internvl2: Better than the best—expanding performance boundaries of open-source multimodal models with the progressive scaling strategy},
  author={Team, OpenGVLab},
  year={2024},
  publisher={Accessed}
}

@article{shu2025earthmind,
  title={EarthMind: Towards Multi-Granular and Multi-Sensor Earth Observation with Large Multimodal Models},
  author={Shu, Yan and Ren, Bin and Xiong, Zhitong and Paudel, Danda Pani and Van Gool, Luc and Demir, Begum and Sebe, Nicu and Rota, Paolo},
  journal={arXiv preprint arXiv:2506.01667},
  year={2025}
}

@inproceedings{clip,
  title={Learning transferable visual models from natural language supervision},
  author={Radford, Alec and Kim, Jong Wook and Hallacy, Chris and Ramesh, Aditya and Goh, Gabriel and Agarwal, Sandhini and Sastry, Girish and Askell, Amanda and Mishkin, Pamela and Clark, Jack and others},
  booktitle={International conference on machine learning},
  pages={8748--8763},
  year={2021},
  organization={PmLR}
}

@inproceedings{jia2021scaling,
  title={Scaling up visual and vision-language representation learning with noisy text supervision},
  author={Jia, Chao and Yang, Yinfei and Xia, Ye and Chen, Yi-Ting and Parekh, Zarana and Pham, Hieu and Le, Quoc and Sung, Yun-Hsuan and Li, Zhen and Duerig, Tom},
  booktitle={International conference on machine learning},
  pages={4904--4916},
  year={2021},
  organization={PMLR}
}

@inproceedings{li2022blip,
  title={Blip: Bootstrapping language-image pre-training for unified vision-language understanding and generation},
  author={Li, Junnan and Li, Dongxu and Xiong, Caiming and Hoi, Steven},
  booktitle={International conference on machine learning},
  pages={12888--12900},
  year={2022},
  organization={PMLR}
}

@article{zhu2023languagebind,
  title={Languagebind: Extending video-language pretraining to n-modality by language-based semantic alignment},
  author={Zhu, Bin and Lin, Bin and Ning, Munan and Yan, Yang and Cui, Jiaxi and Wang, HongFa and Pang, Yatian and Jiang, Wenhao and Zhang, Junwu and Li, Zongwei and others},
  journal={arXiv preprint arXiv:2310.01852},
  year={2023}
}

@article{shao2025allspark,
  title={AllSpark: A Multimodal Spatio-Temporal General Intelligence Model with Ten Modalities via Language as a Reference Framework},
  author={Shao, Run and Yang, Cheng and Li, Qiujun and Xu, LinRui and Yang, Xiang and Li, Xian and Li, MengYao and Zhu, Qing and Zhang, Yongjun and Li, YanSheng and others},
  journal={IEEE Transactions on Geoscience and Remote Sensing},
  year={2025},
  publisher={IEEE}
}

@article{anand2024mathify,
  title={Mathify: Evaluating large language models on mathematical problem solving tasks},
  author={Anand, Avinash and Gupta, Mohit and Prasad, Kritarth and Singla, Navya and Sanjeev, Sanjana and Kumar, Jatin and Shivam, Adarsh Raj and Shah, Rajiv Ratn},
  journal={arXiv preprint arXiv:2404.13099},
  year={2024}
}

@article{ahn2024large,
  title={Large language models for mathematical reasoning: Progresses and challenges},
  author={Ahn, Janice and Verma, Rishu and Lou, Renze and Liu, Di and Zhang, Rui and Yin, Wenpeng},
  journal={arXiv preprint arXiv:2402.00157},
  year={2024}
}

@article{chowdhery2023palm,
  title={Palm: Scaling language modeling with pathways},
  author={Chowdhery, Aakanksha and Narang, Sharan and Devlin, Jacob and Bosma, Maarten and Mishra, Gaurav and Roberts, Adam and Barham, Paul and Chung, Hyung Won and Sutton, Charles and Gehrmann, Sebastian and others},
  journal={Journal of Machine Learning Research},
  volume={24},
  number={240},
  pages={1--113},
  year={2023}
}

@article{wei2022chain,
  title={Chain-of-thought prompting elicits reasoning in large language models},
  author={Wei, Jason and Wang, Xuezhi and Schuurmans, Dale and Bosma, Maarten and Xia, Fei and Chi, Ed and Le, Quoc V and Zhou, Denny and others},
  journal={Advances in neural information processing systems},
  volume={35},
  pages={24824--24837},
  year={2022}
}

@article{guo2025deepseek,
  title={Deepseek-r1: Incentivizing reasoning capability in llms via reinforcement learning},
  author={Guo, Daya and Yang, Dejian and Zhang, Haowei and Song, Junxiao and Zhang, Ruoyu and Xu, Runxin and Zhu, Qihao and Ma, Shirong and Wang, Peiyi and Bi, Xiao and others},
  journal={arXiv preprint arXiv:2501.12948},
  year={2025}
}

@article{li2019visualbert,
  title={Visualbert: A simple and performant baseline for vision and language},
  author={Li, Liunian Harold and Yatskar, Mark and Yin, Da and Hsieh, Cho-Jui and Chang, Kai-Wei},
  journal={arXiv preprint arXiv:1908.03557},
  year={2019}
}

@article{tan2019lxmert,
  title={Lxmert: Learning cross-modality encoder representations from transformers},
  author={Tan, Hao and Bansal, Mohit},
  journal={arXiv preprint arXiv:1908.07490},
  year={2019}
}

@inproceedings{chen2020uniter,
  title={Uniter: Universal image-text representation learning},
  author={Chen, Yen-Chun and Li, Linjie and Yu, Licheng and El Kholy, Ahmed and Ahmed, Faisal and Gan, Zhe and Cheng, Yu and Liu, Jingjing},
  booktitle={European conference on computer vision},
  pages={104--120},
  year={2020},
  organization={Springer}
}

@article{hu2025rsgpt,
  title={Rsgpt: A remote sensing vision language model and benchmark},
  author={Hu, Yuan and Yuan, Jianlong and Wen, Congcong and Lu, Xiaonan and Liu, Yu and Li, Xiang},
  journal={ISPRS Journal of Photogrammetry and Remote Sensing},
  volume={224},
  pages={272--286},
  year={2025},
  publisher={Elsevier}
}

@inproceedings{kuckreja2024geochat,
  title={Geochat: Grounded large vision-language model for remote sensing},
  author={Kuckreja, Kartik and Danish, Muhammad Sohail and Naseer, Muzammal and Das, Abhijit and Khan, Salman and Khan, Fahad Shahbaz},
  booktitle={Proceedings of the IEEE/CVF Conference on Computer Vision and Pattern Recognition},
  pages={27831--27840},
  year={2024}
}

@inproceedings{li2020oscar,
  title={Oscar: Object-semantics aligned pre-training for vision-language tasks},
  author={Li, Xiujun and Yin, Xi and Li, Chunyuan and Zhang, Pengchuan and Hu, Xiaowei and Zhang, Lei and Wang, Lijuan and Hu, Houdong and Dong, Li and Wei, Furu and others},
  booktitle={Computer Vision--ECCV 2020: 16th European Conference, Glasgow, UK, August 23--28, 2020, Proceedings, Part XXX 16},
  pages={121--137},
  year={2020},
  organization={Springer}
}

@article{jiang2022vima,
  title={Vima: General robot manipulation with multimodal prompts},
  author={Jiang, Yunfan and Gupta, Agrim and Zhang, Zichen and Wang, Guanzhi and Dou, Yongqiang and Chen, Yanjun and Fei-Fei, Li and Anandkumar, Anima and Zhu, Yuke and Fan, Linxi},
  journal={arXiv preprint arXiv:2210.03094},
  volume={2},
  number={3},
  pages={6},
  year={2022}
}

@article{driess2023palm,
  title={Palm-e: An embodied multimodal language model},
  author={Driess, Danny and Xia, Fei and Sajjadi, Mehdi SM and Lynch, Corey and Chowdhery, Aakanksha and Wahid, Ayzaan and Tompson, Jonathan and Vuong, Quan and Yu, Tianhe and Huang, Wenlong and others},
  year={2023}
}

@article{irvin2024teochat,
  title={Teochat: A large vision-language assistant for temporal earth observation data},
  author={Irvin, Jeremy Andrew and Liu, Emily Ruoyu and Chen, Joyce Chuyi and Dormoy, Ines and Kim, Jinyoung and Khanna, Samar and Zheng, Zhuo and Ermon, Stefano},
  journal={arXiv preprint arXiv:2410.06234},
  year={2024}
}

@article{luo2024skysensegpt,
  title={Skysensegpt: A fine-grained instruction tuning dataset and model for remote sensing vision-language understanding},
  author={Luo, Junwei and Pang, Zhen and Zhang, Yongjun and Wang, Tingzhu and Wang, Linlin and Dang, Bo and Lao, Jiangwei and Wang, Jian and Chen, Jingdong and Tan, Yihua and others},
  journal={arXiv preprint arXiv:2406.10100},
  year={2024}
}

@article{li2024unicl,
  title={Unicl: A universal contrastive learning framework for large time series models},
  author={Li, Jiawei and Peng, Jingshu and Li, Haoyang and Chen, Lei},
  journal={arXiv preprint arXiv:2405.10597},
  year={2024}
}

@article{shen2022k,
  title={K-lite: Learning transferable visual models with external knowledge},
  author={Shen, Sheng and Li, Chunyuan and Hu, Xiaowei and Xie, Yujia and Yang, Jianwei and Zhang, Pengchuan and Gan, Zhe and Wang, Lijuan and Yuan, Lu and Liu, Ce and others},
  journal={Advances in Neural Information Processing Systems},
  volume={35},
  pages={15558--15573},
  year={2022}
}

@article{fan2023improving,
  title={Improving clip training with language rewrites},
  author={Fan, Lijie and Krishnan, Dilip and Isola, Phillip and Katabi, Dina and Tian, Yonglong},
  journal={Advances in Neural Information Processing Systems},
  volume={36},
  pages={35544--35575},
  year={2023}
}

@article{wang2021simvlm,
  title={Simvlm: Simple visual language model pretraining with weak supervision},
  author={Wang, Zirui and Yu, Jiahui and Yu, Adams Wei and Dai, Zihang and Tsvetkov, Yulia and Cao, Yuan},
  journal={arXiv preprint arXiv:2108.10904},
  year={2021}
}

@inproceedings{qu2016deep,
  title={Deep semantic understanding of high resolution remote sensing image},
  author={Qu, Bo and Li, Xuelong and Tao, Dacheng and Lu, Xiaoqiang},
  booktitle={2016 International conference on computer, information and telecommunication systems (Cits)},
  pages={1--5},
  year={2016},
  organization={IEEE}
}

@article{yuan2022exploring,
  title={Exploring a fine-grained multiscale method for cross-modal remote sensing image retrieval},
  author={Yuan, Zhiqiang and Zhang, Wenkai and Fu, Kun and Li, Xuan and Deng, Chubo and Wang, Hongqi and Sun, Xian},
  journal={arXiv preprint arXiv:2204.09868},
  year={2022}
}

@article{liu2024remoteclip,
  title={Remoteclip: A vision language foundation model for remote sensing},
  author={Liu, Fan and Chen, Delong and Guan, Zhangqingyun and Zhou, Xiaocong and Zhu, Jiale and Ye, Qiaolin and Fu, Liyong and Zhou, Jun},
  journal={IEEE Transactions on Geoscience and Remote Sensing},
  year={2024},
  publisher={IEEE}
}

@article{vivanco2023geoclip,
  title={Geoclip: Clip-inspired alignment between locations and images for effective worldwide geo-localization},
  author={Vivanco Cepeda, Vicente and Nayak, Gaurav Kumar and Shah, Mubarak},
  journal={Advances in Neural Information Processing Systems},
  volume={36},
  pages={8690--8701},
  year={2023}
}

@article{lu2019vilbert,
  title={Vilbert: Pretraining task-agnostic visiolinguistic representations for vision-and-language tasks},
  author={Lu, Jiasen and Batra, Dhruv and Parikh, Devi and Lee, Stefan},
  journal={Advances in neural information processing systems},
  volume={32},
  year={2019}
}

@inproceedings{he2017mask,
  title={Mask r-cnn},
  author={He, Kaiming and Gkioxari, Georgia and Doll{\'a}r, Piotr and Girshick, Ross},
  booktitle={Proceedings of the IEEE international conference on computer vision},
  pages={2961--2969},
  year={2017}
}

@article{su2019vl,
  title={Vl-bert: Pre-training of generic visual-linguistic representations},
  author={Su, Weijie and Zhu, Xizhou and Cao, Yue and Li, Bin and Lu, Lewei and Wei, Furu and Dai, Jifeng},
  journal={arXiv preprint arXiv:1908.08530},
  year={2019}
}

@inproceedings{kim2021vilt,
  title={Vilt: Vision-and-language transformer without convolution or region supervision},
  author={Kim, Wonjae and Son, Bokyung and Kim, Ildoo},
  booktitle={International conference on machine learning},
  pages={5583--5594},
  year={2021},
  organization={PMLR}
}

@article{yang2023fine,
  title={Fine-grained visual prompting},
  author={Yang, Lingfeng and Wang, Yueze and Li, Xiang and Wang, Xinlong and Yang, Jian},
  journal={Advances in Neural Information Processing Systems},
  volume={36},
  pages={24993--25006},
  year={2023}
}

@inproceedings{sun2024alpha,
  title={Alpha-clip: A clip model focusing on wherever you want},
  author={Sun, Zeyi and Fang, Ye and Wu, Tong and Zhang, Pan and Zang, Yuhang and Kong, Shu and Xiong, Yuanjun and Lin, Dahua and Wang, Jiaqi},
  booktitle={Proceedings of the IEEE/CVF conference on computer vision and pattern recognition},
  pages={13019--13029},
  year={2024}
}

@article{wang2023all,
  title={The all-seeing project: Towards panoptic visual recognition and understanding of the open world},
  author={Wang, Weiyun and Shi, Min and Li, Qingyun and Wang, Wenhai and Huang, Zhenhang and Xing, Linjie and Chen, Zhe and Li, Hao and Zhu, Xizhou and Cao, Zhiguo and others},
  journal={arXiv preprint arXiv:2308.01907},
  year={2023}
}

@article{shi2017can,
  title={Can a machine generate humanlike language descriptions for a remote sensing image?},
  author={Shi, Zhenwei and Zou, Zhengxia},
  journal={IEEE Transactions on Geoscience and Remote Sensing},
  volume={55},
  number={6},
  pages={3623--3634},
  year={2017},
  publisher={IEEE}
}

@inproceedings{sun2022visual,
  title={Visual grounding in remote sensing images},
  author={Sun, Yuxi and Feng, Shanshan and Li, Xutao and Ye, Yunming and Kang, Jian and Huang, Xu},
  booktitle={Proceedings of the 30th ACM International conference on Multimedia},
  pages={404--412},
  year={2022}
}

@article{zhan2023rsvg,
  title={Rsvg: Exploring data and models for visual grounding on remote sensing data},
  author={Zhan, Yang and Xiong, Zhitong and Yuan, Yuan},
  journal={IEEE Transactions on Geoscience and Remote Sensing},
  volume={61},
  pages={1--13},
  year={2023},
  publisher={IEEE}
}

@article{lu2023few,
  title={Few-shot object detection in aerial imagery guided by text-modal knowledge},
  author={Lu, Xiaonan and Sun, Xian and Diao, Wenhui and Mao, Yongqiang and Li, Junxi and Zhang, Yidan and Wang, Peijin and Fu, Kun},
  journal={IEEE Transactions on Geoscience and Remote Sensing},
  volume={61},
  pages={1--19},
  year={2023},
  publisher={IEEE}
}

@article{liu2023visual,
  title={Visual instruction tuning},
  author={Liu, Haotian and Li, Chunyuan and Wu, Qingyang and Lee, Yong Jae},
  journal={Advances in neural information processing systems},
  volume={36},
  pages={34892--34916},
  year={2023}
}

@article{chen2023minigpt,
  title={Minigpt-v2: large language model as a unified interface for vision-language multi-task learning},
  author={Chen, Jun and Zhu, Deyao and Shen, Xiaoqian and Li, Xiang and Liu, Zechun and Zhang, Pengchuan and Krishnamoorthi, Raghuraman and Chandra, Vikas and Xiong, Yunyang and Elhoseiny, Mohamed},
  journal={arXiv preprint arXiv:2310.09478},
  year={2023}
}

@article{dai2023instructblip,
  title={Instructblip: Towards general-purpose vision-language models with instruction tuning},
  author={Dai, Wenliang and Li, Junnan and Li, Dongxu and Tiong, Anthony and Zhao, Junqi and Wang, Weisheng and Li, Boyang and Fung, Pascale N and Hoi, Steven},
  journal={Advances in neural information processing systems},
  volume={36},
  pages={49250--49267},
  year={2023}
}

@article{bai2023qwen,
  title={Qwen technical report},
  author={Bai, Jinze and Bai, Shuai and Chu, Yunfei and Cui, Zeyu and Dang, Kai and Deng, Xiaodong and Fan, Yang and Ge, Wenbin and Han, Yu and Huang, Fei and others},
  journal={arXiv preprint arXiv:2309.16609},
  year={2023}
}

@inproceedings{liu2024LLaVA1.5,
  title={Improved baselines with visual instruction tuning},
  author={Liu, Haotian and Li, Chunyuan and Li, Yuheng and Lee, Yong Jae},
  booktitle={Proceedings of the IEEE/CVF conference on computer vision and pattern recognition},
  pages={26296--26306},
  year={2024}
}

@article{dai2023instructblipgeneralpurposevisionlanguagemodels,
      title={InstructBLIP: Towards General-purpose Vision-Language Models with Instruction Tuning}, 
      author={Wenliang Dai and Junnan Li and Dongxu Li and Anthony Meng Huat Tiong and Junqi Zhao and Weisheng Wang and Boyang Li and Pascale Fung and Steven Hoi},
      year={2023},
      eprint={2305.06500},
      archivePrefix={arXiv},
      primaryClass={cs.CV},
      url={https://arxiv.org/abs/2305.06500}, 
}

@article{wang2023caption,
  title={Caption anything: Interactive image description with diverse multimodal controls},
  author={Wang, Teng and Zhang, Jinrui and Fei, Junjie and Zheng, Hao and Tang, Yunlong and Li, Zhe and Gao, Mingqi and Zhao, Shanshan},
  journal={arXiv preprint arXiv:2305.02677},
  year={2023}
}

@article{lobry2020rsvqa,
  title={RSVQA: Visual question answering for remote sensing data},
  author={Lobry, Sylvain and Marcos, Diego and Murray, Jesse and Tuia, Devis},
  journal={IEEE Transactions on Geoscience and Remote Sensing},
  volume={58},
  number={12},
  pages={8555--8566},
  year={2020},
  publisher={IEEE}
}

@article{al2022open,
  title={Open-ended remote sensing visual question answering with transformers},
  author={Al Rahhal, Mohamad M and Bazi, Yakoub and Alsaleh, Sara O and Al-Razgan, Muna and Mekhalfi, Mohamed Lamine and Al Zuair, Mansour and Alajlan, Naif},
  journal={International Journal of Remote Sensing},
  volume={43},
  number={18},
  pages={6809--6823},
  year={2022},
  publisher={Taylor \& Francis}
}

@inproceedings{chappuis2022prompt,
  title={Prompt-RSVQA: Prompting visual context to a language model for remote sensing visual question answering},
  author={Chappuis, Christel and Zermatten, Val{\'e}rie and Lobry, Sylvain and Le Saux, Bertrand and Tuia, Devis},
  booktitle={Proceedings of the IEEE/CVF conference on computer vision and pattern recognition},
  pages={1372--1381},
  year={2022}
}

@inproceedings{maniparambil2024vision,
  title={Do Vision and Language Encoders Represent the World Similarly?},
  author={Maniparambil, Mayug and Akshulakov, Raiymbek and Djilali, Yasser Abdelaziz Dahou and El Amine Seddik, Mohamed and Narayan, Sanath and Mangalam, Karttikeya and O'Connor, Noel E},
  booktitle={Proceedings of the IEEE/CVF Conference on Computer Vision and Pattern Recognition},
  pages={14334--14343},
  year={2024}
}

@article{pang2023frozen,
  title={Frozen transformers in language models are effective visual encoder layers},
  author={Pang, Ziqi and Xie, Ziyang and Man, Yunze and Wang, Yu-Xiong},
  journal={arXiv preprint arXiv:2310.12973},
  year={2023}
}

@article{adeniji2023language,
  title={Language reward modulation for pretraining reinforcement learning},
  author={Adeniji, Ademi and Xie, Amber and Sferrazza, Carmelo and Seo, Younggyo and James, Stephen and Abbeel, Pieter},
  journal={arXiv preprint arXiv:2308.12270},
  year={2023}
}

@article{mall2023remote,
  title={Remote sensing vision-language foundation models without annotations via ground remote alignment},
  author={Mall, Utkarsh and Phoo, Cheng Perng and Liu, Meilin Kelsey and Vondrick, Carl and Hariharan, Bharath and Bala, Kavita},
  journal={arXiv preprint arXiv:2312.06960},
  year={2023}
}

@article{vaswani2017attention,
  title={Attention is all you need},
  author={Vaswani, Ashish and Shazeer, Noam and Parmar, Niki and Uszkoreit, Jakob and Jones, Llion and Gomez, Aidan N and Kaiser, {\L}ukasz and Polosukhin, Illia},
  journal={Advances in neural information processing systems},
  volume={30},
  year={2017}
}

@article{mizoguchi1983prolog,
  title={PROLOG based expert system},
  author={Mizoguchi, Fumio},
  journal={New Generation Computing},
  volume={1},
  number={1},
  pages={99--104},
  year={1983},
  publisher={Springer}
}

@article{wang2024rs,
  title={RS-DFM: A Remote Sensing Distributed Foundation Model for Diverse Downstream Tasks},
  author={Wang, Zhechao and Cheng, Peirui and Tian, Pengju and Wang, Yuchao and Chen, Mingxin and Duan, Shujing and Wang, Zhirui and Li, Xinming and Sun, Xian},
  journal={arXiv preprint arXiv:2406.07032},
  year={2024}
}

@article{feng2023self,
  title={A self-supervised cross-modal remote sensing foundation model with multi-domain representation and cross-domain fusion},
  author={Feng, Yingchao and Wang, Peijin and Diao, Wenhui and He, Qibin and Hu, Huiyang and Bi, Hanbo and Sun, Xian and Fu, Kun},
  booktitle={IGARSS 2023-2023 IEEE International Geoscience and Remote Sensing Symposium},
  pages={2239--2242},
  year={2023},
  organization={IEEE}
}

@article{openai_gpt4_2023,
  title   = {GPT-4 Technical Report},
  author  = {{OpenAI}},
  journal = {arXiv preprint arXiv:2303.08774},
  year    = {2023}
}

@inproceedings{wu2024v,
  title={V?: Guided visual search as a core mechanism in multimodal llms},
  author={Wu, Penghao and Xie, Saining},
  booktitle={Proceedings of the IEEE/CVF Conference on Computer Vision and Pattern Recognition},
  pages={13084--13094},
  year={2024}
}

@article{zhang2024beyond,
  title={Beyond LLaVA-HD: Diving into High-Resolution Large Multimodal Models},
  author={Zhang, Yi-Fan and Wen, Qingsong and Fu, Chaoyou and Wang, Xue and Zhang, Zhang and Wang, Liang and Jin, Rong},
  journal={CoRR},
  year={2024}
}

@article{anthropic2024claude,
  title={Claude 3.5 sonnet model card addendum},
  author={Anthropic, AI},
  journal={Claude-3.5 Model Card},
  volume={3},
  number={6},
  year={2024}
}

@article{li2024mini,
  title={Mini-Gemini: Mining the Potential of Multi-modality Vision Language Models},
  author={Li, Yanwei and Zhang, Yuechen and Wang, Chengyao and Zhong, Zhisheng and Chen, Yixin and Chu, Ruihang and Liu, Shaoteng and Jia, Jiaya},
  journal={CoRR},
  year={2024}
}

@inproceedings{singh2024geollm,
  title={Geollm-engine: A realistic environment for building geospatial copilots},
  author={Singh, Simranjit and Fore, Michael and Stamoulis, Dimitrios},
  booktitle={Proceedings of the IEEE/CVF Conference on Computer Vision and Pattern Recognition},
  pages={585--594},
  year={2024}
}

@misc{openai_gpt4o_2024,
  title        = {Hello {GPT-4o}},
  author       = {{OpenAI}},
  howpublished = {\url{https://openai.com/index/hello-gpt-4o/}},
  year         = {2024},
  month        = may,
  note         = {Accessed: 2025-XX-XX}
}

@article{lu2023chameleon,
  title={Chameleon: Plug-and-play compositional reasoning with large language models},
  author={Lu, Pan and Peng, Baolin and Cheng, Hao and Galley, Michel and Chang, Kai-Wei and Wu, Ying Nian and Zhu, Song-Chun and Gao, Jianfeng},
  journal={Advances in Neural Information Processing Systems},
  volume={36},
  pages={43447--43478},
  year={2023}
}

@article{fourney2024magentic,
  title={Magentic-one: A generalist multi-agent system for solving complex tasks},
  author={Fourney, Adam and Bansal, Gagan and Mozannar, Hussein and Tan, Cheng and Salinas, Eduardo and Niedtner, Friederike and Proebsting, Grace and Bassman, Griffin and Gerrits, Jack and Alber, Jacob and others},
  journal={arXiv preprint arXiv:2411.04468},
  year={2024}
}

@article{lee2025multi,
  title={Multi-Agent Geospatial Copilots for Remote Sensing Workflows},
  author={Lee, Chaehong and Paramanayakam, Varatheepan and Karatzas, Andreas and Jian, Yanan and Fore, Michael and Liao, Heming and Yu, Fuxun and Li, Ruopu and Anagnostopoulos, Iraklis and Stamoulis, Dimitrios},
  journal={arXiv preprint arXiv:2501.16254},
  year={2025}
}

@article{singh2024evaluating,
  title={Evaluating tool-augmented agents in remote sensing platforms},
  author={Singh, Simranjit and Fore, Michael and Stamoulis, Dimitrios},
  journal={arXiv preprint arXiv:2405.00709},
  year={2024}
}

@article{xu2024rs,
  title={RS-Agent: Automating Remote Sensing Tasks through Intelligent Agent},
  author={Xu, Wenjia and Yu, Zijian and Mu, Boyang and Wei, Zhiwei and Zhang, Yuanben and Li, Guangzuo and Peng, Mugen},
  journal={arXiv preprint arXiv:2406.07089},
  year={2024}
}

@inproceedings{stamoulis2025geo,
  title={Geo-olm: Enabling sustainable earth observation studies with cost-efficient open language models \& state-driven workflows},
  author={Stamoulis, Dimitrios and Marculescu, Diana},
  booktitle={Proceedings of the ACM SIGCAS/SIGCHI Conference on Computing and Sustainable Societies},
  pages={608--619},
  year={2025}
}

@article{zhu2025semantic,
  title={Semantic-cd: Remote sensing image semantic change detection towards open-vocabulary setting},
  author={Zhu, Yongshuo and Li, Lu and Chen, Keyan and Liu, Chenyang and Zhou, Fugen and Shi, Zhenwei},
  journal={arXiv preprint arXiv:2501.06808},
  year={2025}
}

@article{grattafiori2024llama,
  title={The llama 3 herd of models},
  author={Grattafiori, Aaron and Dubey, Abhimanyu and Jauhri, Abhinav and Pandey, Abhinav and Kadian, Abhishek and Al-Dahle, Ahmad and Letman, Aiesha and Mathur, Akhil and Schelten, Alan and Vaughan, Alex and others},
  journal={arXiv preprint arXiv:2407.21783},
  year={2024}
}

@article{brown2020language,
  title={Language models are few-shot learners},
  author={Brown, Tom and Mann, Benjamin and Ryder, Nick and Subbiah, Melanie and Kaplan, Jared D and Dhariwal, Prafulla and Neelakantan, Arvind and Shyam, Pranav and Sastry, Girish and Askell, Amanda and others},
  journal={Advances in neural information processing systems},
  volume={33},
  pages={1877--1901},
  year={2020}
}

@article{liu2024deepseek,
  title={Deepseek-v3 technical report},
  author={Liu, Aixin and Feng, Bei and Xue, Bing and Wang, Bingxuan and Wu, Bochao and Lu, Chengda and Zhao, Chenggang and Deng, Chengqi and Zhang, Chenyu and Ruan, Chong and others},
  journal={arXiv preprint arXiv:2412.19437},
  year={2024}
}

@inproceedings{hackel2023lit,
  title={Lit-4-rsvqa: Lightweight transformer-based visual question answering in remote sensing},
  author={Hackel, Leonard and Clasen, Kai Norman and Ravanbakhsh, Mahdyar and Demir, Beg{\"u}m},
  booktitle={IGARSS 2023-2023 IEEE International Geoscience and Remote Sensing Symposium},
  pages={2231--2234},
  year={2023},
  organization={IEEE}
}

@article{zhang2023geogpt,
  title={Geogpt: Understanding and processing geospatial tasks through an autonomous gpt},
  author={Zhang, Yifan and Wei, Cheng and Wu, Shangyou and He, Zhengting and Yu, Wenhao},
  journal={arXiv preprint arXiv:2307.07930},
  year={2023}
}

@inproceedings{muhtar2024lhrs,
  title={Lhrs-bot: Empowering remote sensing with vgi-enhanced large multimodal language model},
  author={Muhtar, Dilxat and Li, Zhenshi and Gu, Feng and Zhang, Xueliang and Xiao, Pengfeng},
  booktitle={European Conference on Computer Vision},
  pages={440--457},
  year={2024},
  organization={Springer}
}

@article{zhan2025skyeyegpt,
  title={Skyeyegpt: Unifying remote sensing vision-language tasks via instruction tuning with large language model},
  author={Zhan, Yang and Xiong, Zhitong and Yuan, Yuan},
  journal={ISPRS Journal of Photogrammetry and Remote Sensing},
  volume={221},
  pages={64--77},
  year={2025},
  publisher={Elsevier}
}

@article{zhang2024earthgpt,
  title={Earthgpt: A universal multi-modal large language model for multi-sensor image comprehension in remote sensing domain},
  author={Zhang, Wei and Cai, Miaoxin and Zhang, Tong and Zhuang, Yin and Mao, Xuerui},
  journal={IEEE Transactions on Geoscience and Remote Sensing},
  year={2024},
  publisher={IEEE}
}

@article{renze2024self,
  title={Self-reflection in llm agents: Effects on problem-solving performance},
  author={Renze, Matthew and Guven, Erhan},
  journal={arXiv preprint arXiv:2405.06682},
  year={2024}
}

@article{madaan2023self,
  title={Self-refine: Iterative refinement with self-feedback},
  author={Madaan, Aman and Tandon, Niket and Gupta, Prakhar and Hallinan, Skyler and Gao, Luyu and Wiegreffe, Sarah and Alon, Uri and Dziri, Nouha and Prabhumoye, Shrimai and Yang, Yiming and others},
  journal={Advances in Neural Information Processing Systems},
  volume={36},
  pages={46534--46594},
  year={2023}
}

@inproceedings{guo2024remote,
  title={Remote sensing chatgpt: Solving remote sensing tasks with chatgpt and visual models},
  author={Guo, Haonan and Su, Xin and Wu, Chen and Du, Bo and Zhang, Liangpei and Li, Deren},
  booktitle={IGARSS 2024-2024 IEEE International Geoscience and Remote Sensing Symposium},
  pages={11474--11478},
  year={2024},
  organization={IEEE}
}

@article{zhang2022automatic,
  title={Automatic chain of thought prompting in large language models},
  author={Zhang, Zhuosheng and Zhang, Aston and Li, Mu and Smola, Alex},
  journal={arXiv preprint arXiv:2210.03493},
  year={2022}
}

@inproceedings{zhong2024memorybank,
  title={Memorybank: Enhancing large language models with long-term memory},
  author={Zhong, Wanjun and Guo, Lianghong and Gao, Qiqi and Ye, He and Wang, Yanlin},
  booktitle={Proceedings of the AAAI Conference on Artificial Intelligence},
  volume={38},
  number={17},
  pages={19724--19731},
  year={2024}
}

@inproceedings{yao2023react,
  title={React: Synergizing reasoning and acting in language models},
  author={Yao, Shunyu and Zhao, Jeffrey and Yu, Dian and Du, Nan and Shafran, Izhak and Narasimhan, Karthik and Cao, Yuan},
  booktitle={International Conference on Learning Representations (ICLR)},
  year={2023}
}

@article{dongre2024respact,
  title={ReSpAct: Harmonizing Reasoning, Speaking, and Acting Towards Building Large Language Model-Based Conversational AI Agents},
  author={Dongre, Vardhan and Yang, Xiaocheng and Acikgoz, Emre Can and Dey, Suvodip and Tur, Gokhan and Hakkani-T{\"u}r, Dilek},
  journal={arXiv preprint arXiv:2411.00927},
  year={2024}
}

@article{church2017word2vec,
  title={Word2Vec},
  author={Church, Kenneth Ward},
  journal={Natural Language Engineering},
  volume={23},
  number={1},
  pages={155--162},
  year={2017},
  publisher={Cambridge University Press}
}

@inproceedings{pennington2014glove,
  title={Glove: Global vectors for word representation},
  author={Pennington, Jeffrey and Socher, Richard and Manning, Christopher D},
  booktitle={Proceedings of the 2014 conference on empirical methods in natural language processing (EMNLP)},
  pages={1532--1543},
  year={2014}
}

@article{joulin2016fasttext,
  title={Fasttext. zip: Compressing text classification models},
  author={Joulin, Armand and Grave, Edouard and Bojanowski, Piotr and Douze, Matthijs and J{\'e}gou, H{\'e}rve and Mikolov, Tomas},
  journal={arXiv preprint arXiv:1612.03651},
  year={2016}
}

@article{sarzynska2021detecting,
  title={Detecting formal thought disorder by deep contextualized word representations},
  author={Sarzynska-Wawer, Justyna and Wawer, Aleksander and Pawlak, Aleksandra and Szymanowska, Julia and Stefaniak, Izabela and Jarkiewicz, Michal and Okruszek, Lukasz},
  journal={Psychiatry Research},
  volume={304},
  pages={114135},
  year={2021},
  publisher={Elsevier}
}

@article{koroteev2021bert,
  title={BERT: a review of applications in natural language processing and understanding},
  author={Koroteev, Mikhail V},
  journal={arXiv preprint arXiv:2103.11943},
  year={2021}
}

@article{radford2018improving,
  title={Improving language understanding by generative pre-training},
  author={Radford, Alec and Narasimhan, Karthik and Salimans, Tim and Sutskever, Ilya and others},
  year={2018},
  publisher={San Francisco, CA, USA}
}

@article{pezeshkpour2018embedding,
  title={Embedding multimodal relational data for knowledge base completion},
  author={Pezeshkpour, Pouya and Chen, Liyan and Singh, Sameer},
  journal={arXiv preprint arXiv:1809.01341},
  year={2018}
}

@inproceedings{kim2020mule,
  title={Mule: Multimodal universal language embedding},
  author={Kim, Donghyun and Saito, Kuniaki and Saenko, Kate and Sclaroff, Stan and Plummer, Bryan},
  booktitle={Proceedings of the AAAI Conference on Artificial Intelligence},
  volume={34},
  number={07},
  pages={11254--11261},
  year={2020}
}

@article{jiang2024e5,
  title={E5-v: Universal embeddings with multimodal large language models},
  author={Jiang, Ting and Song, Minghui and Zhang, Zihan and Huang, Haizhen and Deng, Weiwei and Sun, Feng and Zhang, Qi and Wang, Deqing and Zhuang, Fuzhen},
  journal={arXiv preprint arXiv:2407.12580},
  year={2024}
}

@article{zhang2024earthmarker,
  title={Earthmarker: Visual prompt learning for region-level and point-level remote sensing imagery comprehension},
  author={Zhang, Wei and Cai, Miaoxin and Zhang, Tong and Li, Jun and Zhuang, Yin and Mao, Xuerui},
  journal={arXiv preprint arXiv:2407.13596},
  year={2024}
}

@article{aizawa2003information,
  title={An information-theoretic perspective of tf--idf measures},
  author={Aizawa, Akiko},
  journal={Information Processing \& Management},
  volume={39},
  number={1},
  pages={45--65},
  year={2003},
  publisher={Elsevier}
}

@article{cover1967nearest,
  title={Nearest neighbor pattern classification},
  author={Cover, Thomas and Hart, Peter},
  journal={IEEE transactions on information theory},
  volume={13},
  number={1},
  pages={21--27},
  year={1967},
  publisher={IEEE}
}

@article{arya1998optimal,
  title={An optimal algorithm for approximate nearest neighbor searching fixed dimensions},
  author={Arya, Sunil and Mount, David M and Netanyahu, Nathan S and Silverman, Ruth and Wu, Angela Y},
  journal={Journal of the ACM (JACM)},
  volume={45},
  number={6},
  pages={891--923},
  year={1998},
  publisher={ACM New York, NY, USA}
}

@article{shaw2018self,
  title={Self-attention with relative position representations},
  author={Shaw, Peter and Uszkoreit, Jakob and Vaswani, Ashish},
  journal={arXiv preprint arXiv:1803.02155},
  year={2018}
}

@article{hou2019cross,
  title={Cross attention network for few-shot classification},
  author={Hou, Ruibing and Chang, Hong and Ma, Bingpeng and Shan, Shiguang and Chen, Xilin},
  journal={Advances in neural information processing systems},
  volume={32},
  year={2019}
}

@article{sun2024block,
  title={Block-Attention for Efficient RAG},
  author={Sun, East and Wang, Yan and Tian, Lan},
  journal={arXiv preprint arXiv:2409.15355},
  year={2024}
}

@article{mi2024knowledge,
  title={Knowledge-aware text-image retrieval for remote sensing images},
  author={Mi, Li and Dai, Xianjie and Castillo-Navarro, Javiera and Tuia, Devis},
  journal={IEEE Transactions on Geoscience and Remote Sensing},
  year={2024},
  publisher={IEEE}
}

@article{hu2022view,
  title={View planning for object pose estimation using point clouds: An active robot perception approach},
  author={Hu, Jie and Pagilla, Prabhakar R},
  journal={IEEE Robotics and Automation Letters},
  volume={7},
  number={4},
  pages={9248--9255},
  year={2022},
  publisher={IEEE}
}

@article{saito2021select,
  title={How to select and use tools?: Active perception of target objects using multimodal deep learning},
  author={Saito, Namiko and Ogata, Tetsuya and Funabashi, Satoshi and Mori, Hiroki and Sugano, Shigeki},
  journal={IEEE Robotics and Automation Letters},
  volume={6},
  number={2},
  pages={2517--2524},
  year={2021},
  publisher={IEEE}
}

@book{johnson2005pid,
  title={PID control},
  author={Johnson, Michael A and Moradi, Mohammad H},
  year={2005},
  publisher={Springer}
}

@article{guardeno2019mimo,
  title={MIMO PID controller tuning method for quadrotor based on LQR/LQG theory},
  author={Guarde{\~n}o, Rafael and L{\'o}pez, Manuel J and S{\'a}nchez, V{\'\i}ctor M},
  journal={Robotics},
  volume={8},
  number={2},
  pages={36},
  year={2019},
  publisher={MDPI}
}

@article{amos2018differentiable,
  title={Differentiable mpc for end-to-end planning and control},
  author={Amos, Brandon and Jimenez, Ivan and Sacks, Jacob and Boots, Byron and Kolter, J Zico},
  journal={Advances in neural information processing systems},
  volume={31},
  year={2018}
}

@article{tsounis2020deepgait,
  title={Deepgait: Planning and control of quadrupedal gaits using deep reinforcement learning},
  author={Tsounis, Vassilios and Alge, Mitja and Lee, Joonho and Farshidian, Farbod and Hutter, Marco},
  journal={IEEE Robotics and Automation Letters},
  volume={5},
  number={2},
  pages={3699--3706},
  year={2020},
  publisher={IEEE}
}

@inproceedings{marchesini2021centralizing,
  title={Centralizing state-values in dueling networks for multi-robot reinforcement learning mapless navigation},
  author={Marchesini, Enrico and Farinelli, Alessandro},
  booktitle={2021 IEEE/RSJ International Conference on Intelligent Robots and Systems (IROS)},
  pages={4583--4588},
  year={2021},
  organization={IEEE}
}

@article{jin2023robot,
  title={Robot pilot: a new autonomous system toward flying manned aerial vehicles},
  author={Jin, Zibo and Li, Daochun and Xiang, Jinwu},
  journal={Engineering},
  volume={27},
  pages={242--253},
  year={2023},
  publisher={Elsevier}
}

@article{min2024toward,
  title={Toward Fully Autonomous Aviation: PIBOT, a Humanoid Robot Pilot for Human-Centric Aircraft Cockpits},
  author={Min, Sungjae and Kang, Gyuree and Kim, Hyungjoo and Shim, David Hyunchul},
  journal={IEEE Robotics \& Automation Magazine},
  year={2024},
  publisher={IEEE}
}

@article{qin2023toolllm,
  title={Toolllm: Facilitating large language models to master 16000+ real-world apis},
  author={Qin, Yujia and Liang, Shihao and Ye, Yining and Zhu, Kunlun and Yan, Lan and Lu, Yaxi and Lin, Yankai and Cong, Xin and Tang, Xiangru and Qian, Bill and others},
  journal={arXiv preprint arXiv:2307.16789},
  year={2023}
}

@article{levine2018learning,
  title={Learning hand-eye coordination for robotic grasping with deep learning and large-scale data collection},
  author={Levine, Sergey and Pastor, Peter and Krizhevsky, Alex and Ibarz, Julian and Quillen, Deirdre},
  journal={The International journal of robotics research},
  volume={37},
  number={4-5},
  pages={421--436},
  year={2018},
  publisher={SAGE Publications Sage UK: London, England}
}

@article{pandya2023automating,
  title={Automating Customer Service using LangChain: Building custom open-source GPT Chatbot for organizations},
  author={Pandya, Keivalya and Holia, Mehfuza},
  journal={arXiv preprint arXiv:2310.05421},
  year={2023}
}

@article{yang2023auto,
  title={Auto-gpt for online decision making: Benchmarks and additional opinions},
  author={Yang, Hui and Yue, Sifu and He, Yunzhong},
  journal={arXiv preprint arXiv:2306.02224},
  year={2023}
}

@inproceedings{shridhar2022cliport,
  title={Cliport: What and where pathways for robotic manipulation},
  author={Shridhar, Mohit and Manuelli, Lucas and Fox, Dieter},
  booktitle={Conference on robot learning},
  pages={894--906},
  year={2022},
  organization={PMLR}
}

@article{schick2023toolformer,
  title={Toolformer: Language models can teach themselves to use tools},
  author={Schick, Timo and Dwivedi-Yu, Jane and Dess{\`\i}, Roberto and Raileanu, Roberta and Lomeli, Maria and Hambro, Eric and Zettlemoyer, Luke and Cancedda, Nicola and Scialom, Thomas},
  journal={Advances in Neural Information Processing Systems},
  volume={36},
  pages={68539--68551},
  year={2023}
}

@article{chen2024re,
  title={Re-Invoke: Tool invocation rewriting for zero-shot tool retrieval},
  author={Chen, Yanfei and Yoon, Jinsung and Sachan, Devendra Singh and Wang, Qingze and Cohen-Addad, Vincent and Bateni, Mohammadhossein and Lee, Chen-Yu and Pfister, Tomas},
  journal={arXiv preprint arXiv:2408.01875},
  year={2024}
}

@article{nakano2021webgpt,
  title={Webgpt: Browser-assisted question-answering with human feedback},
  author={Nakano, Reiichiro and Hilton, Jacob and Balaji, Suchir and Wu, Jeff and Ouyang, Long and Kim, Christina and Hesse, Christopher and Jain, Shantanu and Kosaraju, Vineet and Saunders, William and others},
  journal={arXiv preprint arXiv:2112.09332},
  year={2021}
}

@article{qin2023webcpm,
  title={Webcpm: Interactive web search for chinese long-form question answering},
  author={Qin, Yujia and Cai, Zihan and Jin, Dian and Yan, Lan and Liang, Shihao and Zhu, Kunlun and Lin, Yankai and Han, Xu and Ding, Ning and Wang, Huadong and others},
  journal={arXiv preprint arXiv:2305.06849},
  year={2023}
}

@article{hu2022lora,
  title={Lora: Low-rank adaptation of large language models.},
  author={Hu, Edward J and Shen, Yelong and Wallis, Phillip and Allen-Zhu, Zeyuan and Li, Yuanzhi and Wang, Shean and Wang, Lu and Chen, Weizhu and others},
  journal={ICLR},
  volume={1},
  number={2},
  pages={3},
  year={2022}
}

@inproceedings{huh2024position,
  title={Position: The platonic representation hypothesis},
  author={Huh, Minyoung and Cheung, Brian and Wang, Tongzhou and Isola, Phillip},
  booktitle={Forty-first International Conference on Machine Learning},
  year={2024}
}

@article{anderson1997act,
  title={ACT-R: A theory of higher level cognition and its relation to visual attention},
  author={Anderson, John R and Matessa, Michael and Lebiere, Christian},
  journal={Human--Computer Interaction},
  volume={12},
  number={4},
  pages={439--462},
  year={1997},
  publisher={Taylor \& Francis}
}

@inproceedings{wagner2024generative,
  title={A generative artificial intelligence framework for earth observation analysis},
  author={Wagner, Otto and Gordon, Jeffrey and Mousa, Aya and Terry, Brian and Baptist, Joshua and Yetkin, Oguz and Borges, David and Gowda, Sanjay},
  booktitle={IGARSS 2024-2024 IEEE International Geoscience and Remote Sensing Symposium},
  pages={6986--6990},
  year={2024},
  organization={IEEE}
}

@article{hochreiter1997long,
  title={Long short-term memory},
  author={Hochreiter, Sepp and Schmidhuber, J{\"u}rgen},
  journal={Neural computation},
  volume={9},
  number={8},
  pages={1735--1780},
  year={1997},
  publisher={MIT press}
}

@article{huang2025survey,
  title={A survey on remote sensing foundation models: From vision to multimodality},
  author={Huang, Ziyue and Yan, Hongxi and Zhan, Qiqi and Yang, Shuai and Zhang, Mingming and Zhang, Chenkai and Lei, YiMing and Liu, Zeming and Liu, Qingjie and Wang, Yunhong},
  journal={arXiv preprint arXiv:2503.22081},
  year={2025}
}

@article{li2024vision,
  title={Vision-language models in remote sensing: Current progress and future trends},
  author={Li, Xiang and Wen, Congcong and Hu, Yuan and Yuan, Zhenghang and Zhu, Xiao Xiang},
  journal={IEEE Geoscience and Remote Sensing Magazine},
  volume={12},
  number={2},
  pages={32--66},
  year={2024},
  publisher={IEEE}
}

@article{zhu2017deep,
  title={Deep learning in remote sensing: A comprehensive review and list of resources},
  author={Zhu, Xiao Xiang and Tuia, Devis and Mou, Lichao and Xia, Gui-Song and Zhang, Liangpei and Xu, Feng and Fraundorfer, Friedrich},
  journal={IEEE geoscience and remote sensing magazine},
  volume={5},
  number={4},
  pages={8--36},
  year={2017},
  publisher={IEEE}
}

@article{li2020object,
  title={Object detection in optical remote sensing images: A survey and a new benchmark},
  author={Li, Ke and Wan, Gang and Cheng, Gong and Meng, Liqiu and Han, Junwei},
  journal={ISPRS journal of photogrammetry and remote sensing},
  volume={159},
  pages={296--307},
  year={2020},
  publisher={Elsevier}
}

@article{xiao2025foundation,
  title={Foundation models for remote sensing and earth observation: A survey},
  author={Xiao, Aoran and Xuan, Weihao and Wang, Junjue and Huang, Jiaxing and Tao, Dacheng and Lu, Shijian and Yokoya, Naoto},
  journal={IEEE Geoscience and Remote Sensing Magazine},
  year={2025},
  publisher={IEEE}
}

@article{lu2025vision,
  title={Vision foundation models in remote sensing: A survey},
  author={Lu, Siqi and Guo, Junlin and Zimmer-Dauphinee, James R and Nieusma, Jordan M and Wang, Xiao and Wernke, Steven A and Huo, Yuankai and others},
  journal={IEEE Geoscience and Remote Sensing Magazine},
  year={2025},
  publisher={IEEE}
}

@inproceedings{manas2021seasonal,
  title={Seasonal contrast: Unsupervised pre-training from uncurated remote sensing data},
  author={Manas, Oscar and Lacoste, Alexandre and Gir{\'o}-i-Nieto, Xavier and Vazquez, David and Rodriguez, Pau},
  booktitle={Proceedings of the IEEE/CVF International Conference on Computer Vision},
  pages={9414--9423},
  year={2021}
}

@article{shabbir2025thinkgeo,
  title={ThinkGeo: Evaluating Tool-Augmented Agents for Remote Sensing Tasks},
  author={Shabbir, Akashah and Munir, Muhammad Akhtar and Dudhane, Akshay and Sheikh, Muhammad Umer and Khan, Muhammad Haris and Fraccaro, Paolo and Moreno, Juan Bernabe and Khan, Fahad Shahbaz and Khan, Salman},
  journal={arXiv preprint arXiv:2505.23752},
  year={2025}
}

@article{carruthers2002cognitive,
  title={The cognitive functions of language},
  author={Carruthers, Peter},
  journal={Behavioral and brain sciences},
  volume={25},
  number={6},
  pages={657--674},
  year={2002},
  publisher={Cambridge University Press}
}

@inproceedings{bountos2025fomo,
  title={Fomo: Multi-modal, multi-scale and multi-task remote sensing foundation models for forest monitoring},
  author={Bountos, Nikolaos Ioannis and Ouaknine, Arthur and Papoutsis, Ioannis and Rolnick, David},
  booktitle={Proceedings of the AAAI Conference on Artificial Intelligence},
  volume={39},
  number={27},
  pages={27858--27868},
  year={2025}
}

@article{hong2023spectralgpt,
  title={SpectralGPT: Spectral remote sensing foundation model},
  author={Hong, Danfeng and Zhang, Bing and Li, Xuyang and Li, Yuxuan and Li, Chenyu and Yao, Jing and Yokoya, Naoto and Li, Hao and Ghamisi, Pedram and Jia, Xiuping and others},
  journal={arXiv preprint arXiv:2311.07113},
  year={2023}
}

@article{wu2025semantic,
  title={A semantic-enhanced multi-modal remote sensing foundation model for Earth observation},
  author={Wu, Kang and Zhang, Yingying and Ru, Lixiang and Dang, Bo and Lao, Jiangwei and Yu, Lei and Luo, Junwei and Zhu, Zifan and Sun, Yue and Zhang, Jiahao and others},
  journal={Nature Machine Intelligence},
  volume={7},
  number={8},
  pages={1235--1249},
  year={2025},
  publisher={Nature Publishing Group UK London}
}

@article{xiong2024neural,
  title={Neural plasticity-inspired multimodal foundation model for earth observation},
  author={Xiong, Zhitong and Wang, Yi and Zhang, Fahong and Stewart, Adam J and Hanna, Jo{\"e}lle and Borth, Damian and Papoutsis, Ioannis and Saux, Bertrand Le and Camps-Valls, Gustau and Zhu, Xiao Xiang},
  journal={arXiv preprint arXiv:2403.15356},
  year={2024}
}

@inproceedings{han2024bridging,
  title={Bridging remote sensors with multisensor geospatial foundation models},
  author={Han, Boran and Zhang, Shuai and Shi, Xingjian and Reichstein, Markus},
  booktitle={Proceedings of the ieee/cvf conference on computer vision and pattern recognition},
  pages={27852--27862},
  year={2024}
}

@inproceedings{yang2010bag,
  title={Bag-of-visual-words and spatial extensions for land-use classification},
  author={Yang, Yi and Newsam, Shawn},
  booktitle={Proceedings of the 18th SIGSPATIAL international conference on advances in geographic information systems},
  pages={270--279},
  year={2010}
}

@article{cheng2017remote,
  title={Remote sensing image scene classification: Benchmark and state of the art},
  author={Cheng, Gong and Han, Junwei and Lu, Xiaoqiang},
  journal={Proceedings of the IEEE},
  volume={105},
  number={10},
  pages={1865--1883},
  year={2017},
  publisher={IEEE}
}

@inproceedings{dota,
  title={DOTA: A large-scale dataset for object detection in aerial images},
  author={Xia, Gui-Song and Bai, Xiang and Ding, Jian and Zhu, Zhen and Belongie, Serge and Luo, Jiebo and Datcu, Mihai and Pelillo, Marcello and Zhang, Liangpei},
  booktitle={Proceedings of the IEEE conference on computer vision and pattern recognition},
  pages={3974--3983},
  year={2018}
}

@article{rsvqa,
  title={RSVQA: Visual question answering for remote sensing data},
  author={Lobry, Sylvain and Marcos, Diego and Murray, Jesse and Tuia, Devis},
  journal={IEEE Transactions on Geoscience and Remote Sensing},
  volume={58},
  number={12},
  pages={8555--8566},
  year={2020},
  publisher={IEEE}
}

@article{nwpu-caption,
  title={NWPU-captions dataset and MLCA-net for remote sensing image captioning},
  author={Cheng, Qimin and Huang, Haiyan and Xu, Yuan and Zhou, Yuzhuo and Li, Huanying and Wang, Zhongyuan},
  journal={IEEE Transactions on Geoscience and Remote Sensing},
  volume={60},
  pages={1--19},
  year={2022},
  publisher={IEEE}
}

@article{million-aid,
  title={On creating benchmark dataset for aerial image interpretation: Reviews, guidances, and million-aid},
  author={Long, Yang and Xia, Gui-Song and Li, Shengyang and Yang, Wen and Yang, Michael Ying and Zhu, Xiao Xiang and Zhang, Liangpei and Li, Deren},
  journal={IEEE Journal of selected topics in applied earth observations and remote sensing},
  volume={14},
  pages={4205--4230},
  year={2021},
  publisher={IEEE}
}

@inproceedings{zhu2025skysense,
  title={Skysense-o: Towards open-world remote sensing interpretation with vision-centric visual-language modeling},
  author={Zhu, Qi and Lao, Jiangwei and Ji, Deyi and Luo, Junwei and Wu, Kang and Zhang, Yingying and Ru, Lixiang and Wang, Jian and Chen, Jingdong and Yang, Ming and others},
  booktitle={Proceedings of the Computer Vision and Pattern Recognition Conference},
  pages={14733--14744},
  year={2025}
}

@article{rsvgd,
  title={Rsvg: Exploring data and models for visual grounding on remote sensing data},
  author={Zhan, Yang and Xiong, Zhitong and Yuan, Yuan},
  journal={IEEE Transactions on Geoscience and Remote Sensing},
  volume={61},
  pages={1--13},
  year={2023},
  publisher={IEEE}
}

@inproceedings{skyscript,
  title={Skyscript: A large and semantically diverse vision-language dataset for remote sensing},
  author={Wang, Zhecheng and Prabha, Rajanie and Huang, Tianyuan and Wu, Jiajun and Rajagopal, Ram},
  booktitle={Proceedings of the AAAI Conference on Artificial Intelligence},
  volume={38},
  number={6},
  pages={5805--5813},
  year={2024}
}

@article{vrsbench,
  title={Vrsbench: A versatile vision-language benchmark dataset for remote sensing image understanding},
  author={Li, Xiang and Ding, Jian and Elhoseiny, Mohamed},
  journal={Advances in Neural Information Processing Systems},
  volume={37},
  pages={3229--3242},
  year={2024}
}

@article{rs5m,
  title={Rs5m and georsclip: A large scale vision-language dataset and a large vision-language model for remote sensing},
  author={Zhang, Zilun and Zhao, Tiancheng and Guo, Yulong and Yin, Jianwei},
  journal={IEEE Transactions on Geoscience and Remote Sensing},
  year={2024},
  publisher={IEEE}
}

@inproceedings{geobench,
  title={Geobench-vlm: Benchmarking vision-language models for geospatial tasks},
  author={Danish, Muhammad and Munir, Muhammad Akhtar and Shah, Syed Roshaan Ali and Kuckreja, Kartik and Khan, Fahad Shahbaz and Fraccaro, Paolo and Lacoste, Alexandre and Khan, Salman},
  booktitle={Proceedings of the IEEE/CVF International Conference on Computer Vision},
  pages={7132--7142},
  year={2025}
}

@article{an2024choice,
  title={CHOICE: Benchmarking the Remote Sensing Capabilities of Large Vision-Language Models},
  author={An, Xiao and Sun, Jiaxing and Gui, Zihan and He, Wei},
  journal={arXiv preprint arXiv:2411.18145},
  year={2024}
}

@article{lu2017exploring,
  title={Exploring models and data for remote sensing image caption generation},
  author={Lu, Xiaoqiang and Wang, Binqiang and Zheng, Xiangtao and Li, Xuelong},
  journal={IEEE Transactions on Geoscience and Remote Sensing},
  volume={56},
  number={4},
  pages={2183--2195},
  year={2017},
  publisher={IEEE}
}

@article{li2018multi,
  title={Multi-modal gated recurrent units for image description},
  author={Li, Xuelong and Yuan, Aihong and Lu, Xiaoqiang},
  journal={Multimedia Tools and Applications},
  volume={77},
  number={22},
  pages={29847--29869},
  year={2018},
  publisher={Springer}
}

@article{wang2019semantic,
  title={Semantic descriptions of high-resolution remote sensing images},
  author={Wang, Binqiang and Lu, Xiaoqiang and Zheng, Xiangtao and Li, Xuelong},
  journal={IEEE Geoscience and Remote Sensing Letters},
  volume={16},
  number={8},
  pages={1274--1278},
  year={2019},
  publisher={IEEE}
}

@article{lu2019sound,
  title={Sound active attention framework for remote sensing image captioning},
  author={Lu, Xiaoqiang and Wang, Binqiang and Zheng, Xiangtao},
  journal={IEEE Transactions on Geoscience and Remote Sensing},
  volume={58},
  number={3},
  pages={1985--2000},
  year={2019},
  publisher={IEEE}
}

@inproceedings{xu2015show,
  title={Show, attend and tell: Neural image caption generation with visual attention},
  author={Xu, Kelvin and Ba, Jimmy and Kiros, Ryan and Cho, Kyunghyun and Courville, Aaron and Salakhudinov, Ruslan and Zemel, Rich and Bengio, Yoshua},
  booktitle={International conference on machine learning},
  pages={2048--2057},
  year={2015},
  organization={PMLR}
}

@article{sumbul2020sd,
  title={SD-RSIC: Summarization-driven deep remote sensing image captioning},
  author={Sumbul, Gencer and Nayak, Sonali and Demir, Beg{\"u}m},
  journal={IEEE Transactions on Geoscience and Remote Sensing},
  volume={59},
  number={8},
  pages={6922--6934},
  year={2020},
  publisher={IEEE}
}

@article{wang2020retrieval,
  title={Retrieval topic recurrent memory network for remote sensing image captioning},
  author={Wang, Binqiang and Zheng, Xiangtao and Qu, Bo and Lu, Xiaoqiang},
  journal={IEEE Journal of Selected Topics in Applied Earth Observations and Remote Sensing},
  volume={13},
  pages={256--270},
  year={2020},
  publisher={IEEE}
}

@article{hoxha2021novel,
  title={A novel SVM-based decoder for remote sensing image captioning},
  author={Hoxha, Genc and Melgani, Farid},
  journal={IEEE Transactions on Geoscience and Remote Sensing},
  volume={60},
  pages={1--14},
  year={2021},
  publisher={IEEE}
}

@article{hoxha2023improving,
  title={Improving image captioning systems with postprocessing strategies},
  author={Hoxha, Genc and Scuccato, Giacomo and Melgani, Farid},
  journal={IEEE Transactions on Geoscience and Remote Sensing},
  volume={61},
  pages={1--13},
  year={2023},
  publisher={IEEE}
}

@article{silva2024large,
  title={Large language models for captioning and retrieving remote sensing images},
  author={Silva, Jo{\~a}o Daniel and Magalh{\~a}es, Jo{\~a}o and Tuia, Devis and Martins, Bruno},
  journal={arXiv preprint arXiv:2402.06475},
  year={2024}
}

@inproceedings{wang2024skyscript,
  title={Skyscript: A large and semantically diverse vision-language dataset for remote sensing},
  author={Wang, Zhecheng and Prabha, Rajanie and Huang, Tianyuan and Wu, Jiajun and Rajagopal, Ram},
  booktitle={Proceedings of the AAAI Conference on Artificial Intelligence},
  volume={38},
  number={6},
  pages={5805--5813},
  year={2024}
}

@article{lin2025rs,
  title={Rs-moe: A vision-language model with mixture of experts for remote sensing image captioning and visual question answering},
  author={Lin, Hui and Hong, Danfeng and Ge, Shuhang and Luo, Chuyao and Jiang, Kai and Jin, Hao and Wen, Congcong},
  journal={IEEE Transactions on Geoscience and Remote Sensing},
  year={2025},
  publisher={IEEE}
}

@inproceedings{soni2025earthdial,
  title={Earthdial: Turning multi-sensory earth observations to interactive dialogues},
  author={Soni, Sagar and Dudhane, Akshay and Debary, Hiyam and Fiaz, Mustansar and Munir, Muhammad Akhtar and Danish, Muhammad Sohail and Fraccaro, Paolo and Watson, Campbell D and Klein, Levente J and Khan, Fahad Shahbaz and others},
  booktitle={Proceedings of the Computer Vision and Pattern Recognition Conference},
  pages={14303--14313},
  year={2025}
}

@article{jaech2024openai,
  title={Openai o1 system card},
  author={Jaech, Aaron and Kalai, Adam and Lerer, Adam and Richardson, Adam and El-Kishky, Ahmed and Low, Aiden and Helyar, Alec and Madry, Aleksander and Beutel, Alex and Carney, Alex and others},
  journal={arXiv preprint arXiv:2412.16720},
  year={2024}
}

@article{hurst2024gpt,
  title={Gpt-4o system card},
  author={Hurst, Aaron and Lerer, Adam and Goucher, Adam P and Perelman, Adam and Ramesh, Aditya and Clark, Aidan and Ostrow, AJ and Welihinda, Akila and Hayes, Alan and Radford, Alec and others},
  journal={arXiv preprint arXiv:2410.21276},
  year={2024}
}

@article{luo2025large,
  title={When Large Vision-Language Model Meets Large Remote Sensing Imagery: Coarse-to-Fine Text-Guided Token Pruning},
  author={Luo, Junwei and Zhang, Yingying and Yang, Xue and Wu, Kang and Zhu, Qi and Liang, Lei and Chen, Jingdong and Li, Yansheng},
  journal={arXiv preprint arXiv:2503.07588},
  year={2025}
}

@misc{li2024llavanext-strong,
    title={LLaVA-NeXT: Stronger LLMs Supercharge Multimodal Capabilities in the Wild},
    url={https://llava-vl.github.io/blog/2024-05-10-llava-next-stronger-llms/},
    author={Li, Bo and Zhang, Kaichen and Zhang, Hao and Guo, Dong and Zhang, Renrui and Li, Feng and Zhang, Yuanhan and Liu, Ziwei and Li, Chunyuan},
    month={May},
    year={2024}
}

@inproceedings{zhai2023sigmoid,
  title={Sigmoid loss for language image pre-training},
  author={Zhai, Xiaohua and Mustafa, Basil and Kolesnikov, Alexander and Beyer, Lucas},
  booktitle={Proceedings of the IEEE/CVF international conference on computer vision},
  pages={11975--11986},
  year={2023}
}

@article{cai2024internlm2,
  title={Internlm2 technical report},
  author={Cai, Zheng and Cao, Maosong and Chen, Haojiong and Chen, Kai and Chen, Keyu and Chen, Xin and Chen, Xun and Chen, Zehui and Chen, Zhi and Chu, Pei and others},
  journal={arXiv preprint arXiv:2403.17297},
  year={2024}
}

@article{peng2023instruction,
  title={Instruction tuning with gpt-4},
  author={Peng, Baolin and Li, Chunyuan and He, Pengcheng and Galley, Michel and Gao, Jianfeng},
  journal={arXiv preprint arXiv:2304.03277},
  year={2023}
}

@article{krause2016multiplicative,
  title={Multiplicative LSTM for sequence modelling},
  author={Krause, Ben and Lu, Liang and Murray, Iain and Renals, Steve},
  journal={arXiv preprint arXiv:1609.07959},
  year={2016}
}

@inproceedings{waqas2019isaid,
  title={isaid: A large-scale dataset for instance segmentation in aerial images},
  author={Waqas Zamir, Syed and Arora, Aditya and Gupta, Akshita and Khan, Salman and Sun, Guolei and Shahbaz Khan, Fahad and Zhu, Fan and Shao, Ling and Xia, Gui-Song and Bai, Xiang},
  booktitle={Proceedings of the IEEE/CVF conference on computer vision and pattern recognition workshops},
  pages={28--37},
  year={2019}
}

@article{xia2017aid,
  title={AID: A benchmark data set for performance evaluation of aerial scene classification},
  author={Xia, Gui-Song and Hu, Jingwen and Hu, Fan and Shi, Baoguang and Bai, Xiang and Zhong, Yanfei and Zhang, Liangpei and Lu, Xiaoqiang},
  journal={IEEE Transactions on Geoscience and Remote Sensing},
  volume={55},
  number={7},
  pages={3965--3981},
  year={2017},
  publisher={IEEE}
}

@article{bommasani2021opportunities,
  title={On the opportunities and risks of foundation models},
  author={Bommasani, Rishi},
  journal={arXiv preprint arXiv:2108.07258},
  year={2021}
}

@inproceedings{liu2024rotated,
  title={Rotated multi-scale interaction network for referring remote sensing image segmentation},
  author={Liu, Sihan and Ma, Yiwei and Zhang, Xiaoqing and Wang, Haowei and Ji, Jiayi and Sun, Xiaoshuai and Ji, Rongrong},
  booktitle={Proceedings of the IEEE/CVF Conference on Computer Vision and Pattern Recognition},
  pages={26658--26668},
  year={2024}
}

@inproceedings{pang2025vhm,
  title={Vhm: Versatile and honest vision language model for remote sensing image analysis},
  author={Pang, Chao and Weng, Xingxing and Wu, Jiang and Li, Jiayu and Liu, Yi and Sun, Jiaxing and Li, Weijia and Wang, Shuai and Feng, Litong and Xia, Gui-Song and others},
  booktitle={Proceedings of the AAAI Conference on Artificial Intelligence},
  volume={39},
  number={6},
  pages={6381--6388},
  year={2025}
}

@article{hu2025ringmo,
  title={RingMo-Agent: A Unified Remote Sensing Foundation Model for Multi-Platform and Multi-Modal Reasoning},
  author={Hu, Huiyang and Wang, Peijin and Feng, Yingchao and Wei, Kaiwen and Yin, Wenxin and Diao, Wenhui and Wang, Mengyu and Bi, Hanbo and Kang, Kaiyue and Ling, Tong and others},
  journal={arXiv preprint arXiv:2507.20776},
  year={2025}
}

@article{feng2025earth,
  title={Earth-agent: Unlocking the full landscape of earth observation with agents},
  author={Feng, Peilin and Lv, Zhutao and Ye, Junyan and Wang, Xiaolei and Huo, Xinjie and Yu, Jinhua and Xu, Wanghan and Zhang, Wenlong and Bai, Lei and He, Conghui and others},
  journal={arXiv preprint arXiv:2509.23141},
  year={2025}
}

@article{liu2025towards,
  title={Towards Faithful Reasoning in Remote Sensing: A Perceptually-Grounded GeoSpatial Chain-of-Thought for Vision-Language Models},
  author={Liu, Jiaqi and Sun, Lang and Fu, Ronghao and Yang, Bo},
  journal={arXiv preprint arXiv:2509.22221},
  year={2025}
}

@article{yuan2023rrsis,
  title={Rrsis: Referring remote sensing image segmentation},
  author={Yuan, Zhenghang and Mou, Lichao and Hua, Yuansheng and Zhu, Xiao Xiang},
  journal={arXiv preprint arXiv:2306.08625},
  year={2023}
}

@article{lei2024exploring,
  title={Exploring fine-grained image-text alignment for referring remote sensing image segmentation},
  author={Lei, Sen and Xiao, Xinyu and Zhang, Tianlin and Li, Heng-Chao and Shi, Zhenwei and Zhu, Qing},
  journal={IEEE Transactions on Geoscience and Remote Sensing},
  year={2024},
  publisher={IEEE}
}

@inproceedings{liu2021swin,
  title={Swin transformer: Hierarchical vision transformer using shifted windows},
  author={Liu, Ze and Lin, Yutong and Cao, Yue and Hu, Han and Wei, Yixuan and Zhang, Zheng and Lin, Stephen and Guo, Baining},
  booktitle={Proceedings of the IEEE/CVF international conference on computer vision},
  pages={10012--10022},
  year={2021}
}

@article{cho2023cross,
  title={Cross-aware early fusion with stage-divided vision and language transformer encoders for referring image segmentation},
  author={Cho, Yubin and Yu, Hyunwoo and Kang, Suk-Ju},
  journal={IEEE Transactions on Multimedia},
  volume={26},
  pages={5823--5833},
  year={2023},
  publisher={IEEE}
}

@article{chen2025rsrefseg,
  title={Rsrefseg: Referring remote sensing image segmentation with foundation models},
  author={Chen, Keyan and Zhang, Jiafan and Liu, Chenyang and Zou, Zhengxia and Shi, Zhenwei},
  journal={arXiv preprint arXiv:2501.06809},
  year={2025}
}

@inproceedings{kirillov2023segment,
  title={Segment anything},
  author={Kirillov, Alexander and Mintun, Eric and Ravi, Nikhila and Mao, Hanzi and Rolland, Chloe and Gustafson, Laura and Xiao, Tete and Whitehead, Spencer and Berg, Alexander C and Lo, Wan-Yen and others},
  booktitle={Proceedings of the IEEE/CVF international conference on computer vision},
  pages={4015--4026},
  year={2023}
}

@inproceedings{yao2025remotesam,
  title={Remotesam: Towards segment anything for earth observation},
  author={Yao, Liang and Liu, Fan and Chen, Delong and Zhang, Chuanyi and Wang, Yijun and Chen, Ziyun and Xu, Wei and Di, Shimin and Zheng, Yuhui},
  booktitle={Proceedings of the 33rd ACM International Conference on Multimedia},
  pages={3027--3036},
  year={2025}
}

@article{ni2025unigeoseg,
  title={UniGeoSeg: Towards Unified Open-World Segmentation for Geospatial Scenes},
  author={Ni, Shuo and Wang, Di and Chen, He and Guo, Haonan and Zhang, Ning and Zhang, Jing},
  journal={arXiv preprint arXiv:2511.23332},
  year={2025}
}

@article{yang2025learning,
  title={Learning on the job: An experience-driven self-evolving agent for long-horizon tasks},
  author={Yang, Cheng and Yang, Xuemeng and Wen, Licheng and Fu, Daocheng and Mei, Jianbiao and Wu, Rong and Cai, Pinlong and Shen, Yufan and Deng, Nianchen and Shi, Botian and others},
  journal={arXiv preprint arXiv:2510.08002},
  year={2025}
}

@article{shao2025asking,
  title={Asking like Socrates: Socrates helps VLMs understand remote sensing images},
  author={Shao, Run and Li, Ziyu and Zhang, Zhaoyang and Xu, Linrui and He, Xinran and Yuan, Hongyuan and He, Bolei and Dai, Yongxing and Yan, Yiming and Chen, Yijun and others},
  journal={arXiv preprint arXiv:2511.22396},
  year={2025}
}

\begin{IEEEbiography}[{\includegraphics[width=1in, height=1.25in,clip,keepaspectratio]{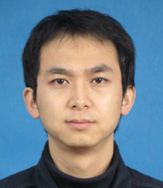}}]{Haifeng Li} received a master's degree in transportation engineering from the South China University of Technology, Guangzhou, China, in 2005 and a Ph.D. degree in photogrammetry and remote sensing from Wuhan University, Wuhan, China, in 2009. He is currently a professor at the School of Geosciences and Info-Physics, Central South University, Changsha, China. He was a research associate at the Department of Land Surveying and Geo-Informatics, The Hong Kong Polytechnic University, Hong Kong, in 2011 and a visiting scholar at the University of Illinois at Urbana-Champaign, Urbana, IL, USA, from 2013 to 2014. He has authored more than 30 journal papers. His current research interests include geo/remote sensing big data, machine/deep learning, and artificial/brain-inspired intelligence. He is a reviewer for many journals.
\end{IEEEbiography}

\begin{IEEEbiography}[{\includegraphics[width=1in,height=1.25in,clip,keepaspectratio]{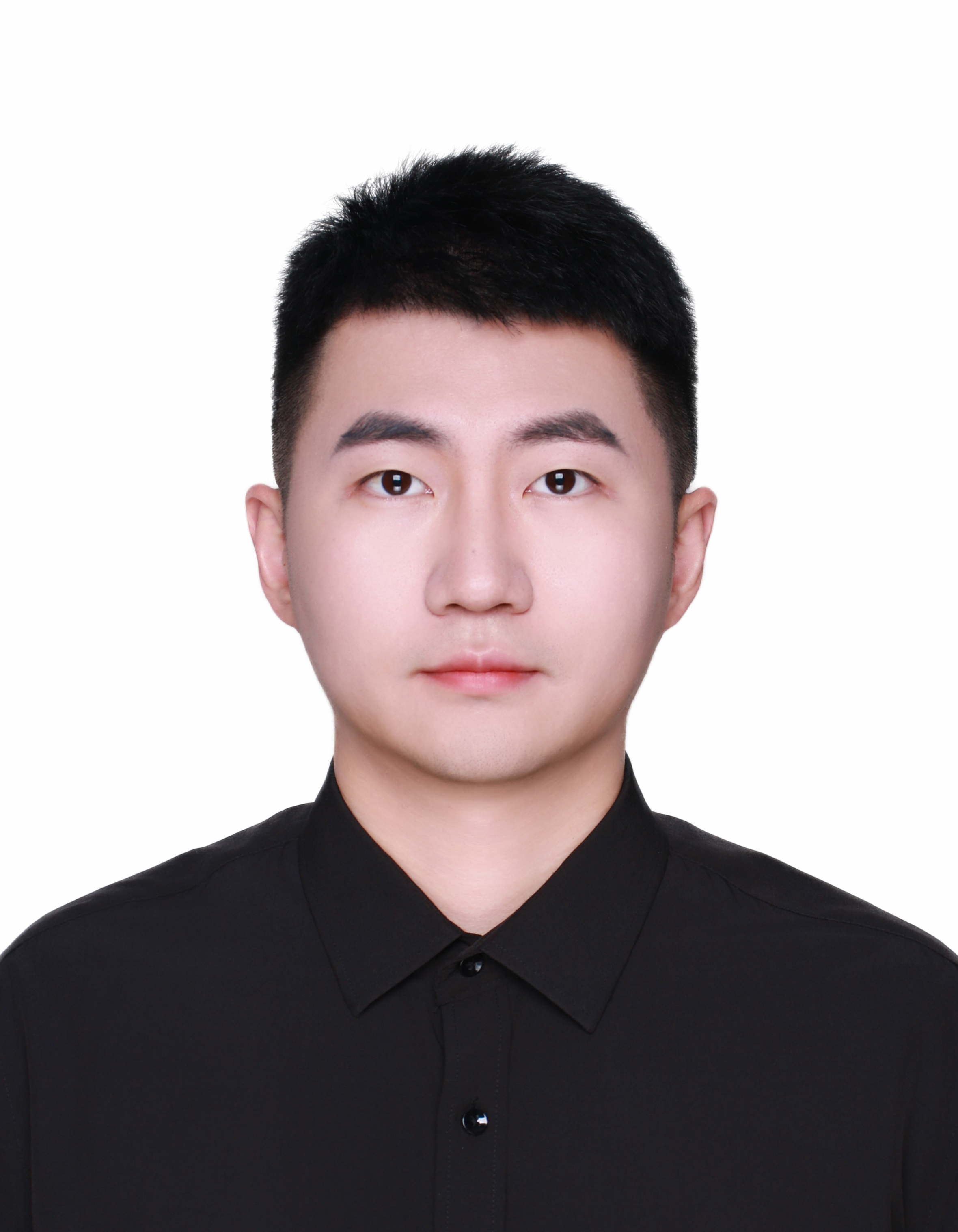}}]{Wang Guo} received an M.Sc. degree in interactive media from University College Cork, Cork, Ireland, in 2019 and is currently pursuing a Ph.D. degree at Central South University, Changsha, China. His research interests include memory models, lifelong learning and remote sensing image processing.
\end{IEEEbiography}

\begin{IEEEbiography}[{\includegraphics[width=1in,height=1.25in,clip,keepaspectratio]{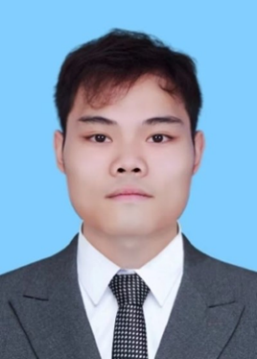}}]{Haiyang Wu} is a Ph.D. candidate in the Department of Geomatics and Remote Sensing Science, Central South University. His research interests include the intelligent interpretation of remote sensing imagery, deep learning, reinforcement learning, and multimodal large-scale models.
\end{IEEEbiography}

\begin{IEEEbiography}[{\includegraphics[width=1in,height=1.25in,clip,keepaspectratio]{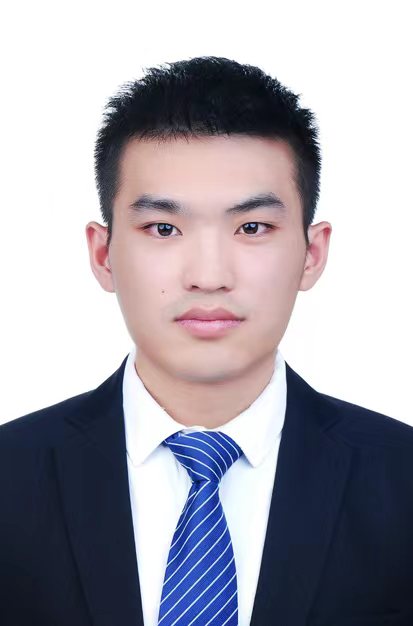}}]{Mengwei Wu} is a Ph.D. student in the Department of Geomatics and Remote Sensing at Central South University. His research interests include multimodal large language models, spatial intelligence, and embodied agents.
\end{IEEEbiography}

\begin{IEEEbiography}[{\includegraphics[width=1in,height=1.25in,clip,keepaspectratio]{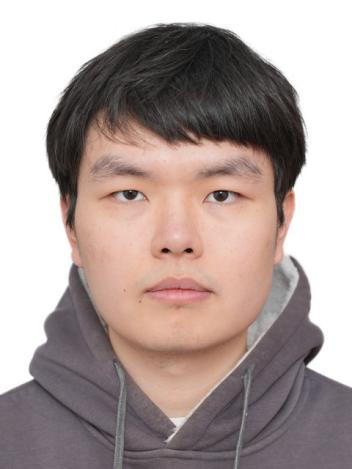}}]{Jipeng Zhang} is a Ph.D. candidate in the Department of Geomatics and Remote Sensing Science, Central South University. His research interests include UAV perception, autonomous navigation, vision-language-action models and deep learning.
\end{IEEEbiography}

\begin{IEEEbiography}[{\includegraphics[width=1in, height=1.25in,clip,keepaspectratio]{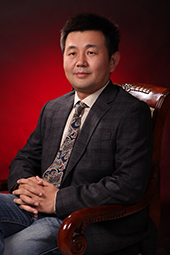}}]{Yu Liu} received B.S., M.S., and Ph.D. degrees from Peking University, Beijing, China, in 1994, 1997, and 2003, respectively. He is currently a Professor at the Institute of Remote Sensing and Geographic Information System, Peking University. His research interests concern mainly big geodata in the humanities and social sciences. 
\end{IEEEbiography}

\begin{IEEEbiography}[{\includegraphics[width=1in, height=1.25in,clip,keepaspectratio]{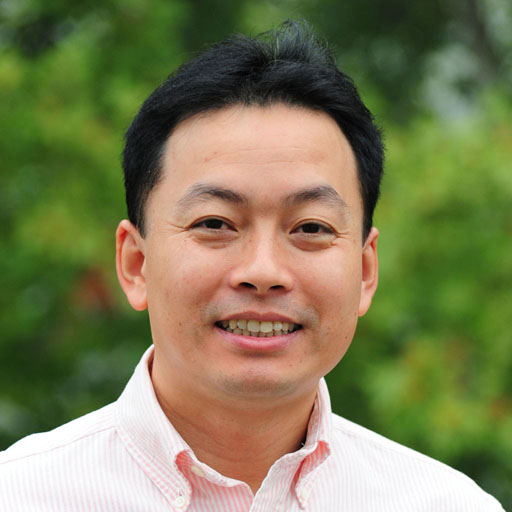}}]{Qing Zhu} received a master's degree in transportation engineering from the South China University of Technology, Guangzhou, China, in 2005 and a Ph.D. degree in photogrammetry and remote sensing from Wuhan University, Wuhan, China, in 2009. He is currently a professor at the School of Geosciences and Info-Physics, Central South University, Changsha, China. He was a research associate in the Department of Land Surveying and Geo-Informatics, The Hong Kong Polytechnic University, Hong Kong, in 2011 and a visiting scholar at the University of Illinois at Urbana-Champaign, Urbana, IL, USA, from 2013 to 2014. He has authored more than 30 journal papers. His current research interests include geo/remote sensing big data, machine/deep learning, and artificial/brain-inspired intelligence. He is a reviewer for many journals.
\end{IEEEbiography}

\begin{IEEEbiography}[{\includegraphics[width=1in,height=1.25in,clip,keepaspectratio]{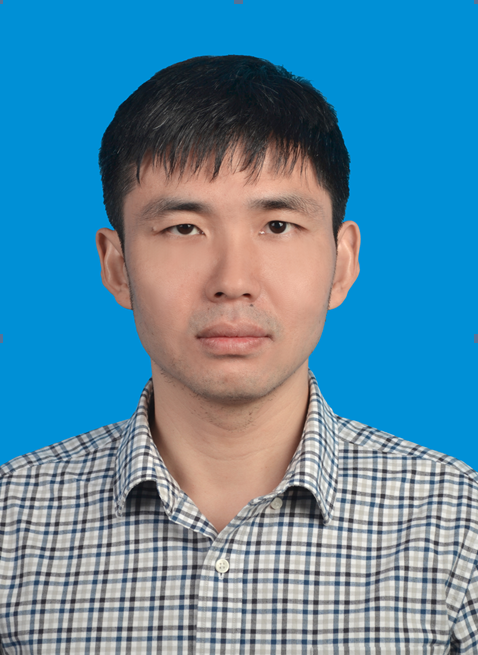}}]{Xin Huang}  (M'13--SM'14--F'25) received a Ph.D. degree in photogrammetry and remote sensing in 2009 from Wuhan University, Wuhan, China. 
He is currently a full Professor at Wuhan University, where he teaches remote sensing, image interpretation, etc. He is the head of the Institute of Remote Sensing Information Processing (IRSIP), Wuhan University. He has published more than 200 peer-reviewed articles (SCI papers) in international journals. He was named as one of the ``Most Cited Chinese Researchers'' by Elsevier and one of the ``Highly Cited Researchers'' by Clarivate.
Dr. Huang's research interests include remote sensing image processing methods and applications. He produced China Land Cover Data (CLCD) from 1980 to 2023, an open-source 30-meter annual land cover dataset of China, and the global impervious surface area from 1972 to 2023 (GISA) with a spatial resolution of 30 meters. He proposed a series of 3D reconstruction and interpretation methods for multiview satellite images, which have been used to generate 3D high-resolution urban land cover maps (ULCM) for China's 50 major cities. 
Dr. Huang has won the First Prize of China Remote Sensing Outstanding Achievement Award, the First Prize of China Geographical Information Science and Technology Progress Award, and the First Prize for China Surveying and Mapping Science and Technology Progress Award. He was the recipient of the Boeing Award from the American Society for Photogrammetry and Remote Sensing (ASPRS) in 2010 and the recipient of the John I. Davidson President's Award from ASPRS in 2018. In addition, he was the winner of the IEEE GRSS Data Fusion Contest in 2014 and 2021. Prof. Huang was an Associate Editor of the IEEE GEOSCIENCE AND REMOTE SENSING LETTERS (2014-2020) and an Associate Editor of the IEEE JOURNAL OF SELECTED TOPICS IN APPLIED EARTH OBSERVATIONS AND REMOTE SENSING (2018-2022) and now serves as an Associate Editor of the IEEE TRANSACTIONS ON GEOSCIENCE AND REMOTE SENSING (since 2022). He is also an editorial board member of the Remote Sensing of Environment (since 2019).
 Dr. Huang was elected as an IEEE Fellow for his contributions to machine learning for remote sensing.
\end{IEEEbiography}

\begin{IEEEbiography}[{\includegraphics[width=1in,height=1.25in,clip,keepaspectratio]{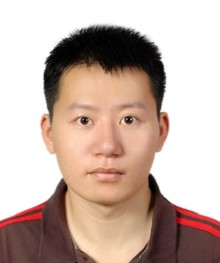}}]{Chao Tao} received a B.S. degree in computational mathematics from the School of Mathematics and Statistics, Huazhong University of Science and Technology, Wuhan, China, in 2007, and a Ph.D. degree in pattern recognition and intelligent systems from the Institution of Pattern Recognition and Artificial Intelligence, Huazhong University of Science and Technology, Wuhan, China, in 2012. He is currently a Professor at the School of Geosciences and Info Physics, Central South University, Changsha, China. He has authored more than 30 peer-reviewed articles in international journals from multiple domains, such as remote sensing and computer vision. His research interests include computer vision, machine learning, deep learning, and their applications in remote sensing. Dr. Tao has frequently served as a reviewer for several international journals, including the IEEE-TGRS, IEEE-GRSL, IEEE-JSTAR, PERS, and ISPRS JPRS. He is also a Communication Evaluation Expert for the National Natural Science Foundation of China.
\end{IEEEbiography}

\end{document}